\documentclass{article}

\usepackage[english]{babel}

\usepackage[letterpaper,top=2cm,bottom=2cm,left=3cm,right=3cm,marginparwidth=1.75cm]{geometry}

\usepackage{gb4e}
\noautomath

\usepackage{amsmath}
\usepackage{graphicx}
\usepackage[final,            
            colorlinks,       
            linkcolor=black,   
            citecolor=black,   
            urlcolor=black,    
            plainpages=false, 
            pdfpagelabels,    
            hypertexnames=false, 
            breaklinks=true
            ]{hyperref}

\usepackage{enumitem, amssymb}
\usepackage[dvipsnames]{xcolor}
\usepackage{multirow,booktabs}
\usepackage{csquotes}
\usepackage{environ}
\usepackage{multicol}
\usepackage{subcaption}
\usepackage{soul}
\let\oldcite\cite
\renewcommand*\cite[1]{[\oldcite{#1}]}

\usepackage{array}
\newcolumntype{P}[1]{>{\centering\arraybackslash}p{#1}}

\usepackage{algorithm}
\usepackage{algpseudocode}

\usepackage{xurl}
\usepackage[style=apa, 
					backend=biber,
            natbib=true,      
            hyperref=true,    
            doi=false,        
            url=false,        
            sortcites=ynt,   
            ]{biblatex}

\newcommand{\proposer}[1]{\textcolor{BurntOrange}{#1}}
\newcommand{\evaluator}[1]{\textcolor{ForestGreen}{#1}}
\newcommand{\reasoner}[1]{\textcolor{RoyalBlue}{#1}}

\usepackage{tikz}

\NewEnviron{prompt}{
\begin{displayquote}
\ttfamily\raggedright\sloppy
\BODY
\end{displayquote}
}

\newcommand{\LM}{%
  \begingroup\normalfont
  \includegraphics[height=\fontcharht\font`B]{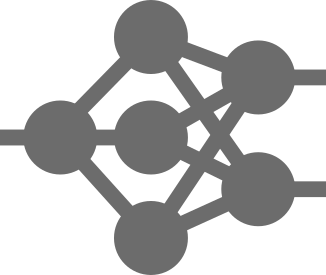}\,%
  \endgroup
}

\newcommand{\RULE}{%
  \begingroup\normalfont
  \includegraphics[height=\fontcharht\font`B]{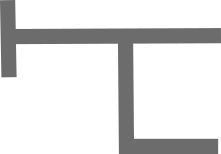}\,%
  \endgroup
}

\newcommand{\SAGE}{%
  \begingroup\normalfont
  \includegraphics[height=\fontcharht\font`B]{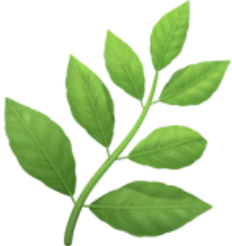}\,%
  \endgroup
}

\usepackage{mdframed}
\newmdenv[
  topline=false,
  bottomline=false,
  rightline=false,
  linecolor=gray,
  linewidth=1pt,
  skipabove=\topsep,
  skipbelow=\topsep,
  fontcolor=gray
]{draftytext}

\title{Computational models of pragmatic reasoning with flexible generation of meaning and expression alternatives}
\date{}
\author{
    Polina Tsvilodub$^{1}$\footnote{Corresponding author: polina.tsvilodub@uni-tuebingen.de}, Fausto Carcassi$^{2}$, Michael Franke$^{1}$\\
    {\small
    $^{1}$University of T\"ubingen, $^{2}$University of Amsterdam
    }
}

\begin{document}
\maketitle


\begin{abstract}
Pragmatic language use requires reasoning about alternatives: the alternative expressions a speaker might have chosen, or the alternative interpretations a listener might entertain.
Formal and computational models of pragmatics must therefore specify the sets of alternatives that interlocutors reason over, which is often done through manual specification.
Here we propose a framework, ScAffolded Generative models for Explanation (SAGE), that combines the explanatory transparency of cognitive models with the generative flexibility of language models (LMs).
SAGE decomposes a pragmatic process into three kinds of modules: \emph{proposers}, which use LMs to generate an open-ended space of candidate alternatives; \emph{evaluators}, which assess those alternatives (e.g., their semantics, complexity, or typicality); and \emph{selectors}, which implement the rule-based computational steps of a cognitively motivated task analysis.
We assess SAGE in three case studies spanning pragmatic generation and interpretation—referential expression generation, manner (M-)implicatures, and Gricean conversational implicatures.
SAGE models are evaluated critically using established methods from computational cognitive modeling, including ablations, baseline comparisons, and quantitative fit to human data.
Across studies, SAGE models achieved high accuracy and often outperformed baselines, but component-level analyses reveal an asymmetry: LM proposers reliably generated alternatives well-suited to pragmatic modeling, whereas LM evaluators are better at providing intuitive judgements rather than judgements of theoretical or formal measures.
We discuss the promise and the limitations of neuro-symbolic models as candidate explanatory accounts of human pragmatic language use.


\end{abstract}

\section{Introduction}
\label{section:introduction}

Humans produce informative yet relevant and efficient utterances and understand them effortlessly across a vast variety of contexts.
Such pragmatic language use has traditionally been explained by identifying general principles of cooperative communication \citep[e.g.,][]{grice1975logic,atlas1981clefts,HornDivisionofLabor1984,SperberWilson1995:Relevance:-Comm,levinson2000presumptive}. If the speaker is assumed to follow these principles, listeners may derive rich contextual meanings beyond the literal meanings of expressions. 
Much subsequent work has attempted to spell out these pragmatic reasoning patterns with formal or computational cognitive models, in which speakers select situationally adequate expressions, while listeners reason about the contextually likely explanations for the speaker's behaviour.
Prominent examples are optimality-theory \citep{BlutnerZeevat2004:Optimality-Theo,Rooijvan-RooijFranke2015:Optimality-Theo}, decision or game theory \citep{Parikh1988:Language-and-st,BenzJager2006:Game-Theory-and}, and Bayesian probabilistic modeling \citep{frank2012predicting,degen2023rational}.
Especially the latter has been quite successful recently in providing explanations for various pragmatic phenomena like non-literal language use \citep{kao2014nonliteral, kao2014formalizing}, vagueness \citep{LassiterGoodman2015:Adjectival-vagu}, or various forms of implicature \citep{BergenLevy2012:Thats-what-she-}.


It is widely acknowledged that pragmatic language use involves reasoning about alternatives: be that alternative expressions that the speaker considers, or alternative interpretations that the listener entertains~\citep[e.g.,][]{FoxKatzir2011:On-the-Characte,BuccolaKriz2021:Conceptual-alte,ReppSpalek2021:The-Role-of-Alt,GotznerRomoli2022:Meaning-and-Alt}.
As a result, formal or computational models have to specify alternative expressions and alternative meanings that the interlocutors reason about.
While principled theories exist for the generation of alternative expressions \citep[e.g.,][]{Katzir2007:Structurally-De} and alternative meanings \citep[e.g.,][]{Franke2011:Quantity-Implic}, they have not led to computational models that can deal with open-ended language, rather than restricted sets of examples.\footnote{The problem of specifying alternatives for reasoning arises more generally in \textit{discrete-state reasoning models}, where decision options, world states or similar discrete entities are subject to weighting or deliberation \citep[e.g.,][]{RutarWolff2022:Structure-Learn, ,wong2025modeling}.}
A crucial difficulty is that alternatives depend on relevant \textit{world knowledge} in open-ended contexts \citep{lake2017building}, which may comprise alternative expressions and meanings, grammatical competence of a language, common sense knowledge about events, objects and their features, knowledge of social conventions and much more. 

Modeling practice offers three main strategies for supplying these alternatives and evaluating them in specific experimental contexts.
First, researchers can manually specify a small space of relevant alternatives and evaluation outputs. 
The specification can be based on theoretical considerations, on researchers' intuitions, or constructed so as to optimize the model's predictions. 
Manual specification is commonly used, but limits the applicability and flexible generalizability of the model to novel and possibly open-ended contexts.
Second, world knowledge can be elicited in auxiliary experiments with human participants. This is often done for probabilistic reasoning models to supply information about intuitive priors \citep[e.g.,][]{kao2014nonliteral,franke2016does}.
However, this approach is expensive and cumbersome, and it has, so far, not been widely used for the elicitation of information about interpretation and utterance alternatives.
Finally, world knowledge could be generated and evaluated by AI models, in particular language models (LMs), statistically trained tools that can easily be queried for reasoning alternatives and evaluations in `\textit{ad hoc} auxiliary elicitation experiments.'
In this paper, we explore this third approach through detailed and comprehensive case studies. 

Recent state-of-the-art LMs are a natural choice for enriching the toolkit for supplying and evaluating alternatives in cognitive modeling, because they complement discrete-state reasoning models. While lacking procedural transparency and, arguably, explanatory potential \citep{Deemter2023:Dimensions-of-E}, LMs produce natural-sounding language, scaling well to different contexts, domains and even to impressive performance on different non-linguistic tasks \citep{devlin-etal-2019-bert, srivastava2023-BIGbench, PerezRinger2023:Discovering-Lan, openai2023gpt4, touvron2023llama}. 
LMs can be used as knowledge bases \citep{petroni2019language} or even as approximations of human intuitive commonsense knowledge \citep{petroni2019language, ParkOBrien2023:Generative-Agen}, leading to a recent line of work which integrates LMs in larger computational procedures \citep[e.g.,][]{yao2023tree, liu-etal-2022-generated}, or with various computational tools for, e.g., mathematical problem solving \citep[e.g.,][]{he2023solving}, complex reasoning \citep[e.g.,][]{creswell2022selection, he2023solving, paranjape2023art, poesia2023certified} or programming tasks \citep[e.g.,][]{gao2022pal}.

\begin{figure*}[t!]
    \centering
    \includegraphics[width=\linewidth]{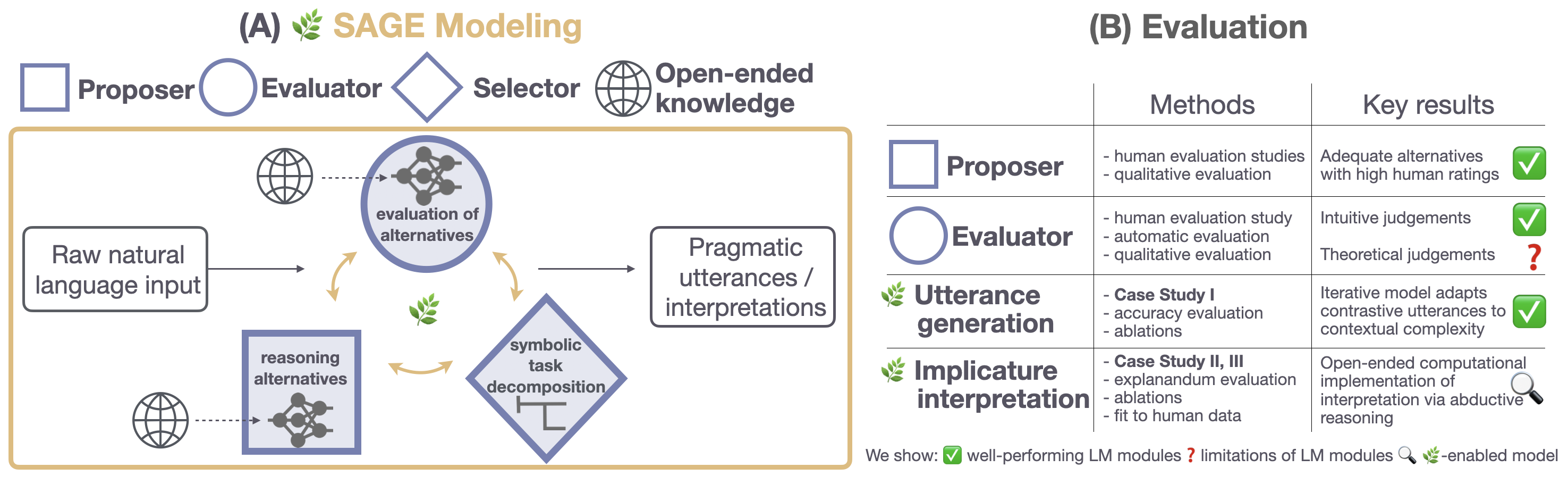}
    \caption{
    \textbf{(A)} Overview of our framework ScAffolded Generative models for Explanation (\textbf{SAGE}, indicate here with the symbol \SAGE) for more open-ended cognitive modeling of human pragmatic language use.
    The framework identifies three kinds of \textcolor[HTML]{7581B3}{\textbf{modules}} that are scaffolded by a cognitively motivated \textcolor[HTML]{DCBC81}{\textbf{explanatory task analysis}}: proposers, evaluators and selectors.
    We use language models to instantiate neural modules (\LM) that propose and flexibly evaluate a more open-ended space of reasoning alternatives often required by computational models, which are harnessed in explicit selector modules (\RULE).
    The model takes as input text datasets and may be designed to output textual or numerical predictions explaining pragmatic language understanding or generation.
    \textbf{(B)} The explanatory adequacy of the computational model and the performance of the neural components is carefully assessed through common methods and measures from computational cognitive modeling.
    Specifically, we stress-test the two new modules (proposer and evaluator), and assess SAGE on case studies of pragmatic language generation as well as interpretation.
    Key results of the case studies and evaluations are summarized in the table.
    }
    \label{fig:sage-overview}
\end{figure*}

In this paper, we outline a novel framework that aims to combine the strength of cognitive models and LMs to yield transparent and, in principle, explanatory computational models of human pragmatic language use that can be applied to contextualized language interpretation and generation beyond pre-specified sets of alternatives. 
We call this framework \textbf{ScAffolded Generative models for Explanation (SAGE \SAGE)}.
Figure~\ref{fig:sage-overview}(A) provides a schematic visualization of the SAGE framework.
The framework implements an explicit task analysis of a cognitive process as information flow between \textit{modules}. 
We distinguish three kinds of modules.
First,  \textit{proposers} generate alternatives for reasoning or choice. 
In our case studies, proposer modules are instantiated with language models, so that the overall model is open-ended and does not completely rely on manual specification of sets of alternatives. 
Second,  \textit{evaluators} process potentially unrestricted alternatives generated in proposer modules. 
This may include, e.g., judging how probable alternative meanings are, or whether a particular expression is true in a given situation.
Evaluators interface rule-based components with LM-generated alternatives.
The information from proposers and generators is harnessed via \textit{selector} modules, which instantiate the computational steps of the assumed task analysis, beyond generation and evaluation of alternatives. 
As we demonstrate in this paper, the resulting computational model can be evaluated through repeated simulation on a dataset of target test items, and can be directly compared to empirical data from human experiments.

SAGE models are a simple form of neuro-symbolic models, combining neural generation and evaluation (using LMs) with a symbolic or algorithmic task-decomposition, thus harnessing the respective strengths of these approaches while compensating their individual limitations \citep{romero2023synergistic, bhuyan2024neuro}.
The SAGE approach also shares clear parallels with recent work on LM agents, where LMs supply information to explanatory discrete-state reasoning models \citep[e.g.,][]{liu2023agentbench, sumers2023cognitive, shinn2023reflexion, wang2024survey}.
However, LM agents have mostly been used for practical applications and improving performance on specific tasks, with less emphasis on \textit{explaining} human reasoning or language use \citep[but see][]{nye2021improving,romero2023synergistic,wong2023word, frank2025cognitive}.
In line with other attempts to bridge the gap between language models and cognitive models \citep[e.g.,][]{bourgin2019cognitive, nye2021improving, binz2023turning, frank2023large, wong2023word, BinzAkata2024:Centaur:-a-foun, FrankeTsvilodub2024:Bayesian-Statis, frank2025cognitive, sucholutsky2025using}, in this paper we explore the potential of neuro-symbolic models as \textit{cognitive models} to explain human language use \citep[similar to][]{tsvilodub2025integrating, tsvilodub2025non, spinoso2025rsa}.

Since the requirements for trustworthy empirical explanations are very high, we focus on exploring the explanatory potential of the SAGE approach for selected tasks, using established tools of computational cognitive science, such as systematic comparisons of models in terms of task performance, predictive accuracy given an explanandum, and quantitative fit to human data (see Figure~\ref{fig:sage-overview}(B) for a schematic overview).
We return to the critical discussion of whether SAGE models might (ever) be explanatory models for the study of human cognition in Section~\ref{section:discussion}.

In the following, we report three case studies selected to cover different aspects of both pragmatic language generation and interpretation across different tasks and for different explanatory goals (see Table~\ref{tab:case-studies-introduction} for overview).
The first case study (Section~\ref{section:utterance-production} ) focuses on generation of referential expressions  \citep{krahmer2012computational}, a task that has traditionally served as a testbed for computational pragmatic modeling and computational linguistics. 
This builds on prior work by us \citep{tsvilodub2024cognitive}, building on a classic algorithmic task analysis of referential expression generation as an iterative process \citep[cf.][]{newell1972human, Ferreira2019:A-Mechanistic-F}, e.g., as instantiated in the \textit{Incremental Algorithm} \citep{dale1995computational}.
This process crucially depends on a proper set of alternative expressions from which the pragmatic speaker will construct a pragmatic expression for identifying the target. 
The SAGE model uses an LM-based proposer to produce the alternatives.
To assess the adequacy of candidate expressions, an LM evaluator module evaluates the semantics of the freely-generated alternatives.
We assess this SAGE model through the lens of task performance, i.e., the ability to single out the target referent uniquely. 

Next, Section~\ref{section:m-impl} focuses on modeling pragmatic language interpretation for the special case of so-called M-implicatures \citep{levinson2000presumptive,Rett2020:Manner-implicat}, a phenomenon wherein a pragmatic listener infers that an atypical, linguistically marked expression signals an atypical state of the world.
According to a prominent algorithmic interpretation procedure suggested for \textit{bi-directional optimality theory} \citep{BlutnerBOTJoS2000} by \citep{Jager2002:Some-Notes-on-t}, calculation of M-implicatures depends on assessing typicality of expressions, states and alternatives.
This case study therefore puts to the test whether an LM proposer can automatically generate alternative expressions that naturally encode typicality (i.e., alternative expressions for more typical situations are more typical, than for atypical situations).
Additionally, we investigate whether LM evaluators can be used to provide explicit assessments about the typicality of world states and expressions.
We assess this resulting model based on its ability to predict the theoretically expected M-implicature, i.e., the explanandum, for a set of variable test items.

In the third case study, reported in Section~\ref{sec:caseStudy-III}, we investigate how a single model of pragmatic interpretation fares with a greater variety of implicatures.
The task scaffolding formulates \textit{abductive inference} rooted in rationalizing the speaker's behavior through reasoning about \textit{Gricean conversational maxims} \citep{grice1975logic}.
This case study innovates over the previous two case studies by using LM proposers for generating situationally adequate interpretations. 
These interpretations are generated using LM evaluators that identify aspects of conversational maxims that help rationalize speaker behavior, putting the modules to the test in a complex social reasoning task. 
Given the high complexity of this task, we conduct ablations and careful evaluation of the model's quantitative fit to human data.

The paper concludes in Section~\ref{section:discussion} with a critical reflection on the lessons learned from these case studies and the implications for the general potential of ``neuro-symbolic cognitive models'' of pragmatic language use. 
Our exploratory case studies showed that, while the simple SAGE models tested here exhibit high accuracy and may outperform baselines, this did not automatically entail satisfactory performance of all components of the model (key results of each case study and the framework assessment are shown in Figure~\ref{fig:sage-overview}(B)-(C)).
Our evaluations reveal that, at least under simple prompting, proposers can provide alternatives well-suited for pragmatic modeling, while evaluators' performance was limited to more intuitive, rather than theoretically-grounded, judgments.
Given promising initial results and adequate proposer performance, we identify promising research directions for improving upon the observed limitations towards more open-ended models of pragmatic language use.
We publish all code, materials and data under \url{https://github.com/CogSciPrag/griceChain}. 

\begin{table}
\small
\centering
\begin{tabular}{p{2.4cm}p{3.8cm}p{3.8cm}p{3.8cm}}
\toprule
 &
\textbf{Case study 1} &
\textbf{Case study 2} &
\textbf{Case study 3}
\\ \midrule
\textbf{Perspective} & production & interpretation & interpretation
\\\addlinespace[0.5em]
\textbf{Task} & referential expressions & manner implicature & conversational
\\ &&& implicatures
\\\addlinespace[0.5em]
\textbf{Task analysis} & incremental & bi-directional & Gricean abductive
\\ &  algorithm & optimality & reasoning
\\\addlinespace[0.5em]
\textbf{Proposer} & expressions & expressions & interpretations
\\\addlinespace[0.5em]
\textbf{Evaluator} & semantics & expression~complexity & maxim violation \& its 
\\ & & state likelihood  & likelihood
\\\addlinespace[0.5em]
\textbf{Goal} & accuracy & predicting explanandum & LH of human data
\\ & task performance & & task performance
\\\addlinespace[0.5em]
\textbf{Checks} & ablation &  baseline model & ablation
\\ & & & module replacement
\\\bottomrule
\end{tabular}
\caption{
Overview of the three case studies reported in this paper, which include LM-components in different roles, focus on different aspects of pragmatic language use, and evaluate the SAGE approach on different goals (performance, explanation, fit to human data).
\label{tab:case-studies-introduction}
}
\end{table}

\section{Case study I: Pragmatic utterance generation}
\label{section:utterance-production}

Our first case study focuses on a well-studied and austere task, tapping into context-dependent language use, namely \textit{contrastive utterance generation} in a \textit{reference game} \citep[e.g.,][]{lewis1969convention}.
This task involves two participants, a speaker and a listener, who jointly observe a context, which consists of a set of states. 
The speaker is assigned a \textit{target} state in the context (unknown to the listener) and aims to produce an utterance that uniquely identifies the target state among a set of the possible alternative \textit{distractor} states in the given context.
Communication is successful if the listener can correctly identify the target, given the utterance. 
Crucially, the utterance choice depends on the context, because the context determines which features of the target differentiate it from the distractors and should, therefore, be mentioned in the utterance.

Referential expression generation has been approached through the lens of well-established algorithmic ideas about human problem solving.
One prominent instantiation of such an algorithmic approach is the Incremental Algorithm (IA) by \citet{dale1995computational}.
In simplified terms, the IA constructs a referential expression by first generating some simple utterances, and then running a loop that alternates between (i) an \textit{evaluation step}, where the current utterances are judged with respect to the context at hand, and (ii) a \textit{generation step}, in which the currently best utterances are enriched.
The loop continues until an utterance is found that identifies the referent in the given context (or a hard limit on iterations is reached). 
This way, the search for a good utterance naturally adapts to the complexity of the contextual goal, thereby aiming for a balanced trade-off between effort (iterations) and task success.

The IA is relatively simple and more performant solutions exist for algorithmic generation of referential expressions \citep{krahmer2012computational,gatt2018survey}.
Moreover, the IA is \textit{not} a strong contender for a cognitively plausible model of referential expression generation in humans.
Nevertheless, the IA not only suffices for our modest exploration purposes, but is arguably ideal in this context because it allows us to focus on the key computational principle at its core.
For one, the general idea of incremental search algorithms has a long history \citep{newell1972human}, demonstrable advantages also for LM-based architectures \citep{yao2023tree}, and is supported in the particular domain of language generation \citep{Ferreira2019:A-Mechanistic-F}.
For another, implementing IA as a SAGE model requires exactly the two kinds of LM-based models, each occurring exactly once in a straightforward manner, that are essential for SAGE models: proposers (of utterance alternatives) and evaluators (of their semantic meaning).

In the next section, we present the \textit{iterative model}, a SAGE implementation of an algorithm that is closely inspired by Incremental Algorithm by \citet{dale1995computational}.
We then discuss an ablated (i.e., non-iterative) version of the model and an LM-only baseline.
We finally compare the performance of the models in the context of a simulated experiment which provides full control over the properties of the contexts and simplifies stringent performance assessment for model comparison and criticism.

\subsection{Iterative model}
\label{sec:cs1-iterative-model}
The \textit{iterative model} (IM) takes as input a set of states, represented in a textual format so as to allow using LM-based modules.
One of these states is the target, while the remaining states are distractors. 
The model uses an LM for proposing an utterance and evaluating whether it is semantically compatible with each state.
Rule-based components subsequently check for contrastivity of each proposed utterance (i.e., from how many distractors is the target differentiated?), and potentially iterate the procedure if no fully contrastive utterance has been found (see Figure~\ref{fig:utteranceProductionLoopFlow} for a graphical representation; and see Algorithm~\ref{alg:utteranceProductionLoopFlow} in the appendix for detailed pseudo-code).

More technically, the model takes as input a set of strings describing states, with a target state $s^{*}$ and a non-empty set of distractor states $D$.
It outputs a description of $s^{*}$ that distinguishes it from the states in $D$.
Internally, the model follows a sequence of steps.
First, a set of utterances describing the target state $s^{*}$ is proposed by an \textit{UtteranceProposer} module.
Second, a \textit{SemanticEvaluator} module determines the truth value of all candidate utterances for each state (i.e., whether each utterance if true or false of the target and distractors).
Then, based on the semantic evaluation in the previous step, the model evaluates the \textit{contrastivity} of each candidate utterance generated so far.
Formally, contrastivity is defined as the proportion of distractors for which the utterance is false.
A fully contrastive, i.e., disambiguating, utterance has $C_i = 1$.
If at least one utterance $u_{i}$ has $C_i=1$, the process stops and returns one utterance chosen randomly among all fully disambiguating utterances.
Otherwise, if none of the utterances is fully contrastive, the model selects the most informative (i.e., contrastive) utterances available, and passes them to the UtteranceProposer which is prompted to add some detail about the target, and a new iteration starts.\footnote{
Note that distractors are not taken into account when extending the description of the target.
Instead, we consider only the utterances produced so far along with the speaker's background knowledge about the target (i.e., all attributes that are true of the target).
This potentially allows the model to be applied to non-contrastive utterance generation tasks in future research.
We stress that this design choice is conservative in the sense that it makes it \textit{more difficult} for the model to propose contrastive utterances, which is desirable for current evaluation purposes.
}
This is repeated until a fully contrastive utterance is found or the maximal number of iterations ($n=5$) is reached.
In the latter case, the IM selects one utterance with the highest contrastivity among the utterances currently entertained, which we implement with an information maximization  \textit{InfoMaxSelector} module (also used in the implementation of the ablated, single-pass model introduced below; see Appendix~\ref{app:rsa-reasoner} for details).

An exhaustive search over minimal contrastive expressions becomes computationally intractable with increasingly complex contexts \citep{krahmer2012computational}. To get around this problem, the IM essentially implements an approximation to a search over a (partial) tree of possible referential utterances.
In each iteration, the model adds details to the utterance, in an order proposed by an LM module (instead of having to specify the preference ordering of target attributes manually).
Therefore, the tree depth roughly corresponds to the number of details included in the utterance, while the width corresponds to the number of sampled utterance proposals on each iteration. If the number of proposals by the utterance proposers is less than the number of features in the scene, not all possible utterances will be considered. Moreover, the search is greedy, in the sense that only the most contrastive utterances are passed to the following iteration. 

The technical novelty of this model is the use of an LM to generate, extend, and evaluate utterances. 
This required careful design of the prompts, which can be found in Appendix\ref{app:im-utt-proposer}--\ref{app:im-sem-eval}.
The precise formulation of the prompts was optimized during the development of the model.
This is particularly important for the SemanticEvaluator module.
In the case at hand, the LM performs semantic evaluation by judging whether a sentence (the proposed utterance) is necessarily true on the assumption that another sentence (the description of the target state) is true.
This is, essentially, logical inference, a problem that is notoriously hard for language models \citep{liu2023evaluating}.
Indeed, the semantic evaluation module is sensitive to the formulation of the prompt.
For that reason, we conducted additional evaluations in order to develop more robust prompts.
Appendix~\ref{app:im-sem-eval} gives details of this search process and the evaluation results.

\subsection{Ablated, single-pass model}
\label{sec:cs1-sp-model}
\begin{figure*}[t!]
\centering
\begin{subfigure}[b]{.43\textwidth}
    \includegraphics[width=\linewidth]{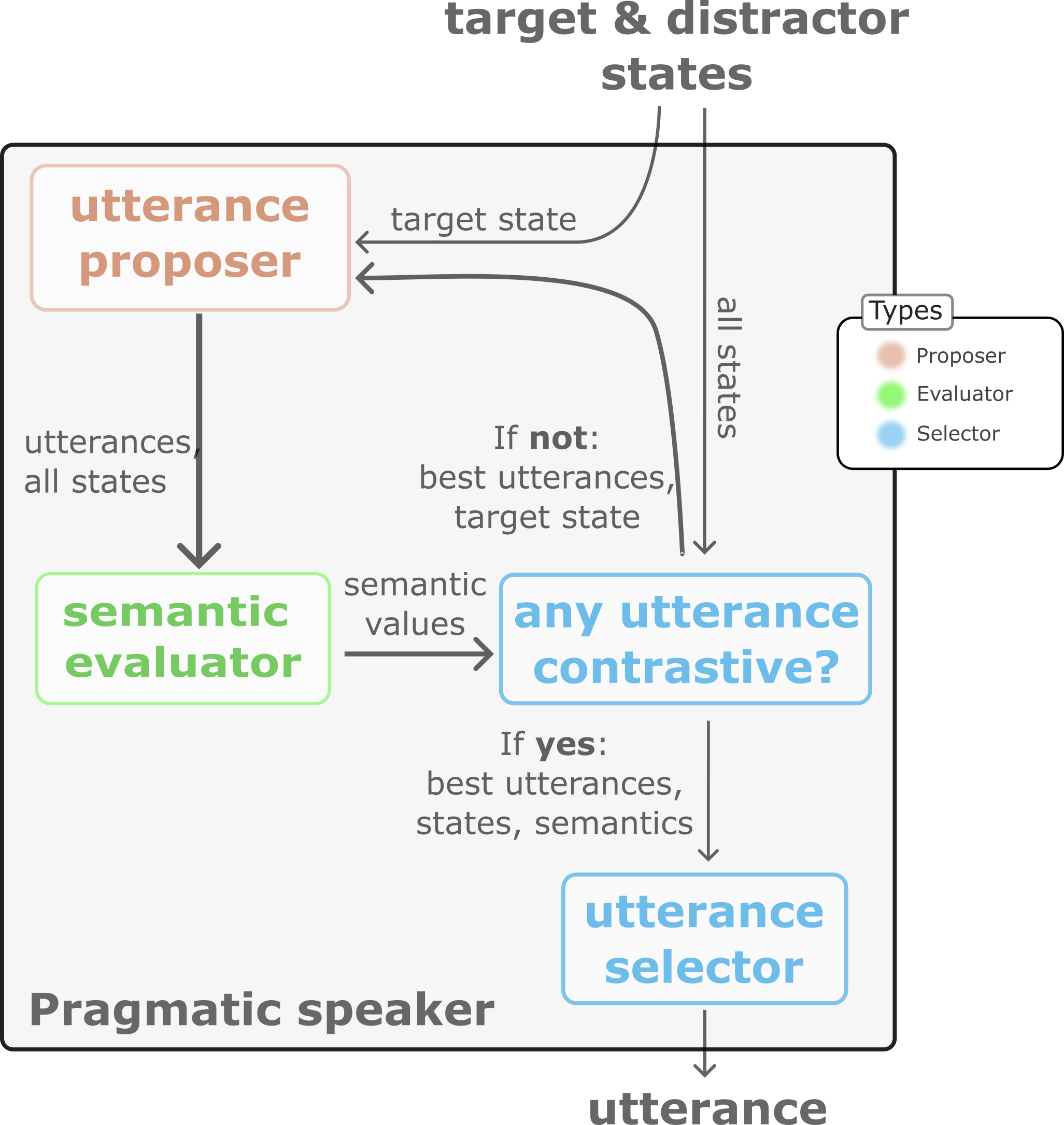}
    \hspace{0.4in}
    \caption{Iterative model\label{fig:utteranceProductionLoopFlow}}
\end{subfigure}
\hfill%
\begin{subfigure}[b]{.4\textwidth}
    \includegraphics[width=\linewidth]{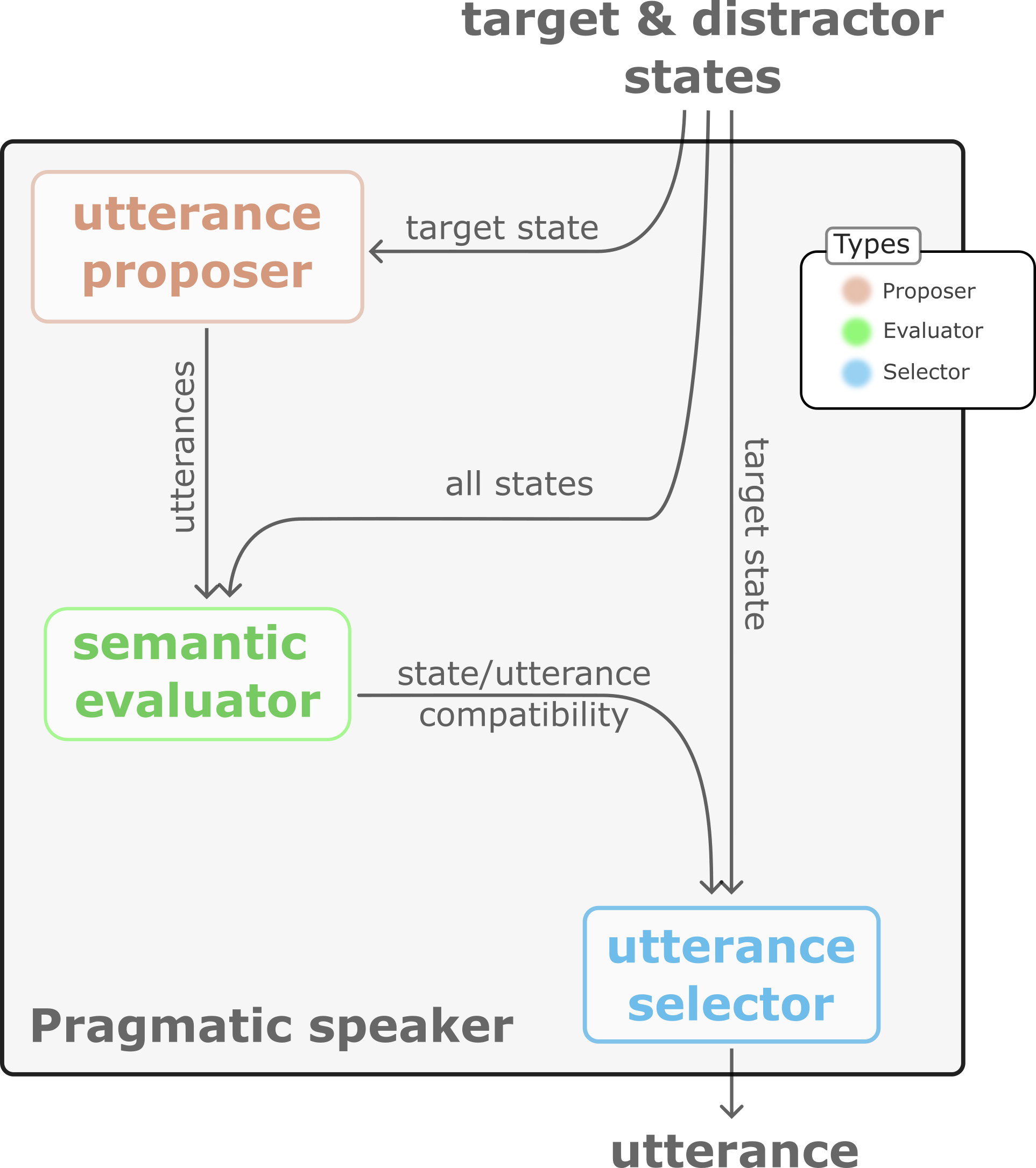}
    \hspace{0.5in}
    \caption{Single-pass model\label{fig:utteranceProductionSimpleFlow}}
\end{subfigure}
\caption{
  Side-by-side illustration of two models for contrastive utterance generation.
  In both models, an utterance proposer initially generates utterances for the target state, which are evaluated for truth by a semantic evaluator for all states in the context (target and distractors).
  At this point the two models diverge.
  In the ablated single-pass model, the utterance selector directly selects an utterance given the semantic evaluator's output.
  In the iterated model, the semantic evaluation result is passed to a contrastivity evaluator.
  If any utterance is contrastive, a contrastive utterance is selected randomly, otherwise the most contrastive utterance is further extended by the utterance proposer and passed back to the semantic evaluator.
  This cycle repeats until a contrastive utterance has been found.
  In sum, the IM can be thought of as an extension where the model dynamically evaluates the produced utterances and improves them in a loop until the task is solved (or a maximum number of iterations is reached).
}
\end{figure*}

The IM presented above can be naturally ablated by stopping the search before any iteration.
In this case, the first batch of produced and interpreted utterances is directly used by the InfoMaxSelector to return the most contrastive utterance, rather than iteratively improving upon it.
Since the task is solved in a single pass, we call this ablated model the \textit{single-pass} (SP) model.
Similarly to the IM, the input to the model is a list of state descriptions, including the target state and one or more distractors.
The model proceeds in three steps, as displayed in Figure~\ref{fig:utteranceProductionSimpleFlow} (for more details see Appendix~\ref{sec:ablated-single-pass} and the Algorithm~\ref{alg:utteranceProductionSimpleFlow} spelled out there).
First, an \textit{UtteranceProposer} module uses an LM call to generate candidate utterances for the target state $s^*$ based on the target state description.
Second, the \textit{SemanticEvaluator} module determines, again via an LM call, the truth value of each candidate utterance for each state (target $s^*$ and distractors $D$).
Lastly, the \textit{InfoMaxSelector} module selects the most specific, i.e., contrastive, true utterance.
All of these modules are functionally similar to the IM modules (see the prompts in Appendix~\ref{sec:utterancesproposer} for full details).


\subsection{Simulation-based model comparison}
\label{section:cs1-results}
\paragraph{Materials.} We test the different models discussed above with states based on the A3DS dataset \citep{tsvilodub2023evaluating}, a textual derivative of the picture-based 3DShapes dataset \citep{burgess20183d}.
The states in the A3DS dataset are unique text-based descriptions of colored geometric shapes in front of various background.
Each state consist of a combination of values for the following attributes:
shape of the object (four different values); size of the object (three values); color of the wall, the floor, the object itself (independent of each other, seven possible values); object position relative to the background (three values).
%
%
There are 12348 possible states in total.
The state descriptions are of the form ``The floor is \{color\}, the wall is \{color\}, the \{color\} \{size\} \{object\} is in the \{position\}'' (see Appendix~\ref{app:im-utt-proposer} for examples of the utterances).\footnote{We also tested supplying state descriptions in the form of lists of ``\{feature-value\}''. However, the unnatural formatting led to poor performance of the LMs.}
This state-space is highly structured and could easily be describes formally with a grammar and a simple interpretation function.
We use this dataset not because it provides an insurmountable obstacle to algorithms like the original IA, but rather because the structured nature of the dataset allows easier systematic construction of controlled context sets and easier annotation of contrastivity in terms of which features distinguish the target from the distractors.

\paragraph{Baseline.} We compare the results of the iterated model (IM) and the single-pass model (SP) against a baseline model.
The baseline consists of a single call to an LM asking for an utterance that solves the task.
For this and all the following experiments, GPT-3.5 (\texttt{gpt-3.5-turbo}, checkpoints of summer 2023) with sampling temperature $\tau=0.1$ was used, unless indicated otherwise.\footnote{
  The backend LM used in our simulations is no longer a frontier model.
  Nothing in these explorations requires using the most advanced model available.
  In fact, using a decent enough non-frontier model may be more insightful for understanding the possibilities but also the limits of the SAGE approach, since our goal is to explore how SAGE-style task-decomposition could possibly work in the context of cognitive modeling.
  We return to this issue more explicitly in the final discussion in Section~\ref{section:discussion}.
}
Following recent results showing that LM performance is improved with examples as well as instructions about the reasoning for solving the task \citep[e.g.,][]{wei2022chain}, we use a one-shot chain-of-thought prompt for the baseline (see Appendix~\ref{app:cs1-baseline} for the full prompt). 

\paragraph{Simulation procedure.}
Each reference game includes a target state sampled at random and one or more distinct distractors.
Each distractor differs from the target by maximally two features, which makes the identification of contrastive features more difficult.
We set the number of distractors to one, four, or eight.
Moreover, for the IM we set the number of utterances generated by the utterance proposer to either four or eight utterances.
For the SP model, the utterance proposer always samples ten utterances.
We test both models as well as the baseline on 100 reference games for each of the simulation configurations.

We evaluate the contrastivity of the final returned utterance, i.e., the proportion of distractors against which the target is set apart by the utterance.
We use careful manual annotation of all simulation runs (performed by one of the authors), because the contrastivity calculation builds on the semantic evaluator, which may be challenging for LMs. This makes the evaluation more robust and less circular.

\begin{figure*}[t!]
\centering
    \includegraphics[width=0.85\linewidth]{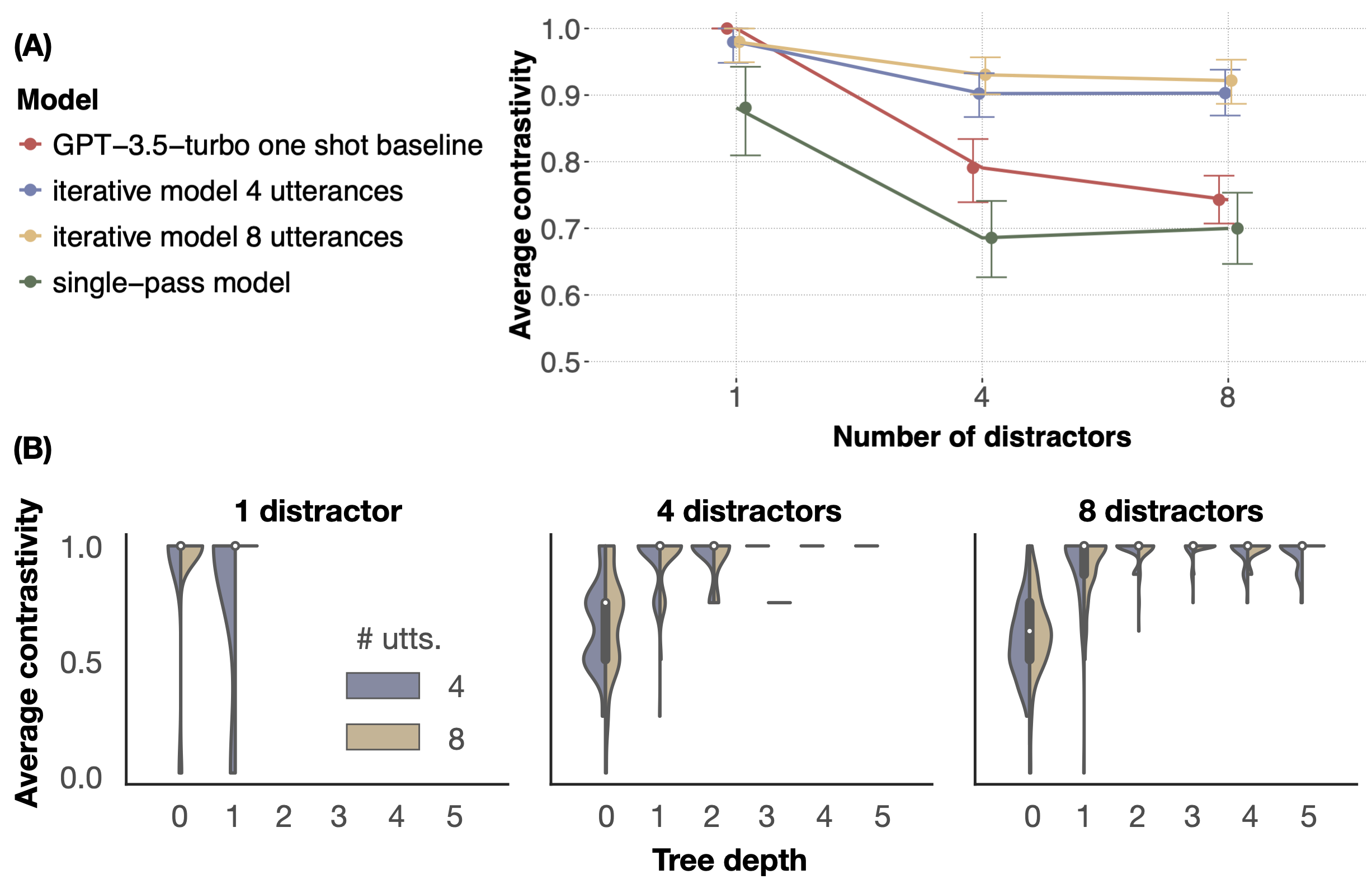}
    \caption{\textbf{A}: Reference game results: distribution over contrastivity values (y-axis) by number of distractors (x-axis) and number of utterances proposed (color). Error bars show bootstrapped 95\%-CIs.
     \textbf{B}: Distribution over contrastivity values (y-axis) over increasing tree depth in the iterated model (extended utterance proposal and evaluation iterations; x-axis), by number of distractors (facets) and tree width (number of proposed utterances; color).
     Dots indicate means, thick bars indicate quartiles, thinner lines indicate minimal values.
    \label{fig:ref-games-summary}}
\end{figure*}

\paragraph{Results.}
The average contrastivity across reference games in each configuration (i.e., for each number of distractors) for each of the three models is shown in Figure~\ref{fig:ref-games-summary}(A).
A bootstrapping analysis ($n=10000$) revealed that the performance of the single-pass model was worse than all other models across number of distractors ($P=1$).
Moreover, in contrast to the single-pass model, the iterative model generated highly contrastive utterances and the contrastivity of utterances produced by the IM remained high even with a larger number of distractors, outperforming the baseline and the single-pass model (see Fig.~\ref{fig:ref-games-summary}(A), 8 distractors).
Bootstrapping analyses confirmed that the average contrastivity of the iterative model for four and eight distractors was above the baseline ($P=1$).
For the condition with only one distractor, no significant difference was observed, possibly due to a ceiling effect.

To better understand the behavior of the IM, Figure~\ref{fig:ref-games-summary}(B) zooms in on the development of contrastivity over successive iterations.
For the conditions with 4 or 8 distractors, the contrastivity of the utterances seems to increase with increasing iterations, at least until hitting a ceiling.
In line with expectations, the number of iterations required in the IM until fully contrastive utterances were produced increased with the number of distractors and thus with the difficulty of identifying contrastive features.

Yet, there also seems to be some discrepancy between the human and the model's evaluations of contrastivity (shown in Figure~\ref{fig:ref-games-summary}(B)).
In the condition with four distractors, the models often had fully disambiguating expressions after one iteration (according to human evaluation), but still iterated further, which is only possible if the model itself did \textit{not} evaluate any utterance to be fully disambiguating.
This observation suggests that, indeed, semantic evaluation is a difficult aspect in this modeling approach (discussed in more detail below in Section~\ref{sec:cs1-discussion}).

Finally, manual inspection of the model output revealed that, when approaching the maximal number of iterations, a small fraction of the utterances were more literary and extensive than proposals from first iterations.
In few cases the utterances repeated the same features within the sentence, and sometimes the proposer came up with additional information or reformulations of the partial utterances from the previous model iteration (see Appendix~\ref{app:im-utt-proposer} for details).

\subsection{Summary \& conclusions}
\label{sec:cs1-discussion}
The key idea of SAGE is to combine an explicit task analysis with LM-based modules that generate a space of alternatives in an open-ended, context-relevant manner for transparent yet scalable models of pragmatic language use.
In this first case study, we used a well-studied example of pragmatic language generation, a reference game task, for a first assessment of whether a open-ended generation and semantic evaluation of referential descriptions can be integrated into an otherwise symbolic process, the Incremental Algorithm by \citet{dale1995computational}.
We compared the full SAGE model (iterative model IM) to an ablated single-pass (SP) model that produced utterances in a single pass without iterative refinement and found that the IM performed better under increasing context size.
Our conclusion from this is not to suggest SAGE as a general recipe for increased downstream task-performance.
Rather we consider these results suggestive evidence that LM-based generation and evaluation of open-ended alternative expressions can work (in the current, still restricted context) and that, in particular, the SAGE model operating on candidate utterances from a neural proposer seems to generate useful referential expressions that are adapted to the complexity of the context at hand.

In order to process alternatives in the iterative model, we used an LM-based semantic evaluator to assess the truth value of each utterance in each scene.
While we observed high-quality model outputs eventually, closer inspection of the model's results suggest that the semantic evaluation is a difficult task in itself, because it is essentially a case of logical inference based on natural language sentences.
For SAGE models to mature to the level of genuinely explanatory models, it would likely require more trustworthy, specialized, and reusable components for the kinds of semantic evaluation that are needed across a number of pragmatic pehnomena (e.g., in probabilistic models of pragmatic language generation; \citealp{degen2023rational}).
Fine-tuning or otherwise prepping LMs for this purpose would be one obvious solution to this; more powerful LMs as a backend can be explored, too.
Another possibility would be to depart from text-based representations of world states and consider multi-modal neural architectures, e.g., especially ones designed to evaluate semantic truth of utterances applied to images for SAGE-like applications.
We will come back to these issues in the final discussion in Section~\ref{section:discussion}.

\section{Case study II: M-implicatures}
\label{section:m-impl}

According to a wide-spread idea, listeners often infer what a speaker meant by an utterance by doing rationalization, abduction or inverse-modeling of the speaker \citep{grice1975logic,ParikhPrashGameImplicature1992,HobbsStickel1993:Interpretation-,BuntBlack2000:Abduction-Belie}.
The listener then needs to formulate an explanation that makes the speaker's utterance appear non-surprising, natural or predictable.
A particularly interesting case of rationalizing a speaker's utterance are so-called \emph{M-implicatures} \citep{levinson2000presumptive,Rett2020:Manner-implicat}, which take their name from Grice's Maxim of Manner, according to which speakers ought to be brief, clear, and orderly.
The utterance in (\ref{bsp:M-imp:marked}) is an example of an M-implicature. 
It invites the inference that the event was \textit{not} typical in some way, because the utterance uses the marked periphrastic causative ``cause to die'' \citep{BlutnerZeevat2004:Optimality-Theo}.
In contrast, the utterance in example (\ref{bsp:M-imp:unmarked}) is naturally interpreted as referring to an event typical for the lexical meaning of the causative verb \emph{to kill}.
This inference to stereotypicality is called \emph{I-implicature} in the linguistic literature \citep{levinson2000presumptive}.

\begin{exe}
  \ex \label{bsp:M-imp:unmarked}
  \begin{xlist}
    \ex \label{bsp:M-imp:unmarked-form}
    Black Bart killed the sheriff.
    \ex \label{bsp:M-imp:unmarked-interpretation}
    $\leadsto$ stereotypical killing, direct causation
    \hfill \textcolor{gray}{[I-implicature]}
  \end{xlist}
  \ex \label{bsp:M-imp:marked}
  \begin{xlist}
    \ex \label{bsp:M-imp:marked-form}
    Black Bart caused the sheriff to die.
    \ex \label{bsp:M-imp:marked-interpretation}
    $\leadsto$ non-stereotypical killing, indirect causation
    \hfill \textcolor{gray}{[M-implicature]}
  \end{xlist}
\end{exe}

M-implicatures are a particularly nice case study for SAGE modeling for a number of reasons.
First, they provide a well-defined example of inference by abductive reasoning, which requires identifying a plausible explanation for the observed behavior in a potentially open-ended space of possible reasons.
Second, it is challenging to explain how they arise.
A number of successful approaches to formalizing pragmatic reasoning struggled to model them, requiring innovative solutions such as bi-directional reasoning for optimality theory \citep{BlutnerBOTJoS2000}, strong belief in rationality in iterated best response models in game-theoretic pragmatics \citep{Franke2009:Signal-to-Act:-}, or inclusion of lexical uncertainty in probabilistic reasoning models \citep{BergenLevy2014:Pragmatic-Reaso}.
They have also been shown to pose a non-trivial task for language models \citep{cong2024manner}.
Finally, despite these technical challenges, prior work suggests a plausible algorithmic account of how M-implicatures might be arrived at during pragmatic interpretation.
One intuition is that M-implicatures from marked utterances like in (\ref{bsp:M-imp:marked}) occur when the interpreter realizes that a stereotypical interpretation could have been expressed with a form that is less marked, i.e., when the availability of a more economic way of expressing the same idea \emph{blocks} the most salient interpretation for a less common expression.
This idea is implicit in the algorithmic computation of bi-directional optimality \citep{Jager2002:Some-Notes-on-t}, but also in line with a number of prominent theories of pragmatic inference, such as the ``intentions-first'' approach suggested by \citet{Geurts2010:Quantity-Implic} and relevance theory 
\citep{SperberWilson1995:Relevance:-Comm}.
In the following, we spell out the \emph{markedness-blocking model}, which implements this general idea in a more open-ended fashion in a SAGE model by outsourcing judgements about complexity of a form and prior plausibility of a state to an LM.

\subsection{The markedness-blocking model for I- and M-implicatures}
\label{sec:cs2-model}



\begin{figure}[t!]
  \centering
  \includegraphics[width=\textwidth]{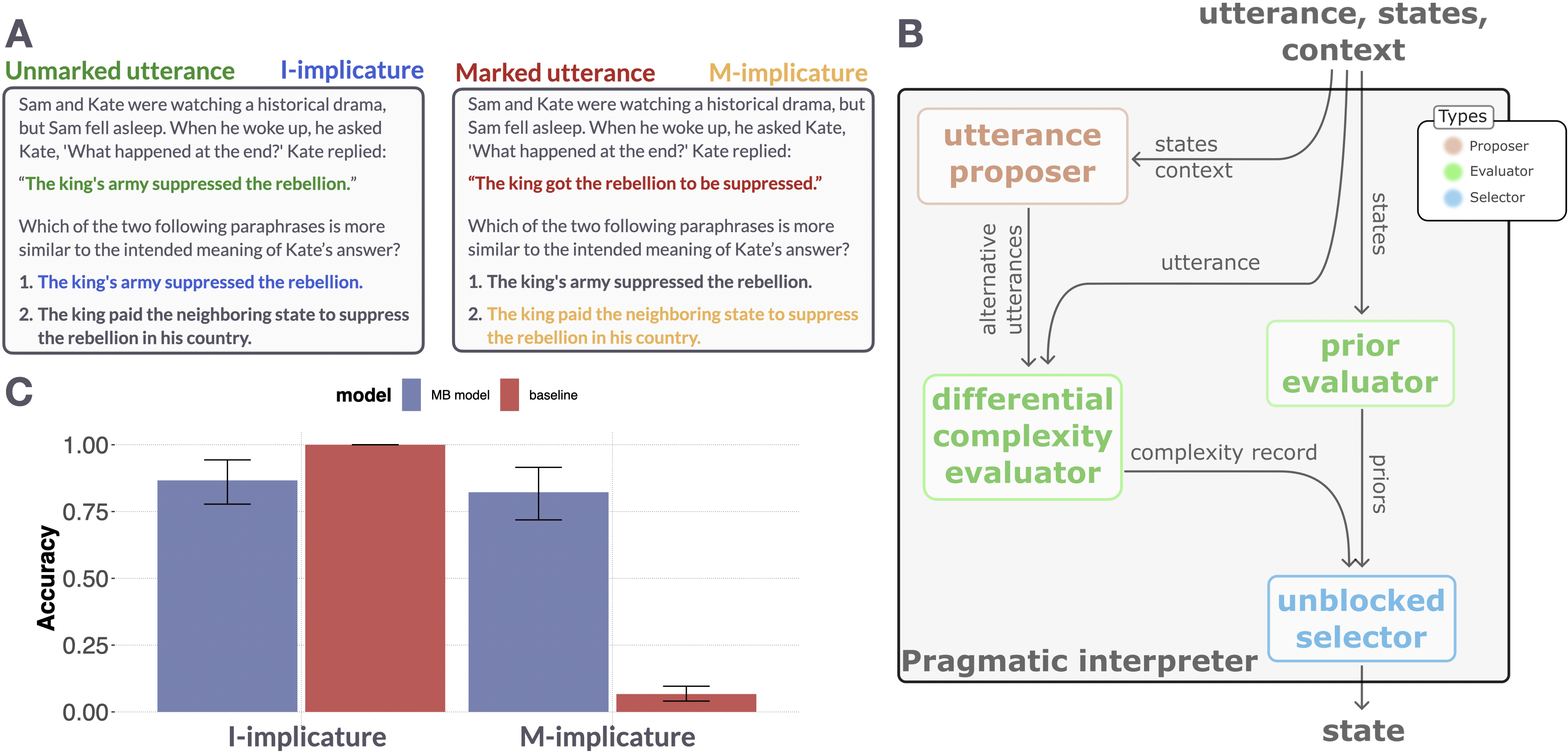}
  \caption{
    \textbf{A:}
    Example of an experimental item for testing I- and M-implicature inferences.
    For each vignette, an unmarked and a marked utterance were presented with two paraphrases describing situations the speaker might have intended to convey, a typical situation and an atypical situation.
    We expect the typical situation to be chosen for the unmarked utterance (the I-implicature).
    We expect the atypical situation to be chosen for the marked utterance (the M-implicature).
    \textbf{B:}
    Diagram of markedness-blocking (MB) model for the interpretation of I- and M-implicatures.
    The model takes as input a context, an utterance (marked or unmarked) and two possible states (typical and atypical).
    Based on the utterance, alternative utterances are proposed.
    The model compares the complexity of these alternatives to the observed utterance, and assesses the prior probabilities of the states.
    Finally, the model returns the most likely state for which there were no simpler alternative utterances than the observed utterance. 
    \textbf{C:}
     M-implicature inference (given the marked utterance) and I-implicature (given the unmarked utterance) results. Error bars show bootstrapped 95\%-CIs.
  }
  \label{fig:M-imp-combined}
\end{figure}

The markedness-blocking (MB) model formalizes the reasoning steps suggested by optimality-theory in the context of a concrete experimental task, while generalizing to accept open-ended linguistic input.
The task is modelled after previous work by \citet{wilson2016manner} and requires selecting a paraphrase for a speaker's utterance (see Figure~\ref{fig:M-imp-combined}A for an example).
The paraphrases are descriptions of a typical or an atypical situation, both of which are compatible with the speaker's unmarked or marked utterance.
We expect the paraphrase describing the typical situation to be selected after the unmarked utterance (I-implicature), and the the atypical situation to be selected after the marked utterance (M-implicature). 

The MB model takes as input a trigger utterance $u$ (marked or unmarked), a set $S$ containing two state descriptions (typical and atypical paraphrases), and a description of the context.
Its output is a selection of a state in $S$.
The model is pictured in Figure~\ref{fig:M-imp-combined}B and spelled out in pseudo-code in Algorithm~\ref{alg:flowSimpleMImplicatureInterpretation} in Appendix~\ref{sec:appendix-modules-MImplicatures}.
Based on the states $S$, the MB model first generates, using the \emph{UtteranceProposer} module, a number of alternative descriptions $u'$ for each state $s \in S$ as plausible alternatives a speaker would normally use to communicate $s$.
The model then checks if any of these $u'$ are less complex than the trigger utterance $u$.
If there are natural alternative descriptions $u'$ which are less complex than $u$, we say that $u$ is \emph{marked} for state $s$, and the pair $(u,s)$ is \textit{blocked}.
This is the task of the \emph{DifferentialComplexityEvaluator} module.
Having established markedness of $u$ for each state $s \in S$, the model also evaluates which $s \in S$ is more likely \emph{a priori} (without considering the trigger utterance) in the \emph{PriorEvaluator} module.
Finally, the \emph{UnblockedSelector} module selects the most likely \textit{unblocked} state as the model's output.

Rather than relying on modeler-specified alternatives and evaluation functions, the alternative utterances for each state $s \in S$, evaluations of their complexity compared to the trigger utterance $u$, and assessments of state priors were done via calls to LM-based modules with respective prompts, all of which can be found in the Appendix~\ref{sec:appendix-modules-MImplicatures}. 
The UtteranceProposer poses a challenging testbed for LM proposers, as the criteria for what constituted viable alternatives in this case study are more nuanced than in the previous case study: not only do the utterances have to truthfully describe the states, but, in order to reflect the theoretical assumptions about the structure of relevant alternatives, we expect them to be more complex given atypical states, and more natural-sounding given typical states.
For a comprehensive empirical assessment of the module's performance, we conduct human evaluations of this module's proposals, reporting full details of the procedure in Appendix~\ref{app:m-implicature-utterance-proposer}, and results in Section~\ref{section:cs2-summary}.

\subsection{Assessing the markedness-blocking model} \label{section:m-impl-experiment}

There are different types of M-implicatures, e.g., resulting from periphrasis, litotes, evaluativity, or plain verbosity \citep{Rett2020:Manner-implicat}.
Our computational experiments are based on \citeauthor{wilson2016manner}'s (\citeyear{wilson2016manner}) material, focusing on M-implicatures triggered by periphrastic causatives, such as \textit{to cause to die}.
Evaluation is based on accuracy, i.e., the ability to retrieve the ``correct'' implicature answer for a given utterance.
Accuracy of the MB model is also compared to a baseline model.

\paragraph{Materials.}
For this analysis we focus on M-implicatures from periphrastic causatives, returning to cases of verbosity in case study III in Section~\ref{sec:caseStudy-III}.
We created 15 vignettes inspired by the materials by \citet{wilson2016manner}.
Each vignette consisted of a context description, a pair of utterances (unmarked and marked) and a pair of state descriptions (typical and atypical), such as shown in Figure~\ref{fig:M-imp-combined}A.

\paragraph{Baseline.}
We compare the MB model to a zero-shot LM baseline that solves the task directly. 
The baseline uses the same LM as the modules of the MB model.
Concretely, the baseline model was prompted with the following instructions, completed with the respective information from each vignette:
\begin{prompt}
    \noindent Consider the following story:            \{context\} \\
           \{utterance\} \\
            Which of the two following paraphrases is more similar to the intended meaning of the answer? \\
            1. \{Atypical state\} \\
            2. \{Typical state\}
\end{prompt}
For each vignette, a second version of the prompt was tested where the interpretation (state) options appeared in reversed order. 
The reported accuracy results were calculated by averaging across the presentation order. 

\paragraph{Simulation procedure.}
We run the MB model four times on the entire set of vignettes, passing in the marked or unmarked utterance as the trigger, respectively.
We sampled $n=3$ alternative utterances with the UtteranceProposer in the MB model.
The baseline model was run for ten iterations over the dataset.

\paragraph{Results.}
For all retrieved predictions, we assessed the accuracy, i.e., whether the typical state was selected, given the unmarked utterance; and whether the atypical state was selected, given the marked utterance.
The average accuracy across runs, vignettes and order of presentation of answer options is displayed in Figure~\ref{fig:M-imp-combined}C.
For the simpler I-implicature case, the baseline model's accuracy is at ceiling, essentially always selecting the stereotypical state, while the MB model has high accuracy at about 80\%.
In the M-implicature condition, the baseline model almost always preferred the stereotypical interpretation, while the MB model was able to retrieve the ``correct'' M-implicature interpretation in about 80\% of the cases.



\subsection{Summary \& conclusions}
\label{section:cs2-summary}

Whereas the first case study looked at production, the second case study focused on interpretation of utterances.
It formulated and evaluated a model of pragmatic interpretation as rationalizing the speaker's utterance for M-implicatures from periphrastic causatives.
This still rather narrow case study further illuminates the potential of SAGE models, because M-implicatures have proven particularly difficult to formalize, and have spawned rather concrete algorithmic solution procedures.
The reported markedness-blocking (MB) model implemented a computational task analysis inspired by game-theoretic and optimality-theoretic approaches, where M-implicatures arise when rationalizing unexpected behavior.
We found that this model, but not a baseline model using only a simple prompt, was able to reliably select the M-implicature interpretation with high accuracy. 
The MB model crucially relies on three LM modules, all of which are interesting in the context of assessing the potential of the SAGE approach: proposing utterances, comparing the complexity of utterances, and evaluating the a priori probability of a state.
We scrutinize each component in turn.

To check the quality of the UtteranceProposer module, we ran human evaluations of the LM-generated proposals.
Humans rated how natural randomly selected proposals were for talking about typical and atypical states, compared to manually constructed marked and unmarked utterances for the respective vignettes.
We found that, across the different states, the proposer-generated alternatives were rated as equally natural as the reference utterance (i.e., the proposals were not credibly different from the unmarked utterance for an atypical state, and marked utterance for a typical state; $\beta =-4.26[-18.54,9.95]$, where here and in the following, we report posterior mean and 95\% credible intervals from Bayesian regression estimates).
The proposals were rated as credibly more natural than the non-matching reference utterances ($\beta=-23.75 [-35.66,-11.78]$). 
Details of the evaluation procedure, analyses and results can be found in Appendix~\ref{app:m-implicature-utterance-proposer}.
While not necessarily providing insights about absolute naturalness or quality of the proposals, these results confirm that the LM-based UtteranceProposer module generated open-ended viable alternative utterances for the purposes of the MB model in the context of our materials.

The MB model also used LM evaluator modules. One evaluator compared the complexity of the LM-sampled alternative utterances and the observed utterance; and another evaluator ranked the typicality of the states.
Predictions of the MB model crucially depend on the results of the utterance complexity evaluation.
If the module judges the alternative utterances to be simpler than the trigger, then the particular utterance-state pair will be blocked.
While the high-accuracy performance of the MB model on both I- and M-implicature indicates that the module performed functionally well, this module was rather sensitive to details of the prompt, which was optimized during development. 
The final prompt used in the simulations and details about the module are reported in Appendix~\ref{app:diff-complexity-eval}.

The observation that the performance of the complexity evaluator is not as stable as the performance of the proposer is in line with our results in case study I, which showed that the semantic evaluation also posed a challenging task for an LM module.
Therefore, developing more faithful methods for assessing the naturalness of utterances is vital to be able to use SAGE for robust modeling of pragmatic language interpretation.
One potential approach for future work could be to compare probabilities assigned by a language model to utterances, given states, rather than assessing them through prompting; recent work has suggested that this approach might arguably be more faithful to the capabilities of language models \citep[cf.][]{hu-levy-2023-prompting}.
We will return to this discussion in Section~\ref{section:discussion}.

\section{Case study III: Implicatures}
\label{sec:caseStudy-III}

The previous case study focused on a narrow but rich case of pragmatic interpretation, namely M-implicature from periphrasis.
The third case study, reported in this section, looks at pragmatic interpretation for a wider range of \textit{conversational implicatures} in a more general interpretation procedure, implementing a well-known but hitherto purely verbal algorithmic picture of Gricean pragmatics.
In the traditional Gricean picture, implicatures are inferences that listeners draw based on the assumption that speakers follow particular patterns of behavior, the so-called Maxims of Conversation: \emph{Quantity, Quality, Manner,} and \emph{Relevance} \citep{grice1975logic}.
Recognizably \emph{flouting}, i.e., intentionally not observing these maxims, can convey content beyond what is literally stated.
For instance, consider the cases in (\ref{bsp:flouting-examples}), where a character, Betsy, gives different answers to the same question, each flouting different common expectations about how such a question should be answered if the speaker were \textit{cooperative} and Gricean.
Different expectation violations, uncertain as they may be, give rise to potentially different \textit{inferences} about why Betsy behaved this way and what she wanted to convey.

\begin{exe}
  \ex \label{bsp:flouting-examples}
  \textbf{Context:} Betsy went to a talent show where your mutual friend Alex performed. You ask Betsy how the performance went, and Betsy replies:
    \begin{xlist}
        \ex Alex sang a song. \\
        \textcolor{gray}{too little information $\leadsto$ performance was probably dull}
        \ex Alex produced a series of sounds corresponding roughly to the score of an aria by Rigoletto. \\
        \textcolor{gray}{too verbose $\leadsto$ performance was probably poor}
        \ex Danny's performance was excellent. \\
        \textcolor{gray}{irrelevant $\leadsto$ speaker probably does not want to talk about it}
    \end{xlist}
\end{exe}

These inferences are frequently explained as the listener's attempt to rationalize the speaker's choice of utterance, given general assumptions about generic speakers' behavior (i.e., being cooperative and following the Maxims of Conversation) \citep[cf.][]{atlas1981clefts}.
However, a computational implementation of this picture is still lacking, since the recognition of flouting of Gricean maxims by the speaker depends, among other factors, on the context, access to alternative utterances, as well as the possibility to generate explanations of the speaker's behavior.

In the following, we formulate a SAGE model for implicature interpretation.
Since this case study has the broadest coverage and is intended to provide an exploratory computational account of very common non-literal language use, we present it in more detail and use it to stress-test the SAGE approach as a predictor of empirical data from human experiments.
We conduct model comparison using common tools from Bayesian data analysis between different SAGE models and human data, exploring how SAGE models can be scrutinized on par with traditional cognitive models.
Moreover, this case study also compares the effect of using different LMs as a backbone in a fixed SAGE model.
Next to using \texttt{GPT-3.5-turbo}  as the LM module backbone \citep{NEURIPS2020_1457c0d6}, we also use \texttt{GPT-4o} \citep[version 2024-08-06,][]{openai2023gpt4} and the open-source \texttt{Llama-3.1-8b-Instruct} \citep{dubey2024llama}.

\subsection{Gricean assumption-evaluation model of pragmatic inference}
\label{sec:models-CS-IIIa}

In the following, we detail a SAGE model for broad coverage of implicature interpretation which we call the \textit{assumption-evaluation (AE) model}.
The model attempts to rationalize the observed behavior of a speaker, starting from a set of standard Gricean assumptions. 
Specifically, the AE model computes pragmatic inferences by checking whether the speaker's utterance violates any of a number of given assumptions about speaker behavior.
For the most likely assumption that is flagged as potentially violated, the AE model produces possible explanations of why this violation may have occurred.

\begin{figure*}[t!]
\centering
\includegraphics[width=\linewidth]{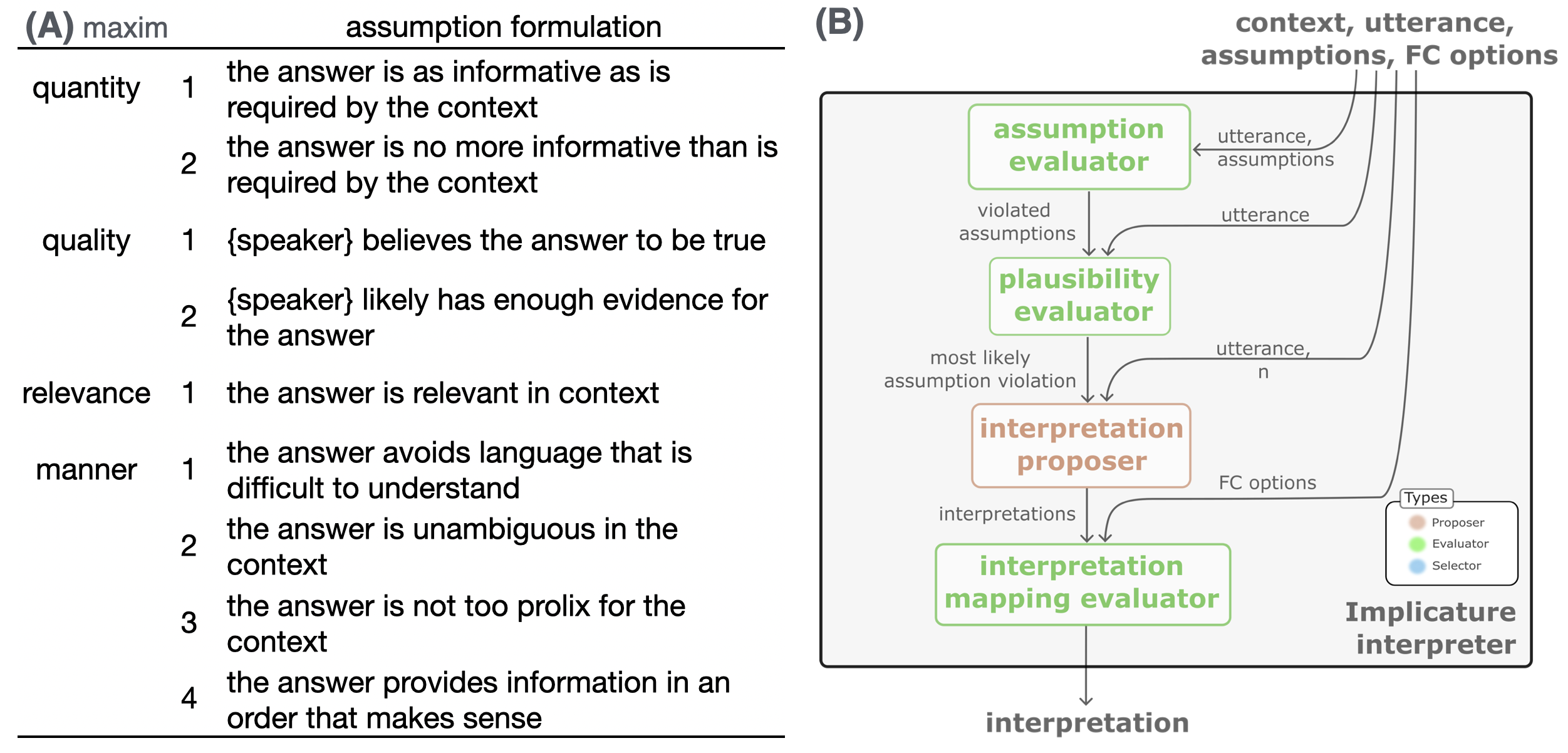}
\caption{\label{fig:interpretationAssumptionsFlow}
\textbf{A}: Gricean assumptions used in the interpretation algorithm for implicatures. \textbf{B}: Flow chart of the Assumption-Evaluation (AE) model for general pragmatic interpretation. The assumption evaluator checks whether each of a list of assumptions holds for the trigger utterance, after which an evaluator select the most likely among the violated assumptions, if the set of assumptions is non-empty. An interpretation proposer module proposes possible interpretations for a given utterance, which are then mapped onto the respective condition's forced choice (FC) options by an interpretation mapping evaluator, after which a majority reasoner selects the most voted interpretation.}
\end{figure*}

Different versions of the AE model of pragmatic interpretation are compared, based on different sets of conversational assumptions $A$ supplied to the model.
We particularly focus on the \textit{Gricean AE model}, which uses assumptions $A$ that directly correspond to the Maxims of Conversation originally proposed by \citet{grice1975logic}.
Like in Grice's original proposal, the maxims are broken down into more specific sub-maxims, which give more fine-grained principles to which the speaker's utterance could be expected to conform (Figure~\ref{fig:interpretationAssumptionsFlow}(A)).

The AE model is formulated to give predictions for items of a multiple-choice task like shown in the example vignette in Figure~\ref{fig:flouting-implicature-example-item}, discussed in more detail below.
The input to the model consists of a context, a trigger utterance $u$ to be interpreted, a set $O$ of four forced-choice options (including a target interpretation of $u$), and a set $A$ of assumptions about speaker behavior.
The model returns the best interpretation among the forced choice options.
While the possible output responses are defined in advance, as will become clear below, the reasoning process that motivates a particular choice is open-ended.


Figure~\ref{fig:interpretationAssumptionsFlow}(B) visually presents the AE model; 
more details are provided in Algorithm~\ref{alg:interpretationAssumptionsFlow} in Appendix~\ref{appendix:modulesCaseStudyIII}, which also lists all prompts and configurations of LM-based modules for this case study.
First, the \emph{AssumptionEvaluator} considers whether each assumption in $A$ holds of the observed utterance $u$ in the given context (e.g., iterating through each assumption in Figure~\ref{fig:interpretationAssumptionsFlow}(A)).\footnote{
If an assumption is flagged as violated, a corresponding pre-written sentence describing the violation is added to a list of violated assumptions.
For instance, if the assumption ``the answer is relevant in context'' is violated, we use the sentence ``the answer is not relevant in context'' in the interpretation proposer (see Tables~\ref{app:tab:cs3-gricean-assumptions}--\ref{app:tab:cs3-lexical-assumptions} in the Appendix for the full list of sentences used for each assumption).}
The output of the AssumptionEvaluator module is a (possibly empty) subset $A_{v} \subseteq A$ of assumptions violated by the utterance $u$.\footnote{
Though not explicitly represented in Figure~\ref{fig:interpretationAssumptionsFlow}(B), the AssumptionEvaluator module also includes the markedness check of the markedness-blocking model from the previous Section~\ref{section:m-impl} (see Figure~\ref{fig:M-imp-combined}(B)).
Concretely, we check whether the utterance $u$ is marked given the target and competitor options as state descriptions.
If the algorithm flags utterance $u$ and target option as unblocked, while utterance $u$ and competitor option as blocked, the following sentence is added to the list of violated assumptions: ``\{speaker\}'s utterance is odd because there is a more usual formulation for \{utterance\}'' (where ``\{speaker\}'' and ``\{utterance\}'' are replaced appropriately for the vignette).
}
Next, the \emph{PlausibilityEvaluator} identifies which assumption violation $a \in A_{v}$ is the most plausible violated assumption from the set $A_{v}$.
This evaluator operated on log probabilities of strings assigned by an LM rather than based on prompting, an alternative approach to leveraging LMs discussed in more detail in Section~\ref{sec:discussion-solutions}.
Specifically, the $a$ was identified via computing a measure of \emph{prior-adjusted conditional probability} of the assumption $a$, given $u$, which is sometimes referred to as \emph{empirical mutual information} and which has been shown to yield superior results for scoring alternatives in benchmark testing in the literature \citep{PoliakNaradowsky2018:Hypothesis-Only,BommasaniHudson2021:On-the-opportun,HoltzmanWest2021:Surface-Form-Co}:\footnote{
  Since this evaluation requires access to conditional probabilities of expressions under the language model, the open-source \texttt{Llama-3.1-8B} model was used for this module for both the \texttt{Llama-3.1-8B} based experiments and the \texttt{GPT-4o} based experiments, since scoring of expressions was not available through the OpenAI API at the time of availability of GPT-4o.
  The model \texttt{text-davinci-003} was used for the \texttt{GPT-3.5-turbo} based experiments.
}
$$ 
\text{plausibility}(u, a) = \log \frac{P(a \mid u)}{P(a)} 
$$

Finally, the assumption $a^{*}$ with the highest plausibility score is used in the \emph{InterpretationProposer} to generate $n$ possible interpretation options $I = \{i_{1}, \dots, i_{n}\}$.
Then, the \emph{InterpretationMapping} module identifies the most likely option among the forced choice options $O$, given the interpretations $I$, through majority voting.
Specifically, for each generated interpretation $i_{j}$, the module selects the most fitting option $o \in O$ that should be chosen if the interpretation was $i_{j}$.
The final prediction of the algorithm is the most frequently selected option $o$ (with random tie breaking) over all $i \in I$.\footnote{
From a more technical point of view, this interpretation mapping architecture for forced-choice tasks is related to work proposing various structured prompting strategies \citep[e.g.,][]{yao2023tree}.
Our implementation closely resembles the work by \citet{liu-etal-2022-generated}, and is similar to various uses of LMs as judges in the literature \citep{bavaresco-etal-2025-llms}.
} 

\subsubsection{Assessing the AE model} 
\label{section:assessment-CS-III}

We test the AE model based on two datasets.
The first dataset was specifically created for this investigation, while the second dataset is taken from prior research \citep{hu-etal-2023-fine}.
For both datasets, we evaluate models based on two criteria:
(i) their accuracy, i.e., the ability to select the intended pragmatic or literal interpretation, and
(ii) their quantitative fit to human data.
Additional critical analyses of the performance of the LM-based components of the AE model, i.e., the assumption evaluator and the interpretation proposer modules, are reported in detail in Appendix~\ref{app:assumption-eval}--\ref{app:implicatures-interpretation-proposer}.

\begin{figure*}[t]
  \centering

  \includegraphics[width=0.8\textwidth]{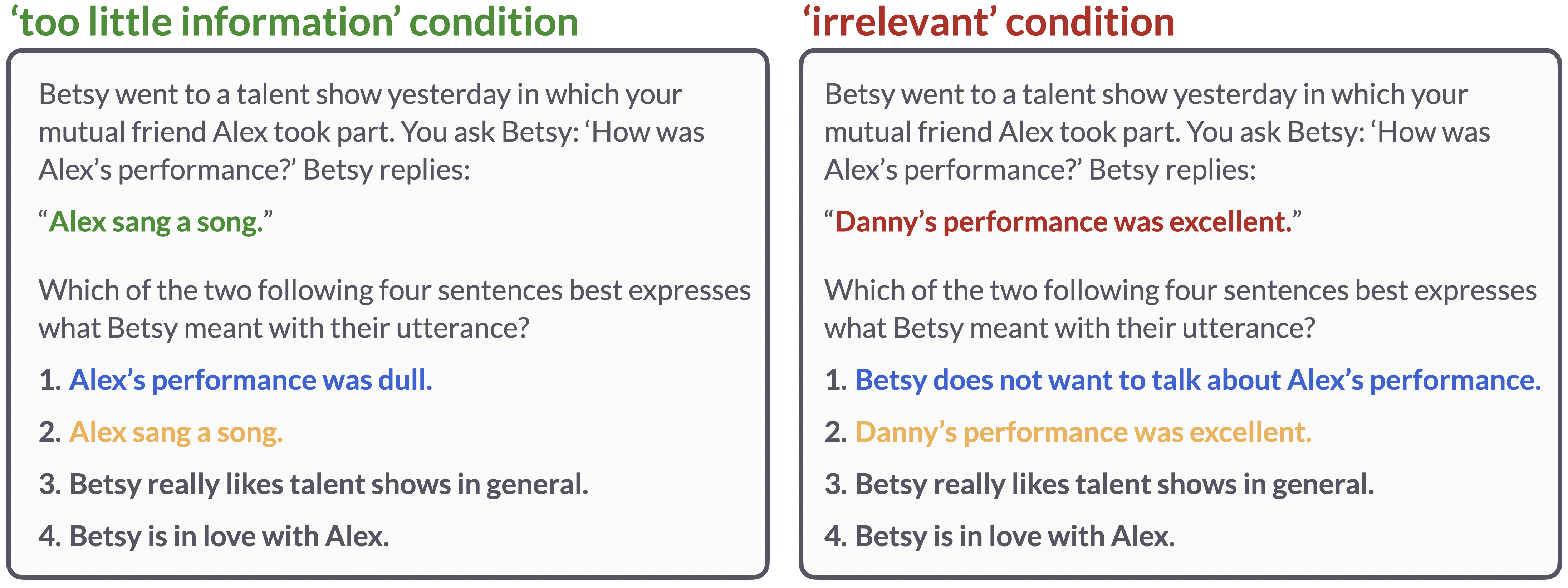}

  \caption{
    Examples of an experimental item for testing implicatures possibly arising from violations of different expectations about the speaker's behavior.
    The examples are for the `too little' and `irrelevant' trigger conditions.
    The target answer, expected from pragmatic reasoning, is listed first.
    The second answer is the competitor option, a literal interpretation without additional pragmatic inference.
    The other two answer options are distractors.
    In experimental trials with humans and LMs the order of options was randomized.
  }
  \label{fig:flouting-implicature-example-item}
\end{figure*}

\paragraph{Materials.}
\label{section:impl-materials}
Materials of the first data set are comprised of 24 novel vignettes.
Each vignette consists of a context description and five trigger sentences: a baseline trigger, which is designed so as not to violate any maxims or assumptions, and one trigger sentence for each of the following conditions, resulting in 120 unique context-trigger pairs:
\begin{enumerate}
    \item Irrelevance\\
    \textcolor{gray}{The utterance provides information that is irrelevant for the purposes of the conversation.}
    \item Markedness\\ 
    \textcolor{gray}{The utterance is unexpectedly complex for describing a typical state of affairs.}
    \item Too little information\\
    \textcolor{gray}{The utterance provides less information than is required for the purposes of the conversation.}
    \item Too much information\\
    \textcolor{gray}{The utterance provides more information than needed.}
\end{enumerate}
The `markedness' condition uses periphrastic causatives, similar to the items from case study II.
The `too much information' condition also taps into M-implicatures, but here triggered by excess verbosity or otherwise marked formulations.
Figure~\ref{fig:flouting-implicature-example-item} shows example trials for two conditions and Table~\ref{table:implicatureVignetteExample} contains example material for one vignette in all conditions.

For each context-trigger pair, four possible interpretation options are constructed: the \emph{target} interpretation of the sentence, a literal \emph{competitor} interpretation, and two infelicitous \emph{distractor} interpretations. 
For the baseline trigger, the competitor interpretation is a false statement.
The distractors are shared across the five conditions.
In the experiments, participants (humans or models) chose the best interpretation option for the speaker's intended meaning.
These interpretation options were embedded into one of the following randomly sampled matrix clauses: ``\{speaker\} \{tries, wants\} to \{communicate, convey, express\} that \{interpretation\}'', to balance out 
possible effects 
of any particular matrix clause.

\begin{table}
\centering
\begin{tabular}{p{1.5cm}p{4cm}p{4cm}p{4cm}}
\toprule
condition &
trigger sentence &
target & 
competitor 
\\ \midrule
too little &
\footnotesize{Alex sang a song.}&
\footnotesize{Alex’s performance was dull.}&
\footnotesize{Alex sang a song.}
\\\addlinespace[0.3em]
too much &
\footnotesize{Alex produced a series of sounds corresponding roughly to the score of an aria from Rigoletto.} &
\footnotesize{Alex’s performance was not very good.} &
\footnotesize{Alex produced a series of sounds corresponding roughly to the score of an aria from Rigoletto.}
\\\addlinespace[0.3em]
irrelevant &
\footnotesize{Danny’s performance was excellent.}&
\footnotesize{Betsy does not want to talk about Alex’s performance.}&
\footnotesize{Danny’s performance was excellent.}
\\\addlinespace[0.3em]
marked &
\footnotesize{Alex caused the audience to laugh.}&
\footnotesize{Alex made the audience laugh unintentionally, e.g., by mistake.}&
\footnotesize{Alex made the audience laugh intentionally e.g., by telling a joke.}
\\\addlinespace[0.3em]
literal baseline &
\footnotesize{Alex performed really well.}&
\footnotesize{Alex performed really well.}&
\footnotesize{Alex didn't take part in the talent show.}
\\\bottomrule
\end{tabular}
\caption{
Example of a vignette in the five conditions. The vignette has the following context: ``Betsy went to a talent show yesterday in which your mutual friend Alex took part. You ask Betsy: `How was Alex’s performance?' Betsy replies: `\{trigger sentence\}' ''.
The vignettes also included two distractor interpretation options, which were shared across all conditions. Distractor 1: ``Betsy really likes talent shows in general.'' Distractor 2: ``Betsy is in love with Alex.''
\label{table:implicatureVignetteExample}
}
\end{table}

In order to compare our models to previous work, we additionally used materials of the ``Maxims'' condition in \citet{hu-etal-2023-fine}.
This dataset has a similar structure and comprises 20 vignettes (five per Gricean maxim).
Each vignette consists of a context and trigger sentence which gives rise to a non-literal interpretation, based on flouting of one of the maxims.
Each vignette is shipped with four interpretation options: correct non-literal interpretation (i.e., target), incorrect literal interpretation (i.e., competitor), incorrect non-literal and incorrect associate interpretations (i.e., two distractors).

\paragraph{Baseline.}\label{section:impl-baseline}
Similarly to previous case studies (Sections~\ref{section:utterance-production}--\ref{section:m-impl}), we compare the AE model to a baseline. In the baseline, the task is solved by the same LM which is used for the modules' backbone, but with a single zero-shot call. We label this baseline the single-pass (SP) model.
The LM is prompted with the following instructions, completed with the respective information from each vignette (Section~\ref{section:impl-materials}):
\begin{verbatim}
  Context:
  {context} "{trigger sentence}"
  Which of the following sentences best expresses what {speaker name} meant 
  with their utterance?
    1. {competitor}   
    2. {target}
    3. {distractor 1} 
    4. {distractor 2}
\end{verbatim}
The order of interpretation options was randomized for each sample. 
The prediction accuracy was determined by matching the returned generation against the forced choice options, complemented with manual categorization when automatic matching failed.

\paragraph{Human data.}
For the novel materials in dataset 1, we ran an interpretation experiment on the crowdsourcing platform Prolific and collected data from 285 participants.
Native speakers of English based in the US with at least a 95\% approval rate and at least five previously completed studies were recruited.
The study was a forced-choice experiment in which participants first read instructions, and then completed two trials.
On each trial, they read the context with the trigger sentence presented below and selected one of four options, presented in randomized order (see example in  Figure~\ref{fig:flouting-implicature-example-item}).
For the critical trials, two vignettes with different trigger conditions were sampled at random.
The study took around two minutes and participants were reimbursed with \pounds 0.30.\footnote{The code and data can be found under \url{https://github.com/magpie-ea/magpie3-interpretations-fc}. 
}
Detailed analyses of the human results for dataset 1 are reported in Appendix~\ref{app:cs3-human-expt}. The data for dataset 2 from a forced-choice experiment with human participants has been made available by \citet{hu-etal-2023-fine}, and we use it here.

\paragraph{Simulation procedure.}

We first compare the Gricean AE model to the SP baseline.
We ran 20 iterations for each model on each item with \texttt{GPT-3.5-turbo} as the module backbone and baseline; we ran ten simulations per model when using \texttt{GPT-4o} and \texttt{Llama-3.1-8b-Instruct}. Details about the backbones are reported in Appendix~\ref{app:sec:lm-configs}.
We sample $n=4$ interpretations for the InterpretationProposer for all AE model simulations in this case study. 
We use temperature $\tau=0.1$ for sampling from GPT-based modules and the baseline, and $\tau=0.8$ for sampling from \texttt{Llama-3.1-8b-Instruct}. 
Additionally, for \texttt{Llama-3.1-8b-Instruct} based modules, the maximal number of predicted tokens is set to 128, the repetition penalty to 1.8 and top $P = 0.9$.
 Here, we report results aggregated across backbones, and report by-backbone results in Appendix~\ref{app:cs3-accuracy}--\ref{app:cs3-human-llh}.

\paragraph{Accuracy-based assessment.}
\begin{figure*}[t!]
  \centering
  \includegraphics[scale = 0.3]{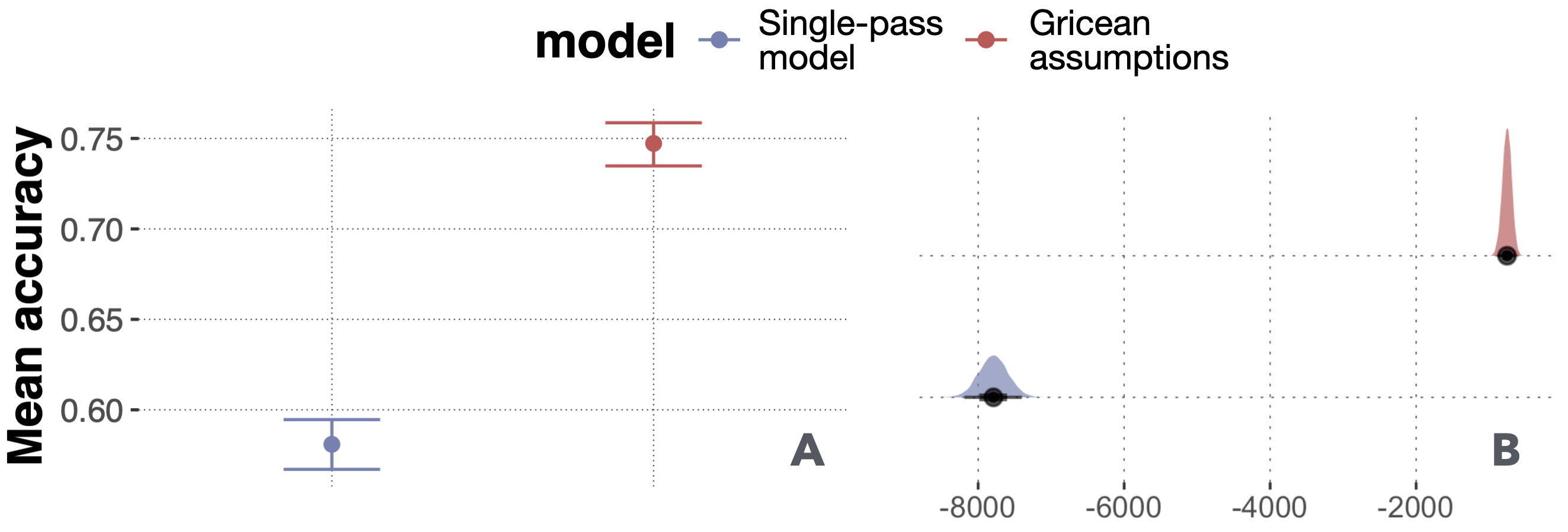}
  \caption{
    Both plots show results across backbones and datasets. \textbf{A}: The average accuracy of selecting the target interpretation (y-axis) for the single-pass LM baseline and the Gricean AE model (x-axis; chance is 25\%). \textbf{B}: Density plots and summary statistics for the posterior distribution of the total log-likelihood of the human data of the single-pass LM baseline and the Gricean AE model (x-axis).
    Black dots indicate means of posterior samples for the total log-likelihood, thick black lines indicate 66\% credible intervals, thinner black lines indicate 95\% credible intervals.
    \label{fig:cs3-SP-Gricean-analyses}
  }
\end{figure*}
Figure~\ref{fig:cs3-SP-Gricean-analyses} (A) reports the accuracy of the single-pass LM baseline and the Gricean AE model, aggregating across the different LM backbones, the two datasets and trigger conditions therein.
Results broken down by LM backbone are reported in Figure~\ref{fig:app-cs3-avg-acc-byBackbone} in the Appendix.
We fit a logistic Bayesian regression model to test whether the Gricean AE model results in higher accuracy than the single-pass LM baseline.
We find that the AE model is credibly more accurate than the single-pass baseline, across datasets and LM backbones ($\beta=0.76\; [0.68, 0.85]$, posterior mean and 95\% credible interval).
That is, at least in terms of predictive accuracy, our task decomposition based on assumption evaluation outperforms the SP baseline.
More specifically, the decomposition helped performance for weaker LM backbones (\texttt{GPT-3.5-turbo} and \texttt{Llama-3.1-8b-Inst}; detailed statistical comparisons are reported in Appendix \ref{app:cs3-accuracy}).
computational cognitive modeling to further scrutinize the results.


\paragraph{Model comparison based on likelihood of human data.}
The previous assessment focused on accuracy in identifying the target interpretation. 
Next, we compare models based on their ability to predict the distribution of answer types (target, competitor, distractor) in the human data.
A model is deemed better than another if it entails a distribution over answer options that is more human-like.

As defined currently, the SAGE models do not give us an explicit likelihood function over answer choices, but only produce single-example samples.
But we can approximate the underlying answer distribution from a set of single-example samples from each model, and thereby obtain a probability distribution (capturing our uncertainty due to using samples) over the model's total likelihood for all human data points (see Appendix~\ref{sec:assess-likel-funct} for formal details of this procedure).
Figure~\ref{fig:cs3-SP-Gricean-analyses} (B) shows summary statistics (means and 95\% credible intervals) and the density for the distribution of total likelihood across both datasets, aggregated across LM backbones.
(Results separated by LM backbone are shown in Appendix~\ref{app:cs3-human-llh} in Figure~\ref{fig:model-comparison}.)
Visually, the baseline SP model is clearly a worse predictor of the distribution of human answer choices than the Gricean AE model, since the log likelihood of human data is by orders of magnitude lower under the SP model that the AE model, at least across LM backbones.
Similarly to the accuracy-based results, this difference holds for the models based on the weaker LM backbones (\texttt{GPT-3.5-turbo} and \texttt{Llama-3.1-8b-Inst}, see Figure~\ref{fig:model-comparison} in the Appendix).
This confirms that the algorithmic analysis inspired by the Gricean picture of implicature interpretation not only generally captured the capability to accurately interpret implicatures (which focused on the rate of target responses), but, in principle, could also lead to an overall more human-like response pattern than zero-shot predictions.

\subsection{Alternative assumptions models} 
\label{sec:models-CS-IIIb}

Our analyses of the AE model have so far not provided deeper insights into the performance of the most relevant, novel modules in architecture --- the interpretation proposer and assumption evaluator.
Especially the assumption evaluation task might pose a challenge both for the LM and for assessment of the module performance itself, since multiple maxims may be flouted by a single utterance.
We therefore assess the robustness of the LM assumption evaluator by comparing alternative models, obtained by using different sets of assumptions about the speaker's behavior.

First, the \emph{intuitive assumptions model} considers a set of \emph{intuitive} assumptions about a cooperative speaker, similar to the Gricean maxims, but here phrased in a way that we deemed more intuitive, supporting reasoning specifically for trigger conditions that we wanted to test (see the ``intuitive assumptions'' column of Table~\ref{table:intuitiveNonsenseAssumptions}).
Second, we consider the \emph{lexical assumptions model} based on a set of \emph{lexical} assumptions, which formulate facts about the speaker that are generally irrelevant to the evaluation task but mention a keyword reflecting the Gricean maxims  (``lexical assumptions'' column of Table~\ref{table:intuitiveNonsenseAssumptions}).
Finally, we consider a \emph{no assumptions (baseline) model} with an \emph{empty} set of assumptions.
This model operates without evaluating any assumptions and relies solely on the part of the model sampling $n$ different interpretation proposals, serving to identify the best fitting forced choice interpretation option, without additional information about which assumptions may have been violated.

\begin{table}[t]
  \centering
  \begin{tabular}{p{3.0cm}p{5.25cm}p{5.25cm}}
    \toprule
    name
    & intuitive assumptions
    & lexical assumptions
    \\ \midrule
    evidence
    &\footnotesize{the answer is adequately supported by evidence}
    &\footnotesize{the answer provides \textit{evidence} that Paris is in France}
    \\
    \addlinespace[0.5em]
    naturalness
    &\footnotesize{the answer is a precise and natural way of expression}
    &\footnotesize{the answer is a \textit{natural} reaction to being threatened with a fork}
    \\
    \addlinespace[0.5em]
    objectivity
    &\footnotesize{the answer is only conveying objective information about the world; it does not convey subjective information like \{speaker\}'s evaluation, judgement or opinion}
    &\footnotesize{the answer is only conveying \textit{objective information} about the world of The Lord of the Rings}
    \\
    \addlinespace[0.5em]
    relevance
    &\footnotesize{the answer only provides information that is relevant for the addressee in the given context}
    &\footnotesize{the answer is \textit{relevant} to building a house}
    \\
    \addlinespace[0.5em]
    inform. satisfaction
    &\footnotesize{the answer conveys sufficient information to answer the question of the addressee in the given context}
    &\footnotesize{the answer conveys \textit{sufficient information} to find the way to the nearest cafe}
    \\
    \addlinespace[0.5em]
    inform. granularity
    &\footnotesize{the answer conveys no more information than can be normally expected in the given context}
    &\footnotesize{the answer conveys \textit{no more information} than can be expected given a sleepy speaker}
    \\
    \bottomrule
  \end{tabular}
  \caption[IntuitiveNonsenseAssumption]{
    Intuitive and lexical assumptions used in the interpretation algorithm for implicatures. The lexical assumptions embed a keyword (italicized) related to Gricean maxims in an otherwise nonsensical sentence with respect to the task at hand.
    \label{table:intuitiveNonsenseAssumptions}
  }
\end{table}

By comparing the Gricean AE model to models with the other assumption sets, we assess the robustness of the assumption evaluation module to variations in prompting through the lens of overall model performance.
By comparing the full AE model and the no-assumptions baseline, we can most stringently test how much added benefit there is in explicitly adding the LM assumption evaluator module at least in its current implementation, relative to adding the interpretation proposer.

\subsubsection{Assessing the models}

\begin{figure*}[t]
  \centering
  \includegraphics[width=0.9\textwidth]{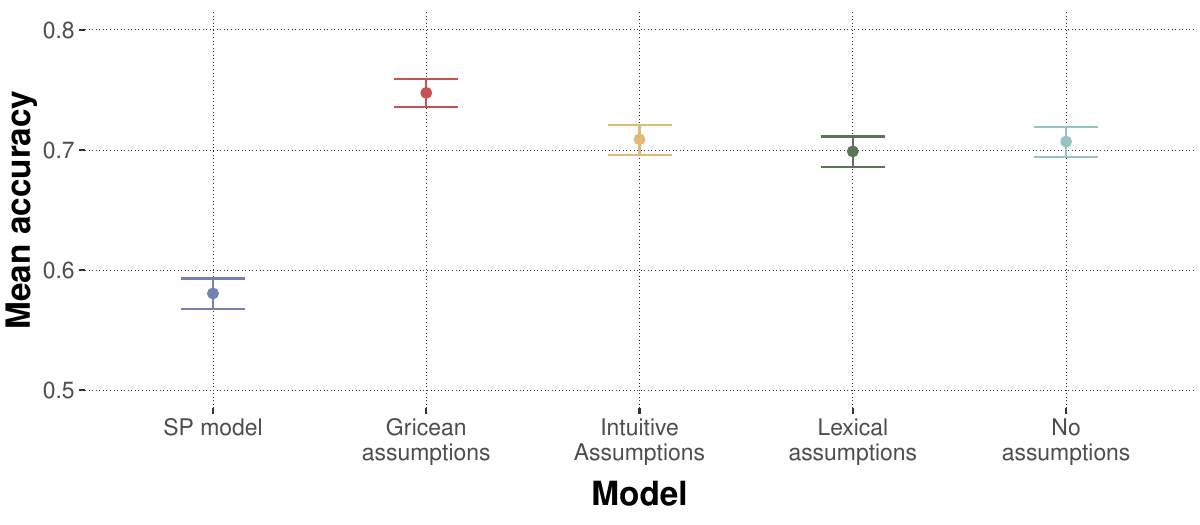}
  \caption{
    Interpretation results in case study III. The average accuracy (i.e., proportion of correct selections of the intended interpretation option, given the trigger utterance in context) is shown across LM backbones, datasets and simulations (y-axis), by model (x-axis). Chance accuracy is 25\%. The error bars indicate 95-\% bootstrapped CIs. SP and Gricean AE model results are replicated from Figure~\ref{fig:cs3-SP-Gricean-analyses} (A) to facilitate visual comparison.
    \label{fig:cs3-acc-summary}}
\end{figure*}
\begin{figure*}[t]
  \centering
  \includegraphics[width=0.7\textwidth]{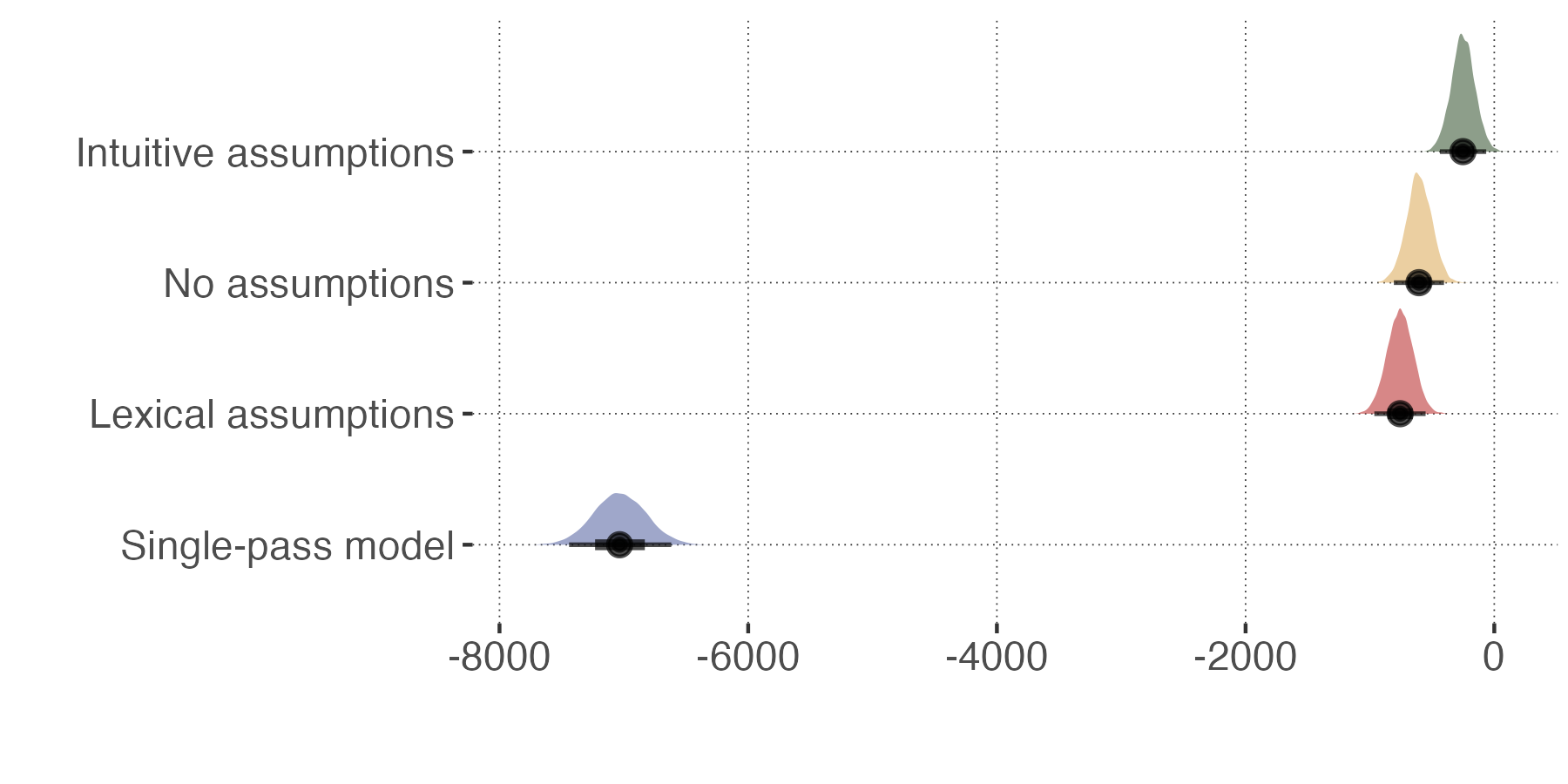}
  \caption{
    Density plots and summary statistics for the posterior distribution of differences in total \textbf{log-likelihood of the human data} between the Gricean-assumptions model and any other model (lines of the plot), across LM backbones and datasets.
    Black dots indicate means of posterior samples for the total log-likelihood, thick black lines indicate 66\% credible intervals, thinner black lines indicate 95\% credible intervals.}
  \label{fig:cs3-model-comparison}
\end{figure*}

We investigate next whether (1) the AE task decomposition provides an improvement over the single-pass (SP) baseline with different assumption sets, and whether (2) the formulation of the assumptions matters, i.e., whether their performance is distinguishable from the Gricean AE model.

\paragraph{Accuracy-based assessment.}
The results comparing the average accuracy across datasets, trigger conditions, and LM backbones are shown in Figure~\ref{fig:cs3-acc-summary}. 
Compared to the SP baseline, all alternative models (i.e., with intuitive, lexical or no assumptions) achieve higher accuracy. 
To address question (1) quantitatively, we use the same Bayesian logistic regression model as for investigating the effect of the Gricean AE model.\footnote{Model in R syntax: \texttt{response\_correctness $\sim$ assumptions\_set + dataset + LM\_backbone}.}
Specifically, we compare each of these assumption sets to the SP baseline and find that they are all credibly more accurate than all other models ($\beta_{intuit}=0.57 \; [0.48, 0.65], \beta_{no}=0.56 \; [0.47, 0.64], \beta_{lex}=0.52 \; [0.44, 0.60]$). 
Further, we compare each of the AE models based on intuitive, lexical and no assumptions to the Gricean AE model, and find that the Gricean model was credibly better than each of the alternative models (posterior estimates of the contrasts: $\beta_{G-int}=0.19 \; [0.10, 0.28], \beta_{G-no}=0.20 \; [0.12, 0.29], \beta_{G-lex} = 0.25 \; [0.16, 0.33]$).

Based on these results, we conclude that our task decomposition generally leads to improved performance over a single-pass monolithic prediction with an LM, and on average, does so robustly over different formulations of the assumptions. 
More detailed analyses in Appendix~\ref{app:cs3-accuracy} suggest that the AE model structure mostly improves performance across assumption sets for \texttt{GPT-3.5-turbo} and \texttt{Llama-3.1-8b-Inst}, but not \texttt{GPT-4o}.
On the other hand, only the larger GPT-based backbones were sensitive to the assumptions being more Gricean (i.e., the Gricean or intuitive rather than lexical $A$). 
That is, the robustenss of the predictive improvement for the AE model may vary depending on the LM backbone, an observation we return to in Section~\ref{sec:discussion-solutions}.
Additionally, the no-assumptions model also provides a substantial improvement over the SP baseline, suggesting that an intermediate level of task decomposition, when using the interpretation proposer only, also leads to performance improvements.

\paragraph{Model comparison based on likelihood of human data.}
We also calculate the likelihood of human data under predictors from models with intuitive, lexical or no assumptions, across LM backbones.
We calculate the differences between the total log likelihood of human data predicted by the Gricean AE model and the total log likelihood predicted by each of the alternative models.
The posterior distributions and respective summary statistics are shown in Figure~\ref{fig:cs3-model-comparison}.
Across datasets and backbones, the AE models based on other sets of assumptions fit human data credibly worse than the Gricean AE model (Figure~\ref{fig:cs3-model-comparison}, credible intervals do not contain 0), although the differences between intuitive and no assumptions didn't appear very large.
More detailed analyses by LM backbone are reported in Appendix \ref{app:cs3-human-llh} in Figure~\ref{fig:model-comparison}.
These additional results suggest that the intuitive AE model is closest to the Gricean AE model in terms of fit to human data across backbones, while the fit is worse for other assumption sets for the more performant GPT-based backbones, but not the small \texttt{Llama-3.1-8b-Inst}~model.

In sum, these results suggest that the implicature interpretation analysis implemented here through SAGE is at least in part a viable step towards converting a verbal account of the cognitive process underlying implicature interpretation into a computational model, as it predicts human-like interpretation patterns and mostly outperforms the SP baseline; and that the task analysis is robust to variations in the formulations of the speaker-focused assumptions. 
The no-assumption model ablation suggests that the improvement of performance and fit to human data over the SP baseline might be carried by either of the neural modules we introduce for implementing the Gricean implicature interpretation account, the AssumptionEvaluator or the InterpretationProposer.
Importantly, this aspect of the current model architecture became apparent through model comparison with respect to its explanatory fit to human data, highlighting the importance of relying on tools from computational cognitive science for critical model comparison.
 
\subsection{Summary \& conclusions}
\label{section:cs3-discussion}
Case study III is the most ambitious model in this paper, focus on analyzing implicature interpretation as abductive rationalization of the speaker's behavior from Gricean conversational maxims.
We generalized the previous case study beyond the narrow domain of M-implicatures, to a rationalization process based also on the Maxims of Quantity, Quality and Relevance.
We report the Gricean assumption evaluation (AE) model that identifies which maxims, if any, are likely flouted by the observed utterance, and generates possible interpretations resulting from the flouting.
We compare the Gricean assumptions to two other assumption sets, as well as to an ablation with an empty assumption set and a single-pass (SP) model baseline.
We run these comparisons for two multiple-choice datasets: a novel dataset manipulating the type of implicature within-vignette, and a dataset from \citet{hu-etal-2023-fine}.
Additionally, we test the AE models with different LMs backbones.
We evaluate the different AE models not only in terms of their accuracy of identifying the correct pragmatic interpretation of the utterance, but also in terms of their explanatory fit to human data collected on the same datasets.

Across different LM backbones and datasets, the Gricean AE model outperformed both the SP model and the other AE models.
Additionally, the Gricean AE model fit human data credibly better than the single-pass LM baseline. 
Yet analyses of AE models based on different LM backbones revealed that AE models based on Gricean and intuitive assumptions showed a similar fit to human data; for \texttt{Llama-3.1-8b-Inst} the Gricean and the no assumptions models showed a similar result, while the fit of the \texttt{GPT-4o} Gricean model was also similar to the SP baseline.
While these results suggest a certain degree of robustness against prompt variation of the LM-based modules, they also indicate variability of the model performance for different LM backbones and raise the question of the explanatory fit and contribution of the single modules to the AE model's overall predictive performance.
 
To address these questions, we scrutinize the two novel neural modules introduced in this model:
(i) the Gricean assumption evaluator assessing different kinds of input utterances, and
(ii) the interpretation proposer module for constructing open-ended interpretations of utterances.
We critically evaluate the performance of these two modules in experiments with human participants, reported here in turn.

First, to assess the assumption evaluator module, we asked participants to evaluate whether assumptions from the Gricean set were violated by trigger utterances in vignettes sampled from dataset 1 (see Appendix~\ref{app:assumption-eval} for more information).
We compared the proportions of violations of different maxims identified for different trigger utterance conditions by humans to the proportions predicted by the LM evaluator using the \texttt{GTP-3.5-turbo} backend.
We found that, across trigger conditions, humans were more reluctant to flag maxims as being violated, while the evaluator flagged almost all maxims as violated.
We also found that different experimental conditions triggered very similar reports of violated assumptions.
This was pronounced for the tested LM-based evaluator but to a certain extent also for humans.
Neither of these observations need be a problem for our SAGE model architecture, as the subsequent plausibility evaluator module could hone in attention to the most likely violated assumption.
Nevertheless, in line with our approach of vigilant testing, we take these results to raise at least two potential issues: (1) this part of the task analysis may not be required for credibly improving over LM's zero-shot predictions, at least for the datasets we tested, as discussed more below; and (2) judgments about maxim violations may rely on less intuitive, less accessible information, making them more difficult to verbalize both for humans and LMs, a point we return to in Section~\ref{section:discussion}.

Second, to assess the interpretation proposer module, we evaluate the quality of the proposals generated by the interpretation proposer, given the trigger utterance in context and the most likely violated assumption.
Full details of the evaluation procedure and results are reported in Appendix~\ref{app:implicatures-interpretation-proposer}.
We ran a human evaluation study wherein we asked participants to rate how natural they think randomly selected LM-proposed interpretations were for what a speaker could plausibly want to convey with a given trigger utterance and assumption violation.
For comparison, participants were also asked to rate incorrect (literal) and target pragmatic interpretations from the forced choice interpretation options of the respective items, generated by us.
We found that, across, trigger conditions, the LM proposals were not rated credibly worse, nor credibly better than the human-written pragmatic interpretations ($\beta = 2.62 [-0.34, 5.49]$), but they were rated credibly worse than the incorrect interpretations ($\beta = 11.4 [7.86, 15.09 ]$).
That is, in line with results from previous case studies, the LM-based proposer seems to provide samples of comparable quality to human written alternatives, suggesting that proposers are a type of neural module where LMs already show promising performance, even under simple zero-shot prompting.

Both human-evaluation studies combined suggest a concise interpretation, namely that the interpretation proposer may not need to rely on information about violated assumptions to sample viable pragmatic interpretations of an utterance in context.
Indeed, the high accuracy and the comparable fit of the no-assumption model and the other AE models to human data also fits into this speculative conclusion.
If correct, the SAGE approach might therefore suggest that an algorithmic task decomposition of Gricean-like interpretation would be well served by a single module that reliably suggests context-dependent interpretation hypotheses, without necessarily having to deduce these interpretation alternatives from a longer-winded process of maxim-checking.
Of course, this is presently entirely speculative, but it points to the kinds of positive evidential contributions that a future, more matured SAGE approach could, in principle, make to cognitive modeling.

There are other aspects of the AE model implementation and the experimental set-up that suggest interesting directions for further testing the potential of SAGE for modeling Gricean theory of pragmatic interpretation.
For instance, the mapping of the LM-proposed interpretations to the multiple-choice dataset options was done through prompting another LM-based module and using majority voting. 
While LM-based scoring has become a common approach, alternative implementations, e.g., by calculating the cosine similarity between the samples and options, or retrieving the conditional probability of options, given the interpretations, could be more faithful \citep[cf.][]{hu-levy-2023-prompting, bavaresco-etal-2025-llms}.
Nonetheless, from a performance optimization perspective, the current interpretation mapping implementation already seems to provide a promising prompting procedure that outperforms zero-shot LMs in a multiple-choice task \citep[corroborated by the improvement of the no-assumptions model over the SP model, in line with, e.g.,][]{yao2023tree}.


In sum, this case study outlines a first attempt to build a simple computational implementation for a traditional Gricean account of a key pragmatic phenomenon, conversational implicature interpretation, that has been lacking within traditional computational modeling because of its complex context-dependence.
We provide a concrete example of how SAGE modules can be used to plausibly open-end the evaluation of traditional pragmatic assumptions about speaker behavior and flexibly sample adequate interpretations, supplying a first set of quantitative modeling results.
Although our results do not yet provide conclusive evidence about cognitively plausible algorithmic implementations of this computational-level Gricean account, we believe that our case study poses an informative starting point for exploring long-standing theoretical predictions from pragmatics through quantitative comparison between models and human data in future work. 

\section{Discussion \& Future Directions}
\label{section:discussion}

Computational modeling has made remarkable progress in explaining how humans go beyond literal interpretation through rich social and common sense reasoning.
Models of this process require contextually appropriate alternatives, such as interpretations or utterances, which the modeler usually had to supply manually.
In this paper, we outlined the computational framework \textbf{SAGE} for modeling open-ended pragmatic language use.
SAGE uses a transparent task-decomposition to scaffold flexible neural modules which dynamically sample and process these alternatives.

Across three case studies, we comprehensively stress-test the idea of using LMs as \textit{proposers} to sample the alternatives, and as \textit{evaluators} to use them in reasoning procedures specified by the computational model scaffold (cf.~Table~\ref{tab:case-studies-introduction} for an overview).
In the first case study, we focus on pragmatic referential expression generation in increasingly complex reference game contexts, using an LM proposer for generating candidate utterances and an LM evaluator for providing literal semantic judgments.
In the second case study, we used SAGE to implement a model of Manner implicature interpretation, using an LM proposer for sampling context-dependent alternative utterances and an LM evaluator for comparing the complexity of different utterance-context pairs. 
Finally, in the third case study, we implement a more encompassing model of conversational implicature interpretation through abductive reasoning based on Gricean conversational maxims.
Here, LM evaluators assessed whether Gricean maxims were flouted in a given context, and an LM proposer generated intended interpretations of the utterance based on the flouting.

Across case studies, SAGE models achieved high accuracy on the tasks, even in complex contexts, but
while performance may be an interesting indicator in applied domains, it is usually less important (than fit to human data) for adequacy of a cognitive model.
Models might achieve high end-to-end human-like accuracy despite non-human-like, theoretically counterintuitive or unreliable performance on subtasks within the computational model.
Careful evaluation of the modules and model comparison are therefore paramount steps of the modeling pipeline.
This is why, we discuss general challenges that SAGE and similar modeling frameworks might face in Section~\ref{sec:discussion-challenges} and ways to improve upon these challenges in Section~\ref{sec:discussion-solutions}.
We close with a general reflection on the contribution of SAGE-style frameworks in computational cognitive modeling and understanding of human pragmatic language in Section~\ref{sec:discussion-role-in-science}. 

\subsection{Challenges of SAGE-like frameworks}
\label{sec:discussion-challenges}

The key idea of SAGE lies in integrating LMs as stand-in components approximating open-ended information needed in models of pragmatic language use.
While literature in computational cognitive science has discussed in detail challenges connected to (probabilistic) cognitive models \citep[e.g.,][]{farrell2018computational}, we here focus on novel challenges that might arise specifically through introducing neural components such as LMs, which we explored in detail in the presented case studies.

\paragraph{Processing of LM outputs into structured formats can be brittle.}
While LM modules are able to flexibly propose and evaluate open-ended language in different contexts, their outputs often have to be processed into structured formats like lists or matrices for downstream computation in the cognitive model.
This is an instance of the common representational interface challenge for neuro-symbolic models \citep{bhuyan2024neuro}, which can suffer from brittleness and general difficulty of mapping outputs onto a limited space of symbols. 
In our case studies, we were able to perform necessary processing with simple regular expression parsing of the outputs and prompt engineering of formatting instructions for LM modules; however, less performant LMs might be less reliable and cause accumulation of errors throughout the scaffolding model (see, e.g., Appendix~\ref{app:sec:lm-configs} for a discussion of modeling with \texttt{Llama-3.1-8B-Inst.}).

\paragraph{Neural modules inherit general problems of LMs.}
Neuro-symbolic cognitive models containing LM modules inherits general challenges connected to LMs. 
For instance, previous work pointed out sensitivity to apparently minor changes to the prompt, lack of interpretability, potential hallucinations \citep[e.g.,][]{bender2021stochastic, ji2023hallucination, liu2023lost, zhao2023explainability}.
The usage of large LMs also raises a host of new ethical challenges, such as non-negligible energy and water consumption and biases.
Furthermore, non human-like interpretation of instructions has been observed in LMs  \citep{shi2023large}.
Recent work has shown that the content of supplementary information in a prompt, like the chain of thought, might not consistently improve results \citep{schaeffer2023invalid}, suggesting that including, e.g., theoretically motivated information in prompts of modules is not guaranteed to have an intuitive causal effect on LM predictions.
Further, LM performance might be susceptible to task and input data statistics relative to the training data \citep{mccoy2023embers}.
While in the context of SAGE, general sensitivity to prompting might affect both proposer and evaluator modules, proposers might struggle more with hallucinations, and evaluators with insensitivity to theoretically motivated prompting.
Qualitative manual analyses of the LM proposer in case study I revealed that a small fraction of candidate referential utterances indeed included details that were not originally available to the LM, especially at late iterations of the IM algorithm (see Section~\ref{section:cs1-results}). 
This issue could be alleviated by a well-performing evaluator to filter out candidates that are false for the target referent, akin to common self-verification prompting approaches in the LM literature \citep{wang2023selfconsistency}.
Proposals in the other case studies received high human evaluations, indicating that there likely were no hallucination issues.


\paragraph{LMs are more robust proposers than evaluators.}
Our detailed analyses of the modules revealed differences in how well LMs can be used as proposers and as evaluators.
In our case studies, proposers were used for two tasks: sampling utterances and sampling interpretations.
Proposer modules across case studies generally performed well, validated by comparable human ratings of the LM-proposed alternatives and human-generated alternatives (see Appendix~\ref{app:im-utt-proposer}, \ref{app:implicatures-interpretation-proposer}). Exploratory analyses during development also suggested that the proposers were mostly robust to variations in prompting.
These analyses also revealed an idiosyncrasy of LM proposers: the proposals might contain several surface form variations of the same kind of alternatives (e.g., similar utterances or several realizations of a pragmatic, rather than literal, interpretation).\footnote{
Arguably, this is not a limitation, but rather a more realistic implementation of alternatives, since instantiating any kind of alternative with only one example glosses over the natural variation in the surface form realization of a message.
}
To investigate whether this affected the models' behavior, in case study I we compared the performance of the SAGE iterative model with different numbers of sampled proposals, finding no significant differences. 

Evaluators were also used for different kinds of tasks: providing judgments in commonsense based tasks (e.g., intuitive assessments of how typical an input is) or in more theoretical tasks (e.g., computing literal semantics or identifying Gricean maxim flouting).
Human evaluation studies in case studies I and III and observations discussed above showed that the latter are more challenging for LM evaluators (see Appendix~\ref{app:im-sem-eval},~\ref{app:assumption-eval} for details).
More intuitive typicality judgments in case study II, on the other hand, had very high accuracy (see Appendix~\ref{app:prior-eval}). 
Additionally, in contrast to proposers, exploratory analyses of evaluator prompts during development showed that their performance was less robust to prompt variations, at least for subtasks in our case studies. 
We speculate that using LM evaluators for judgments that are intuitively easier to access for humans \citep[e.g., humans' subjective beliefs or preferences;][]{franke2016does} might generally lead to better performance than for factual or very theoretical judgments which are also more difficult to access for humans \citep[e.g., assessing objective frequencies;][]{kahneman1972subjective}.

The main take-aways of the module evaluations are summarized in Figure~\ref{fig:sage-overview}(B).
In sum, the result that LM proposers across case studies generally provided adequate reasoning alternatives is reassuring, since the departure point for our framework is the attempt to generate spaces of plausible alternatives \emph{in situ}, rather than manually supplying them a priori (see Section~\ref{section:introduction}).
We conjecture that there are promising paths towards solutions for the challenges posed by evaluators that will allow for more mature SAGE modeling, discussed in the next section.

\subsection{Paths for Improvement \& Recommendations for SAGE Modeling}
\label{sec:discussion-solutions}
Based on the vast research on advancing LM performance and the fast pace of developments in NLP, we next discuss natural directions for expected improvement of LM-based components in the SAGE framework.
Then, we outline general methodological recommendations for SAGE-style neuro-symbolic modeling identified with the help of our case studies.

\paragraph{Appropriate backbone LMs should be used for neural modules.}\label{sec:discussion-backbones}
The challenges described above point to both \textit{content-related} and \textit{format-related} properties of the LMs that should be considered when selecting LMs viable to serve as the backbone for neural modules.

First, the content of the module outputs should be plausible, both to avoid error accumulation and to ensure that the resulting cognitive model operates on sensible samples. 
In the presented case studies this was ensured by instantiating most neural components of the framework with large pretrained language models like \texttt{GPT-4} \citep{openai2023gpt4}, and careful qualitative analyses of the outputs and human evaluation studies, although we also show that, in principle, SAGE models can also be instantiated with small open-source models (as exemplified by the \texttt{Llama-3.1-8b-Inst} model in case study III, see Appendix~\ref{app:sec:lm-configs}).
The generally \textit{high performance} of state-of-the-art LMs across various tasks was sufficient to successfully employ them as proposers and, in part, as evaluators across case studies.
Nonetheless, it is likely that more elaborate prompting \citep[e.g., few-shot, chain-of-thought or ReAct prompting,][]{NEURIPS2020_1457c0d6, wei2022chain, yao2023react} would improve the performance of the modules, and, therefore, SAGE models, at the cost of introducing additional degrees of prompting freedom in the model configuration.
This is especially relevant for improving, e.g., theoretically-driven evaluators.
Additionally, as the development of LMs progresses, concerns like the LMs' insensitivity to instructions are likely to be gradually reduced.

Second, to ensure that the LM outputs can be processed by the cognitive model, structured output formats like lists or matrices are often required.
Formatting is especially important to avoid propagation of errors through the components of the cognitive model.
This requires that the LM is \textit{instruction fine-tuned} and can accurately follow formatting instructions \citep{ouyang2022training}. 
Alternatively, small LMs without instruction fine-tuning could be employed off the shelf for subtasks such as sampling reasoning alternatives formed by single words or short phrases that could easily be generated by continuing a prompt sentence.
In the reported case studies, we added specific instructions on how to format outputs to the module prompts (see Appendices for full prompts).
Given that mostly large LMs were employed, we were able to achieve sufficient robustness of the formatting of the output by engineering appropriate instructions and processing the outputs with regular expressions.
However, for SAGE to become a fully fledged framework, more robust and scalable approaches could be explored. 
One promising avenue is employing recently proposed procedures for retrieving structured or constrained information from LMs for more reliable typing and formatting of the module outputs \citep[e.g.,][]{Moskal2024, lipkin2025fast}.

With modern efficient training techniques like LoRA \citep{hu2022lora}, specialized LM backbones can be created for the purposes of SAGE modeling from smaller, open-source LMs.
For instance, small LMs could be fine-tuned for specific subtasks that commonly occur in computational models of pragmatic language use, such as literal semantic evaluation.
Over time, this could grow into a reusable, lightweight and open-source toolbox for SAGE models in computational pragmatics.
Fine-tuning datasets could be constructed with increasingly popular synthetic data generation methods, e.g., with stronger language models \citep{liu-etal-2022-wanli, kim-etal-2025-evaluating}, or through semi-automatic construction from templates \citep[like in, e.g.,][]{jeretic-etal-2020-natural, gandhi2023understanding}.
We encourage the creation of such a toolbox based on open-source LMs also to ensure reproducibility of modeling results and adherence to open science practices in computational modeling in pragmatics, even when LMs are used \citep[cf.][]{wieling2018reproducibility}.

\paragraph{Neural modules need fair task demands.}
Highly performant LM backbones are not sufficient.
Formulating the subtask to the LM module is also a crucial challenge. 
For instance, we observed that clear prompts, ideally addressing intuitive and contextualized rather than abstract or theoretical judgments, lead to better performance for evaluators.
Optimization of the modules for performance should not jeopardize conceptual soundness of the overall cognitive model; we merely advocate that, when several equivalent ways of posing a task are available, additional task demands should be minimized, such as capabilities presupposed for translating task instructions (e.g., selecting the most plausible alternative) into appropriate mechanisms (e.g., selecting the option with the highest conditional probability) \citep{hu2024auxiliary}.

In the reported case studies, we mostly used prompting to capture the required subtasks with neural modules. 
However, other approaches to leveraging neural models are especially promising for instantiating evaluators, or when using smaller, more efficient LM backbones \citep{hu2024auxiliary}.
For instance, more robust evaluators could use LM-internal information like conditional log probabilities of alternatives given the context, or similarity between embeddings of the LMs. 
Such evaluations might result in higher performance and be more faithful to the capabilities of the LMs \citep{hu-levy-2023-prompting, hu2024auxiliary}, but using correct methods for deriving condition-level estimates is crucial \citep{FrankeTsvilodub2024:Bayesian-Statis}. 
Finally, future work could explore other architectures directly designed for the intended subtasks: for instance, multi-modal models like vision language models could be used for grounded tasks like semantic evaluation of utterances applied to images.

The next step for SAGE is to investigate at larger scale which types of tasks are fair, and which task formulations work well for commonly occurring subtasks in models of human language use \citep[e.g., akin to the systematic evaluations in][]{tsvilodub2025integrating}. 

\subsection{Contributions of SAGE in computational modeling of pragmatic language}
\label{sec:discussion-role-in-science} 

The presented case studies explore SAGE models as a potential addition to the cognitive modeling toolbox, allowing modelers to scale computational models towards wider empirical coverage.
An important concern with SAGE for scientific modeling is that the black-box character of neural modules could undermine the model’s explanatory purpose by sacrificing the transparency needed for cognitive accounts.
Here we emphasize that SAGE models are based on transparent symbolic analyses of optimal computational procedures, their substeps, and how information is passed between these in order to solve the modeled higher-level task.
This analysis is targeted at a particular level of granularity (as adquate for the explanatory purposes).
The precise internal mechanisms of the substeps utilized by a given SAGE model are usually subject to explanations at more fine-grained levels of analysis; independently, of whether we think of the SAGE model itself as being (or its submodules as required) computational-, algorithmic- or even implementation-level in the sense of \citet{marr2010vision}.
That is, the LMs are only intended to instantiate un-analysed subtasks in the overall cognitive model, and not as explanatory components themselves.
Moreover, as stated in Section~\ref{sec:discussion-solutions}, we hypothesize that at least some of these concerns will be alleviated as language models become more robust and the field of interpretability develops.
Additionally, our results indicate that creating small specialized LM modules might be the most promising road towards full-scale SAGE models; sharing the modules and their custom training details within the community would also entail more transparency through open access and reusability across models.

In computational cognitive modeling, key insights about a studied phenomenon come from carefully evaluating a model or comparing it to alternatives using measures like, e.g., accuracy on the target task or predicting the explanandum, and fit to human or held out test data, alongside the usual statistical comparison criteria \citep{farrell2018computational}.
Naturally, SAGE models will also provide insights only under appropriate evaluation. 
We show examples of different evaluation procedures in the different case studies (see Table~\ref{tab:case-studies-introduction} for overview).  
For instance, in Section~\ref{sec:models-CS-IIIb} we outline theoretically informed model comparison of different computational structures through their fit to human data. 
This case study provided a novel computational model, living up to the promise of SAGE to enable quantitative modeling with hitherto verbal theories. 
By design, such novel models might rely on neural components which, in isolation, are rather difficult to compare to human manual specification.
Therefore, the explanatory insights from such novel models specifically rely on evaluation in terms of performance on the modeled task and model comparisons, as outlined above.
In edge cases, bad performance of the neural modules might go unnoticed or affect the overall modeling results in ways that are hard to interpret.
We hope that future work with SAGE-like models, transparent reporting of architectures and evaluations will lead to an accumulation of best practices and insights into robust uses of neural modules.
Additionally, given the importance of neural modules, it is natural to compare SAGE models with simple zero-shot LM baselines.
If a novel model is less accurate than an LM-only baseline, unfair task demands on modules may be hurting the overall model, or the performance of the LM based purely on its capabilities from statistical pretraining might not be improved with task decomposition.

SAGE translations of extant cognitive models (similar to case studies I and II) could contribute a validation of the framework itself (e.g., through benchmarking LM modules against human crowdsourced data), and confirmation that the cognitive model scales beyond manually specified alternatives to open-ended contexts.
We expect that cases where extant cognitive models instantiated with validated LM modules don't scale to open-ended data or underperform relative to the LM-only baseline will raise exciting open questions for the community around the contribution of the hypothesized task decomposition relative to statistical learning, and the criteria under which insights from language models will be taken to inform the revision of theories of human cognition \citep[cf.][]{frank2025cognitive}.

We justify our focus specifically on LMs as neural modules in these first exploratory case studies because, arguably, LMs are currently the best off-the-shelf computational means for sampling many types of context-conditional information that is required for modeling pragmatic language use.
Nonetheless, an important question for future research is whether every kind of information needed for more open-ended proposers and evaluators is retrievable from LMs.
While previous work has discussed what tasks might generally be out of reach for LMs \citep{van2023reclaiming, kambhampati2024llmscantplanhelp}, to our knowledge our framework is the first one to focus specifically on modeling pragmatic communication, so that results of our evaluations of the different neural modules provide a novel nuanced picture of LM capabilities specifically for purposes of modeling human language use.
Concretely, the observed difference in performance between proposers and evaluators suggests that, at least with simple prompting approaches, only simple alternatives and intuitive judgments may be robustly retrievable from LMs. 
That is, the retrievability of the required information from LMs intuitively seems to correlate with the ``effability'' and introspective accessibility of that type of information for humans.
It is an exciting question for future work whether this trend holds up across phenomena, and across LMs trained in different regimes (e.g., only through self-supervised learning on large-scale text data, or additionally fine-tuned on human feedback).

Another important open question for future work is whether relevant LM predictions like proposals should be interpreted as samples from a pool of different personas, where predictions represent some kind of average over the population represented in the training data, or something entirely different \citep{shanahan2024simulacra}.
This is especially relevant for SAGE as a framework for modeling human language, or any other aspect of human cognition, which is bound to inter-individually variable \citep[e.g.][]{StanovichWest2000:Individual-diff,KiddDonnelly2018:Individual-Diff,HaafRouder2019:Some-do-and-som}.
It is an active area of research whether individual variation can be elicited from LMs \citep{aher2023using, salewski2023context}, or if LMs can only predict item-level, but not subject-level variation \citep{FrankeTsvilodub2024:Bayesian-Statis}. 
Recent work suggests that fine-tuning on human experimental data may help improve individualized predictions \citep{binz2025foundation}.








\subsection{Conclusion}

A central question of cognitive science is the respective roles of cognitive or Bayesian models and powerful LMs for explaining domains like pragmatics. 
In line with \citet{futrell2025linguistics}, we argue that we should take the best of both worlds and combine the complementary strengths of both approaches.
In this work, we have proposed one way to do so in practice, by developing the SAGE framework to addresses the key challenges of pragmatic modeling: the specification of a space of reasoning alternatives.
We propose viewing SAGE as one methodological tool, together with other approaches, for generating converging evidence about the mechanisms underlying humans’ impressive ability to flexibly use pragmatic language.
At its current developmental stage, SAGE already offers a concrete roadmap towards computational models that operate on more flexible reasoning alternatives.
We started exploring this path with three case studies, analyzing prominent tasks arising in the domain of pragmatic communication.

Taken together, these case studies include a diverse, but not necessarily exhaustive, set of common components of SAGE models implemented as LM proposers and evaluators.
This provides an informative starting point for many future research directions, such as social language or argumentative language. 
Despite limitations, integrating modern LMs into the toolbox of cognitive scientists might offer a new research agenda on open-ending computational pragmatic models.

Even if the high standards for explanatory adequacy are not yet met by all SAGE models, at the very least the work towards full-fledged models already often produces novel, theoretically informed results on language model performance as a byproduct, potentially even offering useful blueprints for the architecture of LM-based systems \citep[or, agents, as discussed in Section~\ref{section:introduction}; cf.][]{sumers2023cognitive}. 

In the long run, we offer SAGE as a method that could provide a novel perspective, enriching the extant cognitive modeling toolbox.
Specifically, to borrow terminology from \citet{frank2025cognitive}, through the flexible interface of language models, SAGE enables exploring ``stimulus-compatible'' cognitive models, i.e., models that can be applied to the same real-world stimuli as humans are exposed to.
The idea of SAGE as a neuro-symbolic modeling approach is in line with the long-standing discussions pointing towards respective strengths and importance of incorporating both neural and more explicit models for explaining human cognition \citep[e.g.,][]{lake2017building, griffiths2024bayes}.
We believe that a suite of multi-faceted approaches will be needed to explain something as complex as language use across changing communication partners and open-ended contexts, and are hopeful that insights and opportunities offered by SAGE bring us one step closer to a more complete picture of the uniquely human domain of cognition that is language.

\newpage
\section*{Acknowledgments}
We would like to thank Martin Butz for helpful discussions, as well as audiences at the Computational Cognition Conference 2024, the Using Artificial Neural Networks for Studying Human Language Learning and Processing 2024 workshop, and KogWis 2025. MF is a member of the Machine Learning Cluster of Excellence at University of Tübingen, EXC number 2064/2 – Project number 39072764. PT is funded by the Deutsche Forschungsgemeinschaft (DFG, German Research Foundation) under project ID 579368432. MF and PT gratefully acknowledge the support by the state of Baden-Württemberg through bwHPC and the German Research Foundation (DFG) through grant INST 35/1597-1 FUGG.

\printbibliography[heading=bibintoc]

@preamble{"\newcommand{\SortNoop}[1]{}"}

@article{HaafRouder2019:Some-do-and-som,
	author = {Julia M. Haaf and Jeffrey N. Rouder},
	doi = {10.3758/s13423-018-1522-x},
	journal = {Psychonomic Bulletin \& Review},
	pages = {772--789},
	title = {Some do and some don't? Accounting for variability of individual difference structures},
	volume = {26},
	year = {2019}
}

@article{KiddDonnelly2018:Individual-Diff,
	author = {Kidd, Evan and Donnelly, Seamus and Christiansen, Morten H.},
	doi = {10.1016/j.tics.2017.11.006},
	issn = {1364-6613},
	journal = {Trends in Cognitive Sciences},
	number = {2},
	pages = {154--169},
	publisher = {Elsevier BV},
	title = {Individual Differences in Language Acquisition and Processing},
	volume = {22},
	year = {2018}}

@article{StanovichWest2000:Individual-diff,
	author = {Stanovich, Keith E. and West, Richard F.},
	doi = {10.1017/S0140525X00003435},
	journal = {Behavioral and Brain Sciences},
	number = {5},
	pages = {645--665},
	publisher = {Cambridge University Press},
	title = {Individual differences in reasoning: Implications for the rationality debate?},
	volume = {23},
	year = {2000},
}

@article{Jager2002:Some-Notes-on-t,
	author = {Gerhard J{\"{a}}ger},
	journal = {Journal of Logic, Language and Information},
	number = {4},
	pages = {427--451},
	title = {Some Notes on the Formal Properties of Bidirectional Optimality Theory},
	volume = {11},
	year = 2002}

@misc{BinzAkata2024:Centaur:-a-foun,
	archiveprefix = {arXiv},
	author = {Marcel Binz and Elif Akata and Matthias Bethge and Franziska Br{\"a}ndle and Fred Callaway and Julian Coda-Forno and Peter Dayan and Can Demircan and Maria K. Eckstein and No{\'e}mi {\'E}ltet{\H o} and Thomas L. Griffiths and Susanne Haridi and Akshay K. Jagadish and Li Ji-An and Alexander Kipnis and Sreejan Kumar and Tobias Ludwig and Marvin Mathony and Marcelo Mattar and Alireza Modirshanechi and Surabhi S. Nath and Joshua C. Peterson and Milena Rmus and Evan M. Russek and Tankred Saanum and Natalia Scharfenberg and Johannes A. Schubert and Luca M. Schulze Buschoff and Nishad Singhi and Xin Sui and Mirko Thalmann and Fabian Theis and Vuong Truong and Vishaal Udandarao and Konstantinos Voudouris and Robert Wilson and Kristin Witte and Shuchen Wu and Dirk Wulff and Huadong Xiong and Eric Schulz},
	eprint = {2410.20268},
	primaryclass = {cs.LG},
	title = {Centaur: a foundation model of human cognition},
	url = {https://arxiv.org/abs/2410.20268},
	year = {2024}}

@article{RutarWolff2022:Structure-Learn,
	author = {Rutar, Danaja and Wolff, Erwin de and Rooij, Iris van and Kwisthout, Johan},
	doi = {10.1007/s42113-022-00131-8},
	journal = {Computational Brain \& Behavior},
	pages = {1--10},
	title = {{Structure Learning in Predictive Processing Needs Revision}},
	year = {2022}}

@incollection{Rooijvan-RooijFranke2015:Optimality-Theo,
	author = {{\SortNoop{Rooij}}van Rooij, Robert and Michael Franke},
	booktitle = {The Stanford Encyclopedia of Philosophy},
	note = {Winter 2015 Edition},
	publisher = {Stanford University},
	title = {Optimality-Theoretic and Game-Theoretic Approaches to Implicatures},
	year = {2015}}

@phdthesis{Parikh1988:Language-and-st,
	author = {Prashant Parikh},
	school = {Stanford University},
	title = {Language and strategic inference},
	year = {1988}}

@incollection{HornDivisionofLabor1984,
	address = {Washington},
	author = {Laurence R. Horn},
	booktitle = {Meaning, Form, and Use in Context},
	editor = {Deborah Shiffrin},
	pages = {11-42},
	publisher = {Georgetown University Press},
	title = {Towards a New Taxonomy for Pragmatic Inference: {Q}-based and {R}-based Implicature},
	year = 1984}

@article{Katzir2007:Structurally-De,
	author = {Roni Katzir},
	doi = {doi:10.1007/s10988-008-9029-y},
	journal = {Linguistics and Philosophy},
	number = {6},
	pages = {669--690},
	title = {Structurally-Defined Alternatives},
	volume = {30},
	year = {2007}}

@article{GotznerRomoli2022:Meaning-and-Alt,
	author = {Gotzner, Nicole and Romoli, Jacopo},
	doi = {10.1146/annurev-linguistics-031220-012013},
	issn = {2333-9691},
	journal = {Annual Review of Linguistics},
	number = {1},
	pages = {213--234},
	publisher = {Annual Reviews},
	title = {Meaning and Alternatives},
	url = {http://dx.doi.org/10.1146/annurev-linguistics-031220-012013},
	volume = {8},
	year = {2022}}

@book{BenzJager2006:Game-Theory-and,
	address = {Hampshire},
	editor = {Anton Benz and Gerhard J{\"{a}}ger and {\SortNoop{Rooij}}van Rooij, Robert},
	publisher = {Palgrave MacMillan},
	title = {Game Theory and Pragmatics},
	year = {2006}}

@article{Franke2011:Quantity-Implic,
	author = {Michael Franke},
	doi = {doi:10.3765/sp.4.1},
	journal = {Semantics \& Pragmatics},
	number = {1},
	pages = {1--82},
	title = {Quantity Implicatures, Exhaustive Interpretation, and Rational Conversation},
	volume = {4},
	year = {2011}}

@article{BuccolaKriz2021:Conceptual-alte,
	author = {Buccola, Brian and Kri{\v z}, Manuel and Chemla, Emmanuel},
	doi = {10.1007/s10988-021-09327-w},
	journal = {Linguistics and Philosophy},
	number = {2},
	pages = {265--291},
	title = {Conceptual alternatives: Competition in language and beyond},
	url = {http://dx.doi.org/10.1007/s10988-021-09327-w},
	volume = {45},
	year = {2021}}

@article{FoxKatzir2011:On-the-Characte,
	author = {Danny Fox and Roni Katzir},
	doi = {10.1007/s11050-010-9065-3},
	journal = {Natural Language Semantics},
	pages = {87--107},
	title = {On the Characterization of Alternatives},
	volume = {19},
	year = {2011}}

@article{ReppSpalek2021:The-Role-of-Alt,
	author = {Repp, Sophie and Spalek, Katharina},
	doi = {10.3389/fcomm.2021.682009},
	journal = {Frontiers in Communication},
	publisher = {Frontiers Media SA},
	title = {The Role of Alternatives in Language},
	url = {http://dx.doi.org/10.3389/fcomm.2021.682009},
	volume = {6},
	year = {2021}}

@book{SperberWilson1995:Relevance:-Comm,
	address = {Oxford},
	author = {Dan Sperber and Deirdre Wilson},
	publisher = {Blackwell},
	title = {Relevance: Communication and Cognition (2nd ed.)},
	year = {1995}}

@book{Geurts2010:Quantity-Implic,
	address = {Cambridge, UK},
	author = {Bart Geurts},
	publisher = {Cambridge University Press},
	title = {Quantity Implicatures},
	year = {2010}}

@article{LassiterGoodman2015:Adjectival-vagu,
	author = {Daniel Lassiter and Noah D. Goodman},
	doi = {10.1007/s11229-015-0786-1},
	journal = {Synthese},
	number = {10},
	pages = {3801--3836},
	title = {Adjectival vagueness in a Bayesian model of interpretation},
	volume = {194},
	year = {2017}}

@inproceedings{BergenLevy2012:Thats-what-she-,
	author = {Leon Bergen and Roger Levy and Noah D. Goodman},
	booktitle = {Proceedings of the 34\textsuperscript{th} Annual Meeting of the Cognitive Science Society},
	pages = {120--125},
	title = {That's what she (could have) said: {H}ow alternative utterances affect language use},
	volume = {34},
	year = {2012}}

@book{BlutnerZeevat2004:Optimality-Theo,
	editor = {Reinhard Blutner and Henk Zeevat},
	publisher = {Palgrave MacMillan},
	title = {Optimality Theory and Pragmatics},
	year = {2004}}

@article{Burkner2018:Advanced-Bayesi,
	author = {Paul-Christian B{\"u}rkner},
	journal = {{The R Journal}},
	number = {1},
	pages = {395--411},
	title = {{Advanced Bayesian Multilevel Modeling with the R Package brms}},
	volume = {10},
	year = {2018}}

@inproceedings{PoliakNaradowsky2018:Hypothesis-Only,
	author = {Poliak, Adam and Naradowsky, Jason and Haldar, Aparajita and Rudinger, Rachel and Van Durme, Benjamin},
	booktitle = {Proceedings of the Seventh Joint Conference on Lexical and Computational Semantics},
	doi = {10.18653/v1/S18-2023},
	pages = {180--191},
	publisher = {Association for Computational Linguistics},
	title = {Hypothesis Only Baselines in Natural Language Inference},
	url = {https://aclanthology.org/S18-2023},
	year = {2018}}

@article{BommasaniHudson2021:On-the-opportun,
	author = {Bommasani, Rishi and Hudson, Drew A and Adeli, Ehsan and Altman, Russ and Arora, Simran and von Arx, Sydney and Bernstein, Michael S and Bohg, Jeannette and Bosselut, Antoine and Brunskill, Emma and others},
	journal = {arXiv preprint arXiv:2108.07258},
	title = {On the opportunities and risks of foundation models},
	year = {2021}}

@inproceedings{HoltzmanWest2021:Surface-Form-Co,
	address = {Online and Punta Cana, Dominican Republic},
	author = {Holtzman, Ari and West, Peter and Shwartz, Vered and Choi, Yejin and Zettlemoyer, Luke},
	booktitle = {Proceedings of the 2021 Conference on Empirical Methods in Natural Language Processing},
	pages = {7038--7051},
	publisher = {Association for Computational Linguistics},
	title = {Surface Form Competition: Why the Highest Probability Answer Isn{'}t Always Right},
	year = {2021}}

@article{Rett2020:Manner-implicat,
	author = {Jessica Rett},
	journal = {International Review of Pragmatics},
	number = {1},
	pages = {44 - 79},
	title = {Manner implicatures and how to spot them},
	volume = {12},
	year = {2020}}

@article{BergenLevy2014:Pragmatic-Reaso,
	author = {Leon Bergen and Roger Levy and Noah D. Goodman},
	journal = {Semantics \& Pragmatics},
	number = {20},
	pages = {1--83},
	title = {Pragmatic Reasoning through Semantic Inference},
	volume = {9},
	year = {2016}}

@phdthesis{Franke2009:Signal-to-Act:-,
	author = {Michael Franke},
	school = {Universiteit van Amsterdam},
	title = {Signal to Act: {G}ame Theory in Pragmatics},
	year = {2009}}

@article{BlutnerBOTJoS2000,
	author = {Reinhard Blutner},
	journal = {Journal of Semantics},
	pages = {189-216},
	title = {Some Aspects of Optimality in Natural Language Interpretation},
	volume = {17},
	year = {2000}}

@article{Deemter2023:Dimensions-of-E,
	author = {{\SortNoop{Deemter}}{van Deemter}, Kees},
	doi = {10.1162/coli_a_00480},
	journal = {Computational Linguistics},
	title = {Dimensions of Explanatory Value in NLP models},
	year = {2023}}

@inproceedings{ParikhPrashGameImplicature1992,
	address = {San Francisco},
	author = {Prashant Parikh},
	booktitle = {TARK '92: Proceedings of the 4th conference on Theoretical aspects of reasoning about knowledge},
	editor = {Yoram Moses},
	pages = {85--94},
	publisher = {Morgan Kaufmann Publishers Inc.},
	title = {A Game-Theoretic Account of Implicature},
	year = 1992}

@book{BuntBlack2000:Abduction-Belie,
	address = {Amsterdam, Philadelphia},
	author = {Harry Bunt and William Black},
	publisher = {John Benjamins},
	title = {Abduction, Belief and Context in Dialogue},
	year = {2000}}

@article{HobbsStickel1993:Interpretation-,
	author = {Jerry R. Hobbs and Mark Stickel and Paul Martin},
	journal = {Artificial Intelligence},
	pages = {69--142},
	title = {Interpretation as Abduction},
	volume = {63},
	year = {1993}}

@article{ouyang2022training,
  title={Training language models to follow instructions with human feedback},
  author={Ouyang, Long and Wu, Jeffrey and Jiang, Xu and Almeida, Diogo and Wainwright, Carroll and Mishkin, Pamela and Zhang, Chong and Agarwal, Sandhini and Slama, Katarina and Ray, Alex and others},
  journal={Advances in Neural Information Processing Systems},
  volume={35},
  pages={27730--27744},
  year={2022}
}

@misc{tsvilodub2023evaluating,
      title={Evaluating Pragmatic Abilities of Image Captioners on A3DS}, 
      author={Polina Tsvilodub and Michael Franke},
      year={2023},
      eprint={2305.12777},
      archivePrefix={arXiv},
      primaryClass={cs.CL}
}

@article{wilson2016manner,
  title={In a manner of speaking: an empirical investigation of Manner Implicatures},
  author={Wilson, Elspeth and Katsos, Napoleon},
  journal={Pre-proceedings of Trends in Experimental Pragmatics},
  pages={170--176},
  year={2016}
}

@incollection{grice1975logic,
  title={Logic and conversation},
  author={Grice, Herbert P},
  booktitle={Speech acts},
  pages={41--58},
  year={1975},
  publisher={Brill}
}

@book{levinson2000presumptive,
  title={Presumptive meanings: The theory of generalized conversational implicature},
  author={Levinson, Stephen C},
  year={2000},
  publisher={MIT press}
}

@article{krahmer2012computational,
  title={Computational generation of referring expressions: A survey},
  author={Krahmer, Emiel and van Deemter, Kees},
  journal={Computational Linguistics},
  volume={38},
  number={1},
  pages={173--218},
  year={2012},
  publisher={MIT Press One Rogers Street, Cambridge, MA 02142-1209, USA journals-info~…}
}

@article{dale1995computational,
  title={Computational interpretations of the Gricean maxims in the generation of referring expressions},
  author={Dale, Robert and Reiter, Ehud},
  journal={Cognitive science},
  volume={19},
  number={2},
  pages={233--263},
  year={1995},
  publisher={Elsevier}
}

@article{gatt2018survey,
  title={Survey of the state of the art in natural language generation: Core tasks, applications and evaluation},
  author={Gatt, Albert and Krahmer, Emiel},
  journal={Journal of Artificial Intelligence Research},
  volume={61},
  pages={65--170},
  year={2018}
}

@book{newell1972human,
  title={Human problem solving},
  author={Newell, Allen and Simon, Herbert Alexander and others},
  volume={104},
  number={9},
  year={1972},
  publisher={Prentice-hall Englewood Cliffs, NJ}
}

@article{lewis1969convention,
	title={Convention},
	publisher={Harvard university press},
	author={Lewis, David},
	journal={Cambridge, MA},
	year={1969}
}

@article{frank2012predicting,
	title={Predicting pragmatic reasoning in language games},
	author={Frank, Michael C and Goodman, Noah D.},
	journal={Science},
	volume={336},
	number={6084},
	pages={998--998},
	year={2012},
	publisher={American Association for the Advancement of Science}
}

@inproceedings{NEURIPS2020_1457c0d6,
 author = {Brown, Tom and Mann, Benjamin and Ryder, Nick and Subbiah, Melanie and Kaplan, Jared D and Dhariwal, Prafulla and Neelakantan, Arvind and Shyam, Pranav and Sastry, Girish and Askell, Amanda and Agarwal, Sandhini and Herbert-Voss, Ariel and Krueger, Gretchen and Henighan, Tom and Child, Rewon and Ramesh, Aditya and Ziegler, Daniel and Wu, Jeffrey and Winter, Clemens and Hesse, Chris and Chen, Mark and Sigler, Eric and Litwin, Mateusz and Gray, Scott and Chess, Benjamin and Clark, Jack and Berner, Christopher and McCandlish, Sam and Radford, Alec and Sutskever, Ilya and Amodei, Dario},
 booktitle = {Advances in Neural Information Processing Systems},
 editor = {H. Larochelle and M. Ranzato and R. Hadsell and M.F. Balcan and H. Lin},
 pages = {1877--1901},
 publisher = {Curran Associates, Inc.},
 title = {Language Models are Few-Shot Learners},
 volume = {33},
 year = {2020}
}

@misc{openai2023gpt4,
      title={GPT-4 Technical Report}, 
      author={OpenAI},
      year={2023},
      eprint={2303.08774},
      archivePrefix={arXiv},
      primaryClass={cs.CL}
}

@article{touvron2023llama,
  title={Llama: Open and efficient foundation language models},
  author={Touvron, Hugo and Lavril, Thibaut and Izacard, Gautier and Martinet, Xavier and Lachaux, Marie-Anne and Lacroix, Timoth{\'e}e and Rozi{\`e}re, Baptiste and Goyal, Naman and Hambro, Eric and Azhar, Faisal and others},
  journal={arXiv preprint arXiv:2302.13971},
  year={2023}
}

@article{dubey2024llama,
  title={The llama 3 herd of models},
  author={Dubey, Abhimanyu and Jauhri, Abhinav and Pandey, Abhinav and Kadian, Abhishek and Al-Dahle, Ahmad and Letman, Aiesha and Mathur, Akhil and Schelten, Alan and Yang, Amy and Fan, Angela and others},
  journal={arXiv e-prints},
  pages={arXiv--2407},
  year={2024}
}

@article{wei2022chain,
	title={Chain of thought prompting elicits reasoning in large language models},
	author={Wei, Jason and Wang, Xuezhi and Schuurmans, Dale and Bosma, Maarten and Chi, Ed and Le, Quoc and Zhou, Denny},
	journal={arXiv preprint arXiv:2201.11903},
	year={2022}
}

@inproceedings{yao2023react,
  title={React: Synergizing reasoning and acting in language models},
  author={Yao, Shunyu and Zhao, Jeffrey and Yu, Dian and Du, Nan and Shafran, Izhak and Narasimhan, Karthik and Cao, Yuan},
  booktitle={International Conference on Learning Representations (ICLR)},
  year={2023}
}

@article{gao2022pal,
  title={PAL: Program-aided Language Models},
  author={Gao, Luyu and Madaan, Aman and Zhou, Shuyan and Alon, Uri and Liu, Pengfei and Yang, Yiming and Callan, Jamie and Neubig, Graham},
  journal={arXiv preprint arXiv:2211.10435},
  year={2022}
}

@misc{wong2023word,
      title={From Word Models to World Models: Translating from Natural Language to the Probabilistic Language of Thought}, 
      author={Lionel Wong and Gabriel Grand and Alexander K. Lew and Noah D. Goodman and Vikash K. Mansinghka and Jacob Andreas and Joshua B. Tenenbaum},
      year={2023},
      eprint={2306.12672},
      archivePrefix={arXiv},
      primaryClass={cs.CL}
}

@article{yao2023tree,
  title={Tree of thoughts: Deliberate problem solving with large language models},
  author={Yao, Shunyu and Yu, Dian and Zhao, Jeffrey and Shafran, Izhak and Griffiths, Thomas L and Cao, Yuan and Narasimhan, Karthik},
  journal={arXiv preprint arXiv:2305.10601},
  year={2023}
}

@inproceedings{hu-etal-2023-fine,
    title = "A fine-grained comparison of pragmatic language understanding in humans and language models",
    author = "Hu, Jennifer  and
      Floyd, Sammy  and
      Jouravlev, Olessia  and
      Fedorenko, Evelina  and
      Gibson, Edward",
    booktitle = "Proceedings of the 61st Annual Meeting of the Association for Computational Linguistics (Volume 1: Long Papers)",
    month = jul,
    year = "2023",
    address = "Toronto, Canada",
    publisher = "Association for Computational Linguistics",
    url = "https://aclanthology.org/2023.acl-long.230",
    doi = "10.18653/v1/2023.acl-long.230",
    pages = "4194--4213",
}

@inproceedings{liu-etal-2022-generated,
    title = "Generated Knowledge Prompting for Commonsense Reasoning",
    author = "Liu, Jiacheng  and
      Liu, Alisa  and
      Lu, Ximing  and
      Welleck, Sean  and
      West, Peter  and
      Le Bras, Ronan  and
      Choi, Yejin  and
      Hajishirzi, Hannaneh",
    booktitle = "Proceedings of the 60th Annual Meeting of the Association for Computational Linguistics (Volume 1: Long Papers)",
    month = may,
    year = "2022",
    address = "Dublin, Ireland",
    publisher = "Association for Computational Linguistics",
    url = "https://aclanthology.org/2022.acl-long.225",
    doi = "10.18653/v1/2022.acl-long.225",
    pages = "3154--3169",
}

@misc{schaeffer2023invalid,
      title={Invalid Logic, Equivalent Gains: The Bizarreness of Reasoning in Language Model Prompting}, 
      author={Rylan Schaeffer and Kateryna Pistunova and Samar Khanna and Sarthak Consul and Sanmi Koyejo},
      year={2023},
      eprint={2307.10573},
      archivePrefix={arXiv},
      primaryClass={cs.AI}
}

@article{sumers2023cognitive,
  title={Cognitive architectures for language agents},
  author={Sumers, Theodore and Yao, Shunyu and Narasimhan, Karthik and Griffiths, Thomas L},
  journal={arXiv preprint arXiv:2309.02427},
  year={2023}
}

@article{mccoy2023embers,
  title={Embers of autoregression: Understanding large language models through the problem they are trained to solve},
  author={McCoy, R Thomas and Yao, Shunyu and Friedman, Dan and Hardy, Matthew and Griffiths, Thomas L},
  journal={arXiv preprint arXiv:2309.13638},
  year={2023}
}

@article{he2023solving,
  title={Solving math word problems by combining language models with symbolic solvers},
  author={He-Yueya, Joy and Poesia, Gabriel and Wang, Rose E and Goodman, Noah D.},
  journal={arXiv preprint arXiv:2304.09102},
  year={2023}
}

@article{creswell2022selection,
  title={Selection-inference: Exploiting large language models for interpretable logical reasoning},
  author={Creswell, Antonia and Shanahan, Murray and Higgins, Irina},
  journal={arXiv preprint arXiv:2205.09712},
  year={2022}
}

@inproceedings{hu-levy-2023-prompting,
    title = "Prompting is not a substitute for probability measurements in large language models",
    author = "Hu, Jennifer  and
      Levy, Roger",
    editor = "Bouamor, Houda  and
      Pino, Juan  and
      Bali, Kalika",
    booktitle = "Proceedings of the 2023 Conference on Empirical Methods in Natural Language Processing",
    month = dec,
    year = "2023",
    address = "Singapore",
    publisher = "Association for Computational Linguistics",
    url = "https://aclanthology.org/2023.emnlp-main.306/",
    doi = "10.18653/v1/2023.emnlp-main.306",
    pages = "5040--5060"
}

@article{lake2017building,
  title={Building machines that learn and think like people},
  author={Lake, Brenden M and Ullman, Tomer D and Tenenbaum, Joshua B and Gershman, Samuel J},
  journal={Behavioral and brain sciences},
  volume={40},
  pages={e253},
  year={2017},
  publisher={Cambridge University Press}
}

@book{marr2010vision,
  title={Vision: A computational investigation into the human representation and processing of visual information},
  author={Marr, David},
  year={2010},
  publisher={MIT press}
}

@book{farrell2018computational,
  title={Computational modeling of cognition and behavior},
  author={Farrell, Simon and Lewandowsky, Stephan},
  year={2018},
  publisher={Cambridge University Press}
}

@incollection{atlas1981clefts,
  title={It-clefts, informativeness and logical form: Radical pragmatics (revised standard version)},
  author={Atlas, Jay David and Levinson, Stephen C},
  booktitle={Radical pragmatics},
  pages={1--62},
  year={1981},
  publisher={Academic Press}
}

@article{paranjape2023art,
  title={{ART}: Automatic multi-step reasoning and tool-use for large language models},
  author={Paranjape, Bhargavi and Lundberg, Scott and Singh, Sameer and Hajishirzi, Hannaneh and Zettlemoyer, Luke and Ribeiro, Marco Tulio},
  journal={ar{X}iv preprint ar{X}iv:2303.09014},
  year={2023}
}

@misc{poesia2023certified,
      title={Certified Deductive Reasoning with Language Models}, 
      author={Gabriel Poesia and Kanishk Gandhi and Eric Zelikman and Noah D. Goodman},
      year={2023},
      eprint={2306.04031},
      archivePrefix={ar{X}iv},
      primaryClass={cs.AI}
}

@article{frank2023large,
  title={Large language models as models of human cognition},
  author={Frank, Michael C},
  year={2023},
  publisher={PsyArXiv}
}

@inproceedings{
hu2024auxiliary,
title={Auxiliary task demands mask the capabilities of smaller language models},
author={Jennifer Hu and Michael Frank},
booktitle={First Conference on Language Modeling},
year={2024},
url={https://openreview.net/forum?id=U5BUzSn4tD}
}

@inproceedings{bender2021stochastic,
author = {Bender, Emily M. and Gebru, Timnit and McMillan-Major, Angelina and Mitchell, Margaret},
title = {On the Dangers of Stochastic Parrots: Can Language Models Be Too Big?},
year = {2021},
isbn = {9781450383097},
publisher = {Association for Computing Machinery},
address = {New York, NY, USA},
url = {https://doi.org/10.1145/3442188.3445922},
doi = {10.1145/3442188.3445922},
booktitle = {Proceedings of the 2021 ACM Conference on Fairness, Accountability, and Transparency},
pages = {610–623},
numpages = {14},
location = {Virtual Event, Canada},
series = {FAccT '21}
}

@article{zhao2023explainability,
  title={Explainability for large language models: A survey},
  author={Zhao, Haiyan and Chen, Hanjie and Yang, Fan and Liu, Ninghao and Deng, Huiqi and Cai, Hengyi and Wang, Shuaiqiang and Yin, Dawei and Du, Mengnan},
  journal={ACM Transactions on Intelligent Systems and Technology},
  year={2023},
  publisher={ACM New York, NY}
}

@misc{shi2023large,
      title={Large Language Models Can Be Easily Distracted by Irrelevant Context}, 
      author={Freda Shi and Xinyun Chen and Kanishka Misra and Nathan Scales and David Dohan and Ed Chi and Nathanael Schärli and Denny Zhou},
      year={2023},
      eprint={2302.00093},
      archivePrefix={arXiv},
      primaryClass={cs.CL}
}

@misc{liu2023lost,
      title={Lost in the Middle: How Language Models Use Long Contexts}, 
      author={Nelson F. Liu and Kevin Lin and John Hewitt and Ashwin Paranjape and Michele Bevilacqua and Fabio Petroni and Percy Liang},
      year={2023},
      eprint={2307.03172},
      archivePrefix={arXiv},
      primaryClass={cs.CL}
}

@article{ji2023hallucination,
author = {Ji, Ziwei and Lee, Nayeon and Frieske, Rita and Yu, Tiezheng and Su, Dan and Xu, Yan and Ishii, Etsuko and Bang, Ye Jin and Madotto, Andrea and Fung, Pascale},
title = {Survey of Hallucination in Natural Language Generation},
year = {2023},
issue_date = {December 2023},
publisher = {Association for Computing Machinery},
address = {New York, NY, USA},
volume = {55},
number = {12},
issn = {0360-0300},
url = {https://doi.org/10.1145/3571730},
doi = {10.1145/3571730},
journal = {ACM Comput. Surv.},
month = {mar},
articleno = {248},
numpages = {38}
}

@article{van2023reclaiming,
  title={Reclaiming AI as a theoretical tool for cognitive science},
  author={Van Rooij, Iris and Guest, Olivia and Adolfi, Federico and de Haan, Ronald and Kolokolova, Antonina and Rich, Patricia},
  journal={Computational Brain \& Behavior},
  volume={7},
  number={4},
  pages={616--636},
  year={2024},
  publisher={Springer}
}

@inproceedings{romero2023synergistic,
  title={Synergistic integration of large language models and cognitive architectures for robust ai: An exploratory analysis},
  author={Romero, Oscar J and Zimmerman, John and Steinfeld, Aaron and Tomasic, Anthony},
  booktitle={Proceedings of the AAAI Symposium Series},
  volume={2},
  number={1},
  pages={396--405},
  year={2023}
}

@misc{kambhampati2024llmscantplanhelp,
      title={LLMs Can't Plan, But Can Help Planning in LLM-Modulo Frameworks}, 
      author={Subbarao Kambhampati and Karthik Valmeekam and Lin Guan and Mudit Verma and Kaya Stechly and Siddhant Bhambri and Lucas Saldyt and Anil Murthy},
      year={2024},
      eprint={2402.01817},
      archivePrefix={arXiv},
      primaryClass={cs.AI},
      url={https://arxiv.org/abs/2402.01817}, 
}

@article{bhuyan2024neuro,
  title={Neuro-symbolic artificial intelligence: a survey},
  author={Bhuyan, Bikram Pratim and Ramdane-Cherif, Amar and Tomar, Ravi and Singh, TP},
  journal={Neural Computing and Applications},
  pages={1--36},
  year={2024},
  publisher={Springer}
}

@article{srivastava2023-BIGbench,
  title={Beyond the imitation game: Quantifying and extrapolating the capabilities of language models},
  author={Srivastava, Aarohi and Rastogi, Abhinav and Rao, Abhishek and Shoeb, Abu Awal Md and Abid, Abubakar and Fisch, Adam and Brown, Adam R and Santoro, Adam and Gupta, Aditya and Garriga-Alonso, Adri{\`a} and others},
  journal={arXiv preprint arXiv:2206.04615},
  year={2022},
  ulr={https://arxiv.org/abs/2206.04615}
}

@article{PerezRinger2023:Discovering-Lan,
  title={Discovering language model behaviors with model-written evaluations},
  author={Perez, Ethan and Ringer, Sam and Luko{\v{s}}i{\=u}t{\.e}, Kamil{\.e} and Nguyen, Karina and Chen, Edwin and Heiner, Scott and Pettit, Craig and Olsson, Catherine and Kundu, Sandipan and Kadavath, Saurav and others},
  journal={arXiv preprint arXiv:2212.09251},
  year={2022}
}

@article{petroni2019language,
  title={Language models as knowledge bases?},
  author={Petroni, Fabio and Rockt{\"a}schel, Tim and Lewis, Patrick and Bakhtin, Anton and Wu, Yuxiang and Miller, Alexander H and Riedel, Sebastian},
  journal={arXiv preprint arXiv:1909.01066},
  year={2019}
}

@article{nye2021improving,
  title={Improving coherence and consistency in neural sequence models with dual-system, neuro-symbolic reasoning},
  author={Nye, Maxwell and Tessler, Michael and Tenenbaum, Josh and Lake, Brenden M},
  journal={Advances in Neural Information Processing Systems},
  volume={34},
  pages={25192--25204},
  year={2021}
}

@article{tsvilodub2024cognitive,
  title={Cognitive Modeling with Scaffolded LLMs: A Case Study of Referential Expression Generation},
  author={Tsvilodub, Polina and Franke, Michael and Carcassi, Fausto},
  journal={arXiv preprint arXiv:2407.03805},
  year={2024}
}

@inproceedings{ParkOBrien2023:Generative-Agen,
  title={Generative agents: Interactive simulacra of human behavior},
  author={Park, Joon Sung and O'Brien, Joseph and Cai, Carrie Jun and Morris, Meredith Ringel and Liang, Percy and Bernstein, Michael S},
  booktitle={Proceedings of the 36th Annual ACM Symposium on User Interface Software and Technology},
  pages={1--22},
  year={2023}
}

@inproceedings{devlin-etal-2019-bert,
    title = "{BERT}: Pre-training of Deep Bidirectional Transformers for Language Understanding",
    author = "Devlin, Jacob  and
      Chang, Ming-Wei  and
      Lee, Kenton  and
      Toutanova, Kristina",
    editor = "Burstein, Jill  and
      Doran, Christy  and
      Solorio, Thamar",
    booktitle = "Proceedings of the 2019 Conference of the North {A}merican Chapter of the Association for Computational Linguistics: Human Language Technologies, Volume 1 (Long and Short Papers)",
    month = jun,
    year = "2019",
    address = "Minneapolis, Minnesota",
    publisher = "Association for Computational Linguistics",
    url = "https://aclanthology.org/N19-1423",
    doi = "10.18653/v1/N19-1423",
    pages = "4171--4186",
}

@article{binz2023turning,
  title={Turning large language models into cognitive models},
  author={Binz, Marcel and Schulz, Eric},
  journal={arXiv preprint arXiv:2306.03917},
  year={2023}
}

@misc{burgess20183d,
	title={3d shapes dataset},
	author={Burgess, Chris and Kim, Hyunjik},
	year={2018}
}

@misc{FrankeTsvilodub2024:Bayesian-Statis,
	archiveprefix = {arXiv},
	author = {Michael Franke and Polina Tsvilodub and Fausto Carcassi},
	eprint = {2406.09012},
	title = {{Bayesian Statistical Modeling with Predictors from LLMs}},
	year = {2024}}

@article{Ferreira2019:A-Mechanistic-F,
	author = {Victor S. Ferreira},
	journal = {Annual Review of Psychology},
	number = {1},
	pages = {29--51},
	title = {A Mechanistic Framework for Explaining Audience Design in Language Production},
	volume = {70},
	year = 2019}

@article{degen2023rational,
  title={The rational speech act framework},
  author={Degen, Judith},
  journal={Annual Review of Linguistics},
  volume={9},
  number={1},
  pages={519--540},
  year={2023},
  publisher={Annual Reviews}
}

@article{
kao2014nonliteral,
author = {Justine T. Kao  and Jean Y. Wu  and Leon Bergen  and Noah D. Goodman },
title = {Nonliteral understanding of number words},
journal = {Proceedings of the National Academy of Sciences},
volume = {111},
number = {33},
pages = {12002-12007},
year = {2014},
doi = {10.1073/pnas.1407479111},
URL = {https://www.pnas.org/doi/abs/10.1073/pnas.1407479111},
eprint = {https://www.pnas.org/doi/pdf/10.1073/pnas.1407479111}}

@inproceedings{kao2014formalizing,
  title={Formalizing the pragmatics of metaphor understanding},
  author={Kao, Justine T. and Bergen, Leon and Goodman, Noah},
  booktitle={Proceedings of the annual meeting of the Cognitive Science Society},
  volume={36},
  number={36},
  year={2014}
}

@article{wang2024survey,
  title={A survey on large language model based autonomous agents},
  author={Wang, Lei and Ma, Chen and Feng, Xueyang and Zhang, Zeyu and Yang, Hao and Zhang, Jingsen and Chen, Zhiyuan and Tang, Jiakai and Chen, Xu and Lin, Yankai and others},
  journal={Frontiers of Computer Science},
  volume={18},
  number={6},
  pages={186345},
  year={2024},
  publisher={Springer}
}

@article{shinn2023reflexion,
  title={Reflexion: {L}anguage agents with verbal reinforcement learning},
  author={Shinn, Noah and Cassano, Federico and Gopinath, Ashwin and Narasimhan, Karthik and Yao, Shunyu},
  journal={Advances in Neural Information Processing Systems},
  volume={36},
  pages={8634--8652},
  year={2023}
}

@article{liu2023agentbench,
  title={Agentbench: Evaluating {LLMs} as agents},
  author={Liu, Xiao and Yu, Hao and Zhang, Hanchen and Xu, Yifan and Lei, Xuanyu and Lai, Hanyu and Gu, Yu and Ding, Hangliang and Men, Kaiwen and Yang, Kejuan and others},
  journal={arXiv preprint arXiv:2308.03688},
  year={2023}
}

@article{frank2025cognitive,
  title={Cognitive modeling using artificial intelligence. },
  author={Frank, Michael C. and Goodman, Noah D.},
  urls={https://doi.org/10.31234/osf.io/wv7mg_v1},
  year={2025}
}

@article{futrell2025linguistics,
  title={How Linguistics Learned to Stop Worrying and Love the Language Models},
  author={Futrell, Richard and Mahowald, Kyle},
  journal={arXiv preprint arXiv:2501.17047},
  year={2025}
}

@article{tsvilodub2025integrating,
  title={Integrating Neural and Symbolic Components in a Model of Pragmatic Question-Answering},
  author={Tsvilodub, Polina and Hawkins, Robert D. and Franke, Michael},
  journal={Society for Computation in Linguistics},
  volume={8},
  number={1},
  year={2025},
  publisher={University of Massachusetts Amherst Libraries}
}

@article{tsvilodub2025non,
  title={Non-literal Understanding of Number Words by Language Models},
  author={Tsvilodub, Polina and Gandhi, Kanishk and Zhao, Haoran and Fr{\"a}nken, Jan-Philipp and Franke, Michael and Goodman, Noah D},
  journal={arXiv preprint arXiv:2502.06204},
  year={2025}
}

@article{wieling2018reproducibility,
  title={Reproducibility in computational linguistics: Are we willing to share?},
  author={Wieling, Martijn and Rawee, Josine and van Noord, Gertjan},
  journal={Computational Linguistics},
  volume={44},
  number={4},
  pages={641--649},
  year={2018},
  publisher={MIT Press One Rogers Street, Cambridge, MA 02142-1209, USA journals-info~…}
}

@article{binz2025foundation,
  title={A foundation model to predict and capture human cognition},
  author={Binz, Marcel and Akata, Elif and Bethge, Matthias and Br{\"a}ndle, Franziska and Callaway, Fred and Coda-Forno, Julian and Dayan, Peter and Demircan, Can and Eckstein, Maria K and {\'E}ltet{\H{o}}, No{\'e}mi and others},
  journal={Nature},
  pages={1--8},
  year={2025},
  publisher={Nature Publishing Group UK London}
}

@inproceedings{aher2023using,
  title={Using large language models to simulate multiple humans and replicate human subject studies},
  author={Aher, Gati V and Arriaga, Rosa I and Kalai, Adam Tauman},
  booktitle={International Conference on Machine Learning},
  pages={337--371},
  year={2023},
  organization={PMLR}
}

@article{salewski2023context,
  title={In-context impersonation reveals large language models' strengths and biases},
  author={Salewski, Leonard and Alaniz, Stephan and Rio-Torto, Isabel and Schulz, Eric and Akata, Zeynep},
  journal={Advances in neural information processing systems},
  volume={36},
  pages={72044--72057},
  year={2023}
}

@article{sucholutsky2025using,
  title={Using LLMs to Advance the Cognitive Science of Collectives},
  author={Sucholutsky, Ilia and Collins, Katherine M and Jacoby, Nori and Thompson, Bill D and Hawkins, Robert D},
  journal={arXiv preprint arXiv:2506.00052},
  year={2025}
}

@inproceedings{bourgin2019cognitive,
  title={Cognitive model priors for predicting human decisions},
  author={Bourgin, David D and Peterson, Joshua C and Reichman, Daniel and Russell, Stuart J and Griffiths, Thomas L},
  booktitle={International conference on machine learning},
  pages={5133--5141},
  year={2019},
  organization={PMLR}
}

@article{griffiths2024bayes,
  title={Bayes in the age of intelligent machines},
  author={Griffiths, Thomas L and Zhu, Jian-Qiao and Grant, Erin and Thomas McCoy, R},
  journal={Current Directions in Psychological Science},
  volume={33},
  number={5},
  pages={283--291},
  year={2024},
  publisher={Sage Publications Sage CA: Los Angeles, CA}
}

@article{spinoso2025rsa,
  title={(RSA)\^{} 2: A Rhetorical-Strategy-Aware Rational Speech Act Framework for Figurative Language Understanding},
  author={Spinoso-Di Piano, Cesare},
  journal={Society for Computation in Linguistics},
  volume={8},
  number={1},
  year={2025},
  publisher={University of Massachusetts Amherst Libraries}
}

@misc{Moskal2024,
  author = {Moskal, Michal and Musuvathi, Madan and {K\i c\i man}, Emre},
  title = {{AI Controller Interface}},
  year = {2024},
  publisher = {{GitHub}},
  journal = {{GitHub} repository},
  howpublished = {\url{https://github.com/microsoft/aici/}}
}

@article{cong2024manner,
  title={Manner implicatures in large language models},
  author={Cong, Yan},
  journal={Scientific Reports},
  volume={14},
  number={1},
  pages={29113},
  year={2024},
  publisher={Nature Publishing Group UK London}
}

@article{wong2025modeling,
  title={Modeling Open-World Cognition as On-Demand Synthesis of Probabilistic Models},
  author={Wong, Lionel and Collins, Katherine M and Ying, Lance and Zhang, Cedegao E and Weller, Adrian and Gersternberg, Tobias and O'Donnell, Timothy and Lew, Alexander K and Andreas, Jacob D and Tenenbaum, Joshua B and others},
  journal={arXiv preprint arXiv:2507.12547},
  year={2025}
}

@inproceedings{franke2016does,
  title={What does the crowd believe? A hierarchical approach to estimating subjective beliefs from empirical data},
  author={Franke, Michael and Dablander, Fabian and Bennett, Erin and Degen, Judith and Tessler, Michael Henry and Kao, Justine and Goodman, Noah D and others},
  booktitle={Proceedings of the Annual Meeting of the Cognitive Science Society},
  volume={38},
  year={2016}
}

@article{liu2023evaluating,
  title={Evaluating the logical reasoning ability of chatgpt and gpt-4},
  author={Liu, Hanmeng and Ning, Ruoxi and Teng, Zhiyang and Liu, Jian and Zhou, Qiji and Zhang, Yue},
  journal={arXiv preprint arXiv:2304.03439},
  year={2023}
}

@article{wang2019superglue,
  title={Superglue: A stickier benchmark for general-purpose language understanding systems},
  author={Wang, Alex and Pruksachatkun, Yada and Nangia, Nikita and Singh, Amanpreet and Michael, Julian and Hill, Felix and Levy, Omer and Bowman, Samuel},
  journal={Advances in neural information processing systems},
  volume={32},
  year={2019}
}

@inproceedings{bowman-etal-2015-large,
    title = "A large annotated corpus for learning natural language inference",
    author = "Bowman, Samuel R.  and
      Angeli, Gabor  and
      Potts, Christopher  and
      Manning, Christopher D.",
    editor = "M{\`a}rquez, Llu{\'\i}s  and
      Callison-Burch, Chris  and
      Su, Jian",
    booktitle = "Proceedings of the 2015 Conference on Empirical Methods in Natural Language Processing",
    month = sep,
    year = "2015",
    address = "Lisbon, Portugal",
    publisher = "Association for Computational Linguistics",
    url = "https://aclanthology.org/D15-1075",
    doi = "10.18653/v1/D15-1075",
    pages = "632--642",
}

@inproceedings{bavaresco-etal-2025-llms,
    title = "{LLM}s instead of Human Judges? A Large Scale Empirical Study across 20 {NLP} Evaluation Tasks",
    author = "Bavaresco, Anna  and
      Bernardi, Raffaella  and
      Bertolazzi, Leonardo  and
      Elliott, Desmond  and
      Fern{\'a}ndez, Raquel  and
      Gatt, Albert  and
      Ghaleb, Esam  and
      Giulianelli, Mario  and
      Hanna, Michael  and
      Koller, Alexander  and
      Martins, Andre  and
      Mondorf, Philipp  and
      Neplenbroek, Vera  and
      Pezzelle, Sandro  and
      Plank, Barbara  and
      Schlangen, David  and
      Suglia, Alessandro  and
      Surikuchi, Aditya K  and
      Takmaz, Ece  and
      Testoni, Alberto",
    editor = "Che, Wanxiang  and
      Nabende, Joyce  and
      Shutova, Ekaterina  and
      Pilehvar, Mohammad Taher",
    booktitle = "Proceedings of the 63rd Annual Meeting of the Association for Computational Linguistics (Volume 2: Short Papers)",
    month = jul,
    year = "2025",
    address = "Vienna, Austria",
    publisher = "Association for Computational Linguistics",
    url = "https://aclanthology.org/2025.acl-short.20/",
    doi = "10.18653/v1/2025.acl-short.20",
    pages = "238--255",
    ISBN = "979-8-89176-252-7"
}

@article{kahneman1972subjective,
  title={Subjective probability: A judgment of representativeness},
  author={Kahneman, Daniel and Tversky, Amos},
  journal={Cognitive psychology},
  volume={3},
  number={3},
  pages={430--454},
  year={1972},
  publisher={Elsevier}
}

@inproceedings{
    lipkin2025fast,
    title={Fast Controlled Generation from Language Models with Adaptive Weighted Rejection Sampling},
    author={Ben Lipkin and Benjamin LeBrun and Jacob Hoover Vigly and Jo{\~a}o Loula and David R. MacIver and Li Du and Jason Eisner and Ryan Cotterell and Vikash Mansinghka and Timothy J. O'Donnell and Alexander K. Lew and Tim Vieira},
    booktitle={Second Conference on Language Modeling},
    year={2025},
    url={https://openreview.net/forum?id=3BmPSFAdq3}
}

@inproceedings{liu-etal-2022-wanli,
    title = "{WANLI}: Worker and {AI} Collaboration for Natural Language Inference Dataset Creation",
    author = "Liu, Alisa  and
      Swayamdipta, Swabha  and
      Smith, Noah A.  and
      Choi, Yejin",
    editor = "Goldberg, Yoav  and
      Kozareva, Zornitsa  and
      Zhang, Yue",
    booktitle = "Findings of the Association for Computational Linguistics: EMNLP 2022",
    month = dec,
    year = "2022",
    address = "Abu Dhabi, United Arab Emirates",
    publisher = "Association for Computational Linguistics",
    url = "https://aclanthology.org/2022.findings-emnlp.508/",
    doi = "10.18653/v1/2022.findings-emnlp.508",
    pages = "6826--6847"
}

@inproceedings{kim-etal-2025-evaluating,
    title = "Evaluating Language Models as Synthetic Data Generators",
    author = "Kim, Seungone  and
      Suk, Juyoung  and
      Yue, Xiang  and
      Viswanathan, Vijay  and
      Lee, Seongyun  and
      Wang, Yizhong  and
      Gashteovski, Kiril  and
      Lawrence, Carolin  and
      Welleck, Sean  and
      Neubig, Graham",
    editor = "Che, Wanxiang  and
      Nabende, Joyce  and
      Shutova, Ekaterina  and
      Pilehvar, Mohammad Taher",
    booktitle = "Proceedings of the 63rd Annual Meeting of the Association for Computational Linguistics (Volume 1: Long Papers)",
    month = jul,
    year = "2025",
    address = "Vienna, Austria",
    publisher = "Association for Computational Linguistics",
    url = "https://aclanthology.org/2025.acl-long.320/",
    doi = "10.18653/v1/2025.acl-long.320",
    pages = "6385--6403",
    ISBN = "979-8-89176-251-0"
}

@inproceedings{
wang2023selfconsistency,
title={Self-Consistency Improves Chain of Thought Reasoning in Language Models},
author={Xuezhi Wang and Jason Wei and Dale Schuurmans and Quoc V Le and Ed H. Chi and Sharan Narang and Aakanksha Chowdhery and Denny Zhou},
booktitle={The Eleventh International Conference on Learning Representations },
year={2023},
url={https://openreview.net/forum?id=1PL1NIMMrw}
}

@article{hu2022lora,
  title={Lora: Low-rank adaptation of large language models.},
  author={Hu, Edward J and Shen, Yelong and Wallis, Phillip and Allen-Zhu, Zeyuan and Li, Yuanzhi and Wang, Shean and Wang, Lu and Chen, Weizhu and others},
  journal={ICLR},
  volume={1},
  number={2},
  pages={3},
  year={2022}
}

@article{gandhi2023understanding,
  title={Understanding social reasoning in language models with language models},
  author={Gandhi, Kanishk and Fr{\"a}nken, Jan-Philipp and Gerstenberg, Tobias and Goodman, Noah},
  journal={Advances in Neural Information Processing Systems},
  volume={36},
  pages={13518--13529},
  year={2023}
}

@inproceedings{jeretic-etal-2020-natural,
    title = "Are Natural Language Inference Models {IMPPRESsive}? {L}earning {IMPlicature} and {PRESupposition}",
    author = "Jeretic, Paloma  and
      Warstadt, Alex  and
      Bhooshan, Suvrat  and
      Williams, Adina",
    editor = "Jurafsky, Dan  and
      Chai, Joyce  and
      Schluter, Natalie  and
      Tetreault, Joel",
    booktitle = "Proceedings of the 58th Annual Meeting of the Association for Computational Linguistics",
    month = jul,
    year = "2020",
    address = "Online",
    publisher = "Association for Computational Linguistics",
    url = "https://aclanthology.org/2020.acl-main.768/",
    doi = "10.18653/v1/2020.acl-main.768",
    pages = "8690--8705"
}

@article{shanahan2024simulacra,
  title={Simulacra as conscious exotica},
  author={Shanahan, Murray},
  journal={Inquiry},
  pages={1--29},
  year={2024},
  publisher={Taylor \& Francis}
}

@article{landau1988importance,
	title={The importance of shape in early lexical learning},
	author={Landau, Barbara and Smith, Linda B and Jones, Susan S},
	journal={Cognitive development},
	volume={3},
	number={3},
	pages={299--321},
	year={1988},
	publisher={Elsevier}
}

\newpage
\appendix
\section{Case study I: Referential expression generation}
\label{app:cs1}

This following details the implementation and evaluation of the models for pragmatic referential expression generation described in Section~\ref{section:utterance-production}, including the full prompts of the neural proposer and evaluator modules. 
First, we report details of the iterative model in Section~\ref{app:cs1-im}; then, we report details of the ablated non-iterative model in Section~\ref{sec:ablated-single-pass}, and of the LM-only baseline in Section~\ref{app:cs1-baseline}.

\subsection{Iterative Model}
\label{app:cs1-im}

An overview of the algorithm is described in the main text in Section~\ref{sec:cs1-iterative-model}, and a detailed pseudo-code for the algorithm is shown in Algorithm~\ref{alg:utteranceProductionLoopFlow}.
Here and in the following, the algorithm indicates the neural modules with \LM{} (UtteranceProposer and SemanticEvaluator)and the rule-based modules with \RULE{} (ContrastivitySelector and InfoMaxSelector). 
Each module is described in detail in the following, based on evaluation results that we also reported in previous work \citep{tsvilodub2024cognitive}.

\begin{algorithm}[t!]
\caption{Iterative model for contrastive utterance generation}
\begin{algorithmic}[1]
\Procedure{Generate}{$s^{*}$, $D$, $n$} 
\Comment{Takes a target state and a set of distractors.}
    \State $\text{partialUtt} \gets []$
    \While{$i \leq 5$}
        \For{$u' \in \text{partialUtt}$} 
        \State $B \gets \LM{} \text{\proposer{UtteranceProposer}}(s^{*}, n, u')$
        \Comment{Sample (initial) set of utterances $B$}
        \EndFor
        \State $C_{\text{new}} \gets []$
        \Comment Stores the contrastivity of utterances
        \State $T \gets []$
        \Comment Stores the interpretation of utterances
        \For{$b_u \in B$}
        \Comment{Loop over best utterances so far $B$}
            \State \textbf{append } $T_u \gets \LM{} \text{\evaluator{SemEval}}(\{s^{*}\} \cup D, b_u)$ \textbf{to} $T$ 
            \Comment Evaluate truth of $b_u$ on each state
            \State $\text{\textbf{append }} \RULE{} \text{\reasoner{ConstrastivitySelector}}(T_u)$ \textbf{to} $C_{\text{new}}$
            \Comment Evaluate contrastivity of $b_u$
        \EndFor
        \State $\text{indices} \gets C_{\text{new}} = \max (C_{\text{new}})$
        \Comment Mask where contrastivity is maximal
        \State $B^* \gets B[\text{indices}]$
        \Comment Get most contrastive utterances\ldots
        \State $T^* \gets T[\text{indices}]$
        \Comment \ldots and their interpretations
        \If{$\max (C_{\text{new}}) = 1$} 
        \Comment There is at least one contrastive utterance
            \State $u \gets \RULE{} \text{\reasoner{InfoMaxSelector}}(T^*, B^*)$
            \Comment Choose among contrastive utterances
            \State \textbf{return} $u$
        \EndIf 
        \For{$b \in B^*$}
            \State $\text{partialUtt} \gets B^*$
        \EndFor
        \State $B \gets [], \; i++$
    \EndWhile
\EndProcedure
\end{algorithmic}
\label{alg:utteranceProductionLoopFlow}
\end{algorithm}

\subsubsection{UtteranceProposer}
\label{app:im-utt-proposer}
The first kind of module introduced in the SAGE framework are proposers which generate alterantives for reasoning.
In this first case study, we assessed an LM-based \textit{UtteranceProposer} that sampled candidate referential expressions identifying a target state $s^*$ among distractors $D$.
The states were drawn from the structured A3DS dataset and were represented in terms of feature-value descriptions for simple features like color and shape, so that the expected space of plausible utterances for a state was reasonably constrained.

The proposer sampled $n$ candidate utterances on each iteration of the model. 
On the first iteration of the model, there were no previous utterances yet. We prompted the LM to generate initial utterances which should only mention a single feature of the target. We motivate this constraint in the prompt as approximating a production cost pressure, which is present for humans and usually included in cognitive models, but absent in LMs.
During explorations, we also observed that including an instruction to only use details provided in the state description helped reduce possible hallucinations in the candidate utterances. 
We used the following prompt for the UtteranceProposer in first iteration of the IM:
\begin{prompt}
    \noindent You will be given a target description. Please produce \{num\_samples\} sentence(s) that only mention one detail from the target description. The produced sentences should include exclusively content mentioned in the target. Please provide the sentences in a bullet list format.  \\

    Target: \{target state\}\\
    Sentences:
\end{prompt}

\texttt{num\_samples} was set to either four or eight in our simulation.

Qualitative analyses showed that the proposer generated diverse and expected candidate utterances. 
For instance, for the target state ``The floor is blue, the wall is blue, the blue large pill is in the middle.'', a candidate utterance was ``The floor is blue''. 
However, we also observed that whenever the object on the state was mentioned, there was a tendency to mention both the shape of the object and one additional feature (e.g., ``The pill is blue.'', rather than ``The object is a pill.''), which is unsurprising given that human language use is also biased towards focusing on shapes of objects \citep{landau1988importance}. 

For further iterations of the IM, the prompt was adapted to include the most contrastive previously generated utterance \texttt{\{partial\_description\}}:
\begin{prompt}
    \noindent Your task is to produce some sentences. Each sentence should repeat the information in "\{partial\_description\}", but add one more detail taken from "\{full\_description\}" Do not make up any new detail. \\

    Please produce \{num\_samples\} sentence(s) in a bullet list format. Be very concise!\\

    Sentences:
\end{prompt}
The model was prompted to increment the utterances by one feature to keep the approximation of a production cost and prevent unnatural exhaustive enumeration of all features of the target right away.

Qualitative analyses showed that the proposer extended the initial utterances in well-formed manner, e.g., producing the following utterance for the state and initial utterance above: ``The floor and wall are both blue.'' (examples are chosen at random from recorded module outputs).
However, manual inspection also showed that some samples of the LM did not follow the instruction to only add one new feature, so that another candidate from the second iteration could be, e.g., ``The blue large pill is in the middle of the blue floor.''
On the other hand, sometimes no new features at all were included, e.g., producing a candidate like ``The green wall is in the room.'', given the initial utterance ``The wall is green.''.
Finally, at late iterations of the model, the proposals sometimes contained hallucinations about additional details of the state, for instance like the last part of the following utterance: ``The blue small block is in the right corner on the blue floor and surrounded by the pink wall, which has a small dent near the bottom.'', or were somewhat figurative, like ``The red floor and walls create a striking setting for the large purple cylinder that takes center stage in the room.''.
We expect such issues to reduce as LMs' instruction-following abilities improve. 
Additionally, the semantic evaluation, at least in principle, provides a mechanism for filtering out such candidates when their truth value with respect to $s^*$ is evaluated.

\subsubsection{SemanticEvaluator}
\label{app:im-sem-eval}
The second kind of module introduced in the SAGE framework are evaluators.
In this case study, an evaluator was used to assess literal semantics of utterances.
That is, the purpose of the SemanticEvaluator was to check whether candidate utterances were true of each of the states (target and distractors). 
Semantic evaluation was applied to each utterance-state combination to determine whether any utterance among the available candidates was fully disambiguating, i.e., true of the target and false of all distractors in the context.
Because this contrastivity evaluation determined, in the IM, whether the iterative refinement of utterances should be continued, and in the SP model, which utterance was the best referential expression, we conducted separate assessments of the SemanticEvaluator and adjusted its prompt based on evaluation results.

The task was cast as checking the utterance semantics, i.e., if the utterance is literally true of a state. 
This task is essentially a logical entailment problem, where the module needs to determine whether a candidate utterance is true, given a (true) state.
To exploit the model's knowledge of intuitive language use, we formulated the final prompt for determining the semantic value that was used in the reported experiments as an \emph{intuitive entailment} task. 
Given the following prompt, the LM was instructed to return `yes'/`no' and output a chain of thought, given a state description and utterance pair: 

\begin{prompt}
Consider the following sentence: \{state\}

Does the following statement provide exclusively information also contained in the sentence above: \{utterance\} 

Explain your answer step by step. 

Importantly, the last line of your answer should exclusively contain "yes" or "no", and nothing else. 

Here the the structure of the answer:

"""

[step-by-step explanation,

possibly over multiple lines]

[empty line]

[yes/no]

"""

Your answer: 
\end{prompt}

The generated response was processed with a regular expression to extract `yes'/`no' and convert these to 1, 0, respectively. 

During development, variations of this prompt were tested. For instance, we added a one-shot chain of thought prompt exemplifying the reasoning; we also explored various terms to refer to the state and the previous utterance (``statement'', ``sentence'', ``fact''), but no significant performance gains were observed. 
We also explored alternative prompts, e.g., asking for ``logical compatibility'' of the state and utterance, whether the utterance is ``true'', whether there are contradictions between the state and utterance, any new information is present in the utterance, but the prompt above provided the most reliable results.

To analyze and optimize the performance of this module in isolation, we evaluated the SemanticEvaluator on two groups of tests. 
First, we used tasks from the SuperGLUE and SNLI benchmarks \citep{bowman-etal-2015-large, wang2019superglue}. 
One test set was based on five samples from each of the ``axb'', ``axg'', ``copa'', ``rte'' tasks within the SuperGLUE benchmark, and on five entailment and contradiction datapoints each from the SNLI benchmark. 
Pairs of states and utterances for passing them to the prompt were constructed, and the ground truth semantic value was derived from the dataset annotations. 
For SNLI, the sentence 1 was used as the state and sentence 2 as the utterance. 
These tests mostly contained naturalistic sentences and strongly focused on testing natural language inference (NLI). 
We used 39 NLI test pairs for assessing the SemanticEvaluator.
Second, we used tests containing manually constructed example sentences closely matching the A3DS dataset that we used in the main simulation in phrasing and content, checking the prompt with respect to robustness for synonyms, syntactic modification, and interpretation of quantifiers. 
These were intended to broaden the set of tests and include tests matching the reference game setting more closely. 
We constructed 12 test sentences. 

All evaluations were conducted with manual processing of the LM outputs.
The average accuracy of the final SemanticEvaluator with the prompt reported above on these two test sets was 0.82.
Inspecting cases where the module failed throughout development, we identified that the LM with this prompt still might be susceptible to inconsistencies between the chain of thought and the final answer of the LM; 
for instance, if the chain of thought seemed to address the question ``Is there any new information present in the utterance?'', it was more likely to lead the model to the final answer ``no'', although the reasoning suggested ``yes''. 
Additionally, we observed that the regular expression based processing of the final answers of the modules was a step that is susceptible to LMs' failures to follow formatting instructions precisely.
We note that since the time of implementation of the case studies, new methods for retrieving LM generations that fit a specific format and contain specific values have been developed \citep[e.g.,][]{Moskal2024}; such tools could be of great use for future implementations of such modules. 
Additionally, we speculate that fine-tuning a small LM as a reusable semantic evaluator might be a promising direction to improve the performance of the module, since we observed that tasks involving literal semantics turn out to be more challenging for LMs than tasks required for other types of modules (see Section~\ref{sec:discussion-challenges} for detailed discussion).

\subsubsection{ContrastivitySelector}
The results of the semantic evaluation were passed to the next module which evaluated the contrastivity of the utterances, the ContrastivitySelector.
This module is a rule-based module which takes a matrix of truth values $T$ computed by the SemanticEvaluator as input (i.e., a matrix of shape state $\times$ utterance), and checks the proportion of distractors of which each available utterance $u_i$ is \emph{false}. 
That is, it computed
$C_i = \frac{\# \text{distractors for which } [\![u_i]\!](d) = 0}{\# \text{distractors}}$, resulting in a list of contrastivity values for the utterances.
If there was at least one utterance which is fully contrastive (i.e., $max(C) = 1$), the loop was terminated and this utterance was returned. 
If no utterance was fully contrastive, the utterances with the highest contrastivity were selected and passed to the next UtteranceProposer in the next iteration of the model.

\subsubsection{InfoMaxSelector}
\label{app:rsa-reasoner}
When the maximal number of iterations was reached by the IM and no fully contrastive utterance was detected, the contrastivity evaluation results were passed to a simple rule-based  information maximization (InfoMaxSelector) module selecting the best utterance among the avaialble ones. 
This module was also used in the ablated single-pass model.
The InfoMaxSelector takes as input contrastivity values (IM) or derived them from the semantic truth values (SP model), and returned the utterance with the highest contrastivity value:
$u^* = argmax (C)$

We note that this module could be extended to return a distribution over utterances, akin to the pragmatic speaker $S_1$ in RSA models \citep[e.g.,][]{frank2012predicting}:

\begin{align}
P_{S_1}(u \mid s) 
&\propto \exp(\alpha \; (\log \; L_0(s \mid u) - \text{cost})) \\
P_{L_0}(s\mid u) 
&\propto [\![u]\!](s)
\end{align}

The name of the module ``InfoMax'' is couched in the idea that the pragmatic speaker as defined above prefers utterances that are \textit{informative} for a literal listener, i.e., maximally reduces her surprisal. 

\subsection{Ablated Single-Pass Model}
\label{sec:ablated-single-pass}

\begin{algorithm}[t!]
	\caption{Ablated single-pass model for contrastive utterance generation}
	\label{alg:utteranceProductionSimpleFlow}
	\begin{algorithmic}[1]
		\Procedure{Generate}{$s^{*}$, $D$, $n$}
		\Comment{$s^{*}$ is a target state and $D$ a set of distractors}
		\State $U_{s^{*}} \gets \LM{} \text{\proposer{UtteranceProposer}}(s^{*}, n)$
		\Comment{Sample $n$ descriptions of target $s^{*}$}
		\State $T \gets \LM{} \text{\evaluator{SemanticEvaluator}}( \{s^{*}\} \cup D, U_{s^{*}})$
		\Comment{Truth of each utterance for each state}
		\State $u^* \gets \RULE{}\text{\reasoner{InfoMaxSelector}}(T, U_{s^{*}})$
		\Comment Select most informative utterance
		\State \textbf{return} $u^*$
		\EndProcedure
	\end{algorithmic}
\end{algorithm}

We compared the IM to an ablated model, the single-pass (SP) model, which did not iterate over candidate utteranes.
As in the IM, the input to the SP model was a list of state descriptions, including the target state $s^*$ and one or more distractors $D$.
The model proceeded in three steps, described in pseudo-code in Algorithm~\ref{alg:utteranceProductionSimpleFlow}.

First, an UtterancesProposer module generated ten candidate utterances for the target state based on the target state description.
Second, the SemanticEvaluator module determined the truth value of each candidate utterance for each state (target and distractors).
Lastly, the InfoMaxSelector module selected and returned the most contrastive utterance.
The functionality of the SemanticEvalutor and the InfoMaxSelector are identical to the IM.

\subsubsection{UtterancesProposer}
\label{sec:utterancesproposer}

The prompt used for generating the utterance proposals did not include the restriction to mentioning a single feature, i.e., it did not include the production cost approximation.
The following prompt was used: 
\begin{prompt}
    You will be given a target description. Please produce \{num\_samples\} sentence(s) based on the target that leave out some part of the description but are still well-formed. The reduced sentences should include exclusively content already mentioned in Target. Please provide the sentences in a bullet list format.
\end{prompt}

\texttt{\{num\_samples\}} was set to ten across simulations.

Qualitative evaluations confirmed that the candidate utterances were well-formed, and differed in their granularity and contents. For instance, given the state ``The floor is purple, the wall is green, the red small block is in the left corner.'', some candidate utterances were: ``The wall is green and the floor is a different color.'', ``There is a small red block in the left corner.'',  ``The left corner has a red block.'', ``The wall is not purple, but the floor is.''.

\subsection{LM-only Baseline}
\label{app:cs1-baseline}

We used the following one-shot prompt for the LM-only baseline for generating referential expressions: 

\begin{prompt}
\noindent You will be given a target state and one or more distractors. \\
Your task is to describe the target state in natural language in a way that distinguishes it from the distractors. \\
Try to be as concise as possible. You do not need to list all the features of the target state. \\
Please think step by step, motivating why you decide to mention some features.\\

Here is an example of a good answer.\\

Target state:\\
- The floor is purple, the wall is green, the red small block is in the left corner.\\

Distractors:\\
- The floor is red, the wall is green, the red small block is in the middle.\\

Your answer:\\
One difference between the target and the distractor is the color. This difference is enough to distinguish between them.
Utterance: "The target state has a purple floor".\\
Now the real input.
\end{prompt}

Given this prompt, the baseline produced utterances varying in the level of granularity, although often following the syntactic pattern of the one-shot example, e.g.: ``The target state has an orange medium-sized block in the middle.'', ``In the target state, there is a small blue cylinder in the left corner on a yellow floor and an orange wall.''.

\section{Case study II: M-implicatures}
\label{sec:appendix-modules-MImplicatures}

This section reports details of the models from the case study on the inteprretation of M-implicatures described in Section~\ref{section:m-impl}. 
The SAGE model based on ideas from optimality theory --- the markedness-blocking (MB) model --- is described in the main text in Section~\ref{sec:cs2-model}; detailed pseudo-code for the MB model is presented in Algorithm~\ref{alg:flowSimpleMImplicatureInterpretation}.
The implementation of each module is described in the following.
In particular, human evaluation of the candidate utterances provided by the neural UtteranceProposer are reported in Section~\ref{app:m-implicature-utterance-proposer}.

\begin{algorithm}[t!]
\caption{Markedness-blocking algorithm for M-implicature interpretation}\label{alg:flowSimpleMImplicatureInterpretation}
\begin{algorithmic}[1]
\Procedure{Interpret}{$u^*$, states, context, $n$}
    \Comment{Takes utterance $u^*$, possible states as input}
    \State $\text{complexity\_record} \gets []$
    \Comment{Initialize list to record relative complexity}
    \For{$s \in$ states}
        \State $\text{alt}_{s} \gets \LM{} \text{\proposer{UtteranceProposer}}(s, n, \text{context})$
        \Comment{Sample $n$ utterances that naturally describe $s$}
        \For{$a \in \text{alt}_{s}$}
            \Comment{Compare $u^*$ to the alternatives set}
            \State $c \gets \LM{}\text{\evaluator{DifferentialComplexityEvaluator}}(u^*,a)$
            \State \textbf{append} $c$ \textbf{to} complexity\_record
        \EndFor
    \EndFor
    \State priors $\gets []$
    \Comment{Initialize list of priors}
    \For{$s \in$ states}
        \State \textbf{append} \LM{} $\text{\evaluator{PriorEvaluator}}(s)$ \textbf{to} priors
    \EndFor
    \State state $\gets \RULE{}\text{\reasoner{UnblockedSelector}}(\text{priors}, u^*, \text{complexity record})$
    \State \textbf{return} state
\EndProcedure
\end{algorithmic}
\end{algorithm}

\subsection{UtteranceProposer}
\label{app:m-implicature-utterance-proposer}
The first LM module of the MB model is the UtteranceProposer module which generated alternative utterances $u'$, conditioned on a given state $s$, that a speaker could normally use to talk about the state. 
Crucially, the module did not have access to the observed trigger utterance $u$.
We used the following prompt to elicit the alternative utterances:

\begin{prompt}
    Image a person wants to communicate the following information: "\{conditioned\_on\}"

Generate \{n\_proposals\} concise natural expressions for communicating the information. Please provide your response in a bullet list format.
\end{prompt}
where \texttt{conditioned\_on} was a possible interpretation (i.e., typical or atypical state). 
The number of alternatives \texttt{n\_proposals} was set to three throughout the reported simulations.

\begin{figure*}[t!]
    \centering
    \includegraphics[width=0.7\linewidth]{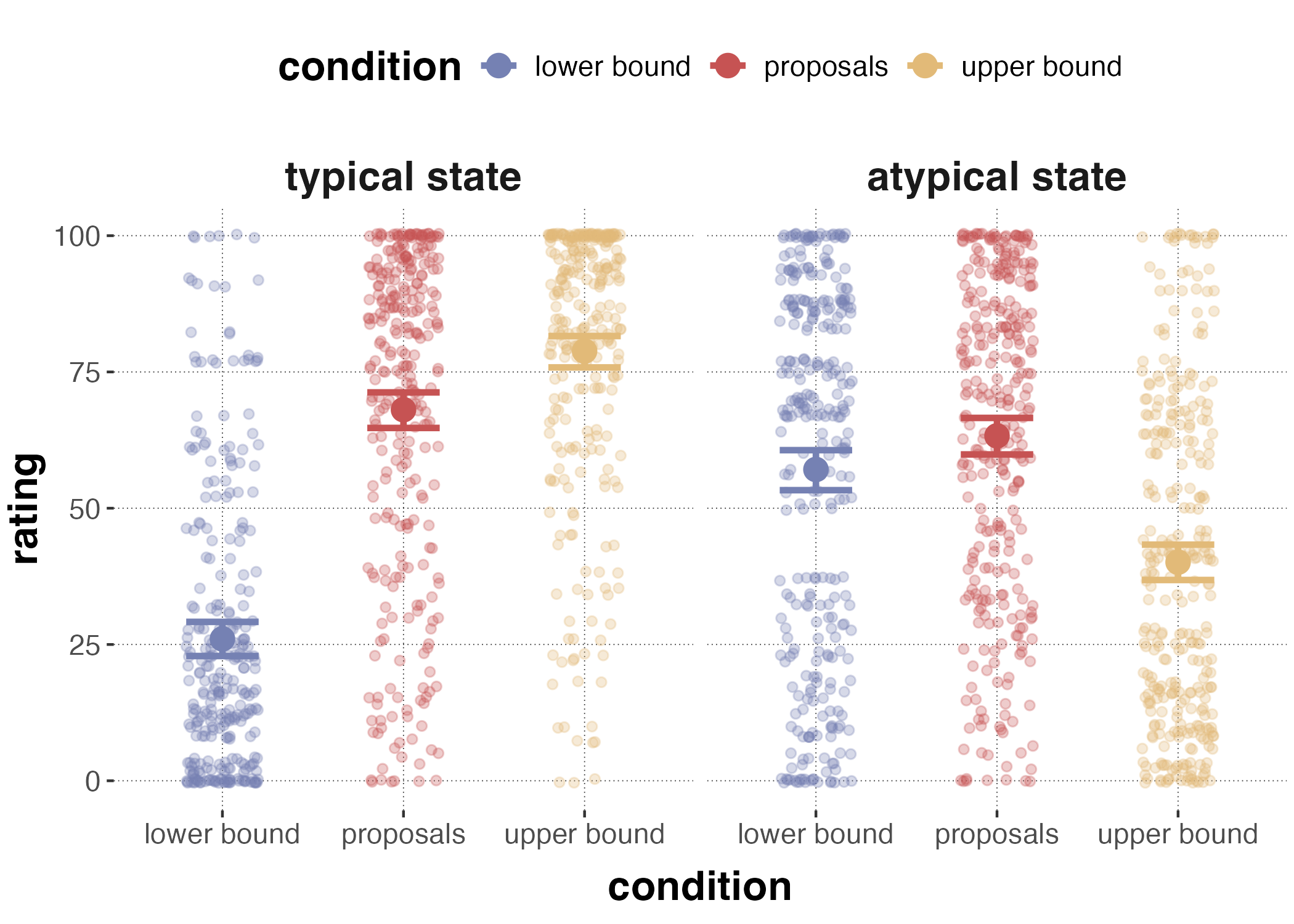}
    \caption{Human ratings (y-axis) of the utterance proposals against human-constructed reference utterances (x-axis) for different interpretation types (facets). Smaller points show individual ratings, large points indicate means, error bars show 95\% bootstrapped CIs.
    \label{fig:cs2-evals-results}}
\end{figure*}

\subsubsection{Human module evaluation} 
This case study builds on more theoretically loaded assumptions about what counts as viable utterance alternatives than in the previous case study, providing more challenging requirements for the LM-based UtteranceProposer. 
Specifically, the alternative utterances have to not only truthfully describe the given state, but we also expect them to be more complex when conditioned on an atypical state, and more natural-sounding given a typical state.
To assess to which extent the LM was able to satisfy there expectations, we conducted human experimental evaluations of the utterance proposer.
The live experiment can be viewed under: \url{https://magpie-ea.github.io/magpie3-sage-module-evals/proposer_evals/cs2_utterances/}, the code and data for the experiment and the analyses can be found under: \url{https://github.com/magpie-ea/magpie3-sage-module-evals/tree/main}.

In this experiment, we triangulated the quality of the alternative utterances generated by the LM proposer against human-written reference utterances, for which we used the marked and unmarked utterances in the dataset (generated by the authors). 
Specifically, for ten randomly selected items, we randomly sampled three proposals generated by the LM in each condition (i.e., given the typical, and the atypical state of the item). 
If the LM proposals are natural alternative utterances for a given state, we expect that they would be rated as good as the human-written matching utterances (e.g., the marked utterance given the atypical state; no credible difference in ratings); we call these utterances the  or upper bound. And we expect that the proposals would be rated higher than the unmatching, or lower bound, human utterances (e.g., the marked utterance given the typical state; credible difference in ratings).
On each trial, participants ($N=34$) read the state description, followed by the three LM proposals, the upper bound and the lower bound utterances, presented in randomly shuffled order on the screen.
Participants were asked to rate how natural they think the presented sentences are for talking about the given situation on a slider ranging from ``terrible'' to ``natural'' (mapped to 0--100). 
Each participant first saw an example trial where the described state was atypical, and the sliders were preset mid to high ratings for the proposals, a lower rating for an unmarked utterance, and a 0 rating for a factually false utterance. 
Then, each participant completed six main trials (three per state type) and one attention checking trial, presented in randomized order.
The attention check visually matched the main trials, and the context contained explicit instructions where to move all sliders (e.g., all to 0).\footnote{29.5\% of the participants failed the attention check by providing at least one rating that was off by more than 6 points from the instructed target rating. 
We ran all analyses reported below excluding these participants and observed fully qualitatively matching results. Therefore, all analyses below are conducted with the full sample of participants.}
At the beginning of the experiment, participants were told that they will help assess the quality of utterances, some of which were generated by language models.
The experiment took six minutes on average, and participants were reimbursed with \pounds1.00. 
English native speaking participants based in the US and UK with at least a 95\% approval rate and at least five previously completed studies were recruited through the crowdsourcing platform Prolific.

The results are shown in Figure~\ref{fig:cs2-evals-results}.
We analyzed the results using a Bayesian linear regression model, regressing the ratings against an intercept, a main effect of utterance type (three levels: LM proposals, human lower bound, human upper bound), a main effect of state type (typical or atypical state), their interaction, by-subject and by-item random intercepts and random slope effects of utterance type, state and their interaction.\footnote{The model in R syntax: \texttt{rating $\sim$ utterance\_type * state\_type + (1 + utterance\_type * state\_type | subject) + (1 + utterance\_type * state\_type | item)}.}
The contrasts were sum-coded.
We report the posterior means of the relevant regression coefficients or their comparisons and 95\% credible intervals. 
As reported in the main text, across the different state types, the proposer-generated alternatives were rated equally natural as the human upper bound utterances (i.e., no credible difference between the upper bound and the proposal ratings, $\beta =-4.26[-18.54,9.95]$), and as credibly more natural than the lower bound ($\beta=-23.75 [-35.66,-11.78]$).
In the typical state condition, the proposals were rated equally to the upper bound, with a tendency towards higher ratings ($\beta=13.70 [-4.73,31.82]$), and credibly higher than the lower bound ($\beta=-40.23 [-53.46,-26.71]$).
In the atypical state condition, the proposal were rated higher than the upper bound ($\beta=-22.22	[-41.76	,-2.32]$) and equal to the lower bound references ($\beta=-7.27	[-24.65, 11.14]$).
Results in both conditions indicate that our requirement for the proposals to at least match the naturalness of the upper bound reference was satisfied. 
In fact, in the atypical state condition, both the proposals and the lower bound reference were rated visibly higher than the upper bound reference, which we take to indicate that in case of an atypical situation, the unmarked utterance is taken to be still an acceptable expression for talking about the state. 
Taken together, these results confirm that the LM proposer generated viable utterance candidates that satisfied the theoretically predicted differences in naturalness, depending on the typicality of the state they described.

\subsection{DifferentialComplexityEvaluator}
\label{app:diff-complexity-eval}
Another important module in the MB model was the neural DifferentialComplexityEvaluator for comparing the complexity of the marked or unmarked input utterance to alternatives generated by the UtteranceProposer.
The role of the evaluator was to capture whether this open-endedly sampled set contains alternatives that are intuitively more common, natural expressions than the observed utterance.

We used the following prompt for this evaluator:

\begin{prompt}
    \noindent Your task is to compare the complexity of the following two sentences and what they describe.\\
Sentence 1: \{target\}\\
Sentence 2: \{alternative\}\\

Reason step by step.\\
Give your answer as a single number corresponding to the sentence which is more complex (1 for sentence 1 and 2 for sentence 2). If sentences are roughly equally complex, write 0.\\

Importantly, the last line of your answer should exclusively contain 0, 1 or 2, and nothing else. \\

Here the the structure of the answer:\\
"""\\
{[step-by-step explanation, \\
possibly over multiple lines]}\\
{[empty line]}\\
{[0/1/2]}\\
"""\\

Your answer: 
\end{prompt}

The performance of the model crucially depended on the output of this evaluator because it determined whether a given utterance-state pair was blocked by more natural alternatives or not.
Therefore, during development, different versions of the prompt were tested.
The prompt was tested with and without the chain of thought instruction, as well as with few-shot prompting or phrased as a \textit{wh}-question (i.e., ``Which of the following two sentences and what they describe is more complex?'').
Additionally, other formulations than ``complexity'' were explored.
We found that especially when phrasing the task as a \textit{wh}-question, the LM showed a strong tendency towards returning a specific sentence, even when they were very similar and a 0 should be returned.
We found the prompt reported above to be the most robust phrasing, at least for the model and materials in our experiments.

\subsection{PriorEvaluator}
\label{app:prior-eval}
The third LM-based component in the MB model was the PriorEvaluator that ranked the likelihood (or, typicality) of the states (i.e., available interpretations for the utterance). 
We used the following prompt:

\begin{prompt}
    \noindent Consider these two situations: \\
Situation1: \{state1\} \\
Situation 2: \{state2\}\\
Your task is to rank these situations from most typical to least typical. In rating the typicality, focus on the content that is being described rather than the precise phrasing of the description.
Output the ranked situations by ordering the words "situation 1", "situation 2" according to the ranking. The last like of your answer should contain only these two expressions. \\
Reason step by step.\\

Here the the structure of the answer:\\

"""\\
{[step-by-step explanation, \\
possibly over multiple lines]}\\
{[empty line]}\\
{[situations]}\\
"""\\

Your answer:
\end{prompt}

This evaluator was less sensitive to variations of the prompting that were explored during development, suggesting that an evaluator performing more intuitive tasks (like judging typicality of states) may be more robust than when performing more complex or theoretically-driven evaluations like in the SemanticEvaluator or the DifferentialComplexityEvaluator.
This was confirmed though manual evaluations of 62 samples, revealing a high accuracy of 0.925 for this module.

\subsection{UnblockedSelector}
Finally, the UnblockedSelector is a symbolic module of the MB model which instantiated the idea of the blocking of interpretations by simpler available alternatives based on the results of the differential complexity evaluation (Section \ref{app:diff-complexity-eval}) and the prior evaluation (Section~\ref{app:prior-eval}). 
If alternative utterances were preferred over the target (i.e., evaluated as less complex than the target), the target utterance and state pair under consideration was blocked.
The unblocked selector selected the most likely \textit{unblocked} interpretation for the given target utterance (i.e., the utterance for which no simpler alternatives were sampled, together with the most likely state). 
The likelihood was determined by the PriorEvaluator.
Runs of the MB where no unblocked state-utterance pairs were available were discarded from analysis.




\section{Case study III: Implicatures}
\label{appendix:modulesCaseStudyIII}

This following details the implementation and evaluation of the Gricean AE model, its ablations and baseline for implicature interpretation described in Section~\ref{sec:caseStudy-III}, including the full prompts of the neural proposer and evaluator modules. 
First, we report details of the AE model and different assumption sets about the speaker's behavior in Section~\ref{app:sec:ae-model}, including results of human evaluations of one neural proposer and evaluator module, respectively, in Sections~\ref{app:assumption-eval},~\ref{app:implicatures-interpretation-proposer}.
We report details of the experiment where we collected human data for the novel test dataset designed in this case study in Section~\ref{app:cs3-human-expt}.
Finally, we provide details of simulations and analyses of the AE model instantiated with different LM backbones in Sections~\ref{app:sec:lm-configs}--\ref{app:cs3-human-llh}.

\subsection{Assumption-Evaluation Model}
\label{app:sec:ae-model}
The assumption-evaluation (AE) model is described in Section~\ref{sec:models-CS-IIIa} in the main text; detailed pseudo-code for the AE model is in Algorithm~\ref{alg:interpretationAssumptionsFlow}.
Below, we provide details of each module.

\begin{algorithm}[t!]
  \caption{Assumption-evaluation model for implicature interpretation}
  \label{alg:interpretationAssumptionsFlow}
  \begin{algorithmic}[1]
    \Procedure{Interpret}{trigger utterance $u$, context, assumptions $A$, forced choice options $O$, $n$}
      \State $\text{violated\_assumptions} \gets []$
      \Comment{Initialize list of violated assumptions}
      \For{$a \in A$}
        \State $a_{v} \gets \LM{} \text{\evaluator{AssumptionEvaluator}}(a, u, \text{context}, O)$
        \Comment{Evaluate if $a$ holds of $u$ in context}
        \State if $a$ violated, \textbf{append} $a_{v}$ \textbf{to} violated\_assumptions
      \EndFor
      \State plausibility\_values $\gets []$
      \Comment{Initialize list of plausibility values}
      \For{$a_v \in$ violated\_assumptions}
        \State \textbf{append} \LM{} $\text{\evaluator{PlausibilityEvaluator}}([u; a_v], a_v, \text{context})$ \textbf{to} plausibility\_values
      \EndFor
      \State $a_c \gets$ max(plausibility\_values)
      \Comment{Select crucial violated assumption}
      \State $\text{int} \gets \LM{}\text{\proposer{InterpretationProposer}}(u, \text{context}, a_c, n)$
      \Comment{Sample interpretations given violation}
      \State most\_likely\_options $\gets []$
      \For{($O, i) \in$ product($O$, int)}
        \Comment{Compare each interpretation to forced choice options}
        \State $o \gets \LM{}\text{\evaluator{InterpretationMappingEvaluator}}(O, i)$
        \State \textbf{append} $o$ \textbf{to} most\_likely\_options
      \EndFor
      \State interpretation $\gets \RULE{}\text{\reasoner{MajoritySelector}}(\text{most\_likely\_options})$
      \State \textbf{return} interpretation
    \EndProcedure
  \end{algorithmic}
\end{algorithm}

\subsubsection{AssumptionEvaluator}
\label{app:assumption-eval}
The first module processing the input utterance and context in the AE model is the AssumptionEvaluator.
This neural module is used to evaluate which assumption(s) about a pragmatic, cooperative speaker might be violated by the observed trigger utterance in the given context. 
More specifically, in the main Gricean AE model the evalutor determined whether the utterance adhered to different conversational maxims, or, assumptions, that usually hold of cooperative pragmatic speakers. 
According to the Gricean tradition in pragmatics \citep{grice1975logic}, if a listener identifies that a particular assumption is violated, or, flouted, she will be able to derive the speaker's intended meaning that goes beyond the literal meaning of the utterance through rationalizing the observed behavior. 
If no violations are identified, the utterance would be interpreted literally.
Therefore, information about violated assumptions is then passed to the subsequent modules to generate rationalizations (i.e., interpretations) of the input utterance.

The assumptions were represented as a list of statements about a speaker, each evaluated through a separate call to the AssumptionEvaluator.
The main AE model evaluated a set of Gricean assumptions, presented in Table~\ref{app:tab:cs3-gricean-assumptions}. 
Additionally, the main Gricean AE model was compared to AE models instantiated with different sets of assumptions: 
intuitive formulations of the Gricean assumptions (see Table~\ref{app:tab:cs3-intuitive-assumptions}), and lexical assumptions, which fomulate generally irrelevant facts about the speaker but mention a keyword related to the Gricean maxims (Table~\ref{app:tab:cs3-lexical-assumptions}).
The tables contain the assumption formulation which was passed as input to the AssumptionEvaluator, and the respective phrase describing the violation of each assumption if it was identified, which was passed to the subsequent PlausibilityEvaluator and InterpretationProposer.

Identifying flouted maxims is a challenging task; pragmatic utterances may be compatible with different maxims being flouted, and the recognition of flouting has mostly been formulated as part of verbal theories.
This evaluator module provides a first attempt at implementing open-ending recognition of the maxim flouting through simple prompt-based evaluation of maxims provided as sentences mentioning individual aspects. 
In light of the observed challenges connected to using LM evaluators in the previous case studies, we adapted the prompt of this module based on manual assessments of the outputs during development, and critically assessed its performance under the final prompt against human data (see below). 

We used the following prompt for the AssumptionEvaluator in the reported simulations with all sets of assumptions:

\begin{prompt}
    \noindent Context: \{context\} "\{trigger\}" \\
\noindent Normally, you would make the following assumption about \{speaker\}'s utterance: \{assumption\}. \\

\noindent Should we doubt that the assumption is true in this context? The last line of your response should contain be a single word and nothing else: "yes" if there are strong reasons to doubt the assumption and "no" if there are no strong reasons to doubt the assumption. \\

\noindent Reason step by step. \\

\noindent Here the the structure of the answer:\\

\noindent"""\\
\noindent [brief step-by-step explanation]\\
\noindent [empty line]\\
\noindent [yes/no] \\
\noindent """ \\

\noindent Your answer:
\end{prompt}

During development, the prompt was optimized for plausibility of the results and good instruction-following regarding the formatting of the outputs.
Manual inspections of the outputs revealed sensitivity towards the exact formulation of the prompt: for instance, the consistency of the chain of thought and the final answer was sensitive to the polarity of the main question (``doubt''~vs.~``is the following assumption true'').
These observations are in line with our observations for LM evaluators in the previous case studies, suggesting that more elaborate strategies than simple zero-shot prompting may be required for incorporating LMs as reliable evaluators in SAGE-style models.

 \begin{table}[h!]
	\centering
	\begin{tabular}{p{1.5cm}p{1.0cm}p{5.25cm}p{5.25cm}}
		\toprule
		maxim
		& submax.
		& assumption formulation
		& violation formulation
		\\ \midrule
		quantity 
		& 1
		&\footnotesize{the answer is as informative as is required by the context}
		&\footnotesize{the answer is less informative than is required by the context}
		\\
		\addlinespace[0.3em]
		& 2
		&\footnotesize{the answer is no more informative than is required by the context}
		&\footnotesize{the answer is more informative than is required by the context}
		\\
		\addlinespace[0.6em]
		quality 
		& 1
		&\footnotesize{\{speaker\} believes the answer to be true}
		&\footnotesize{\{speaker\} does not believe that the answer is true}
		\\ 
		\addlinespace[0.3em]
		& 2
		&\footnotesize{\{speaker\} likely has enough evidence for the answer}
		&\footnotesize{\{speaker\} likely does not have enough evidence for the answer}
		\\
		\addlinespace[0.6em]
		relevance 
		& 1
		&\footnotesize{the answer is relevant in context}
		&\footnotesize{the answer is not relevant in context}
		\\
		\addlinespace[0.6em]
		manner 
		& 1
		&\footnotesize{the answer avoids language that is difficult to understand}
		&\footnotesize{the answer uses language that is difficult to understand}
		\\
		\addlinespace[0.3em]
		& 2
		&\footnotesize{the answer is unambiguous in the context}
		&\footnotesize{the answer is intentionally ambiguous in the context}
		\\
		\addlinespace[0.3em]
		& 3
		&\footnotesize{the answer is not too prolix for the context}
		&\footnotesize{the answer is unnecessarily prolix for the context}
		\\
		\addlinespace[0.3em]
		& 4
		&\footnotesize{the answer provides information in an order that makes sense}
		&\footnotesize{the answer provides information in an order that does not make sense}
		\\ 
		\bottomrule

	\end{tabular}
\caption[Gricean Assumptions]{Set of ``Gricean assumptions'' used by the AssumptionEvaluator the main AE model for implicature interpretation.  \label{app:tab:cs3-gricean-assumptions}}
	\end{table}


 \begin{table}[h!]
	 \centering
	\begin{tabular}{p{3.5cm}p{5.25cm}p{5.25cm}}
		\toprule
		name
		& assumption formulation
		& violation formulation
		\\ \midrule
		evidence
		&\footnotesize{the answer is adequately supported by evidence}
		&\footnotesize{\{speaker\}'s answer is not adequately supported by evidence}
		\\
		\addlinespace[0.5em]
		naturalness
		&\footnotesize{the answer is a precise and natural way of expression}
		&\footnotesize{\{speaker\}'s answer is not a precise and natural way of expression}
		\\
		\addlinespace[0.5em]
		objectivity
		&\footnotesize{the answer is only conveying objective information about the world; it does not convey subjective information like \{speaker\}'s evaluation, judgement or opinion}
		&\footnotesize{\{speaker\}'s answer is not conveying objective information about the world; it conveys subjective information like \{speaker\}'s evaluation, judgement or opinion}
		\\
		\addlinespace[0.5em]
		relevance
		&\footnotesize{the answer only provides information that is relevant for the addressee in the given context}
		&\footnotesize{\{speaker\}'s answer provides information that is irrelevant for the addresse in the given context}
		\\
		\addlinespace[0.5em]
		inform. satisfaction
		&\footnotesize{the answer conveys sufficient information to answer the question of the addressee in the given context}
		&\footnotesize{\{speaker\}'s answer does not convey sufficient information to answer the question of the addressee in the given context}
		\\
		\addlinespace[0.5em]    
		inform. granularity
		&\footnotesize{the answer conveys no more information than can be normally expected in the given context}
		&\footnotesize{\{speaker\}'s answer conveys much more information than can be normally expected in the given context}
		\\
		\bottomrule
	
		\end{tabular}
		\caption[Intuitive Assumptions]{Set of ``intuitive assumptions'' used by the AssumptionEvaluator an ablated version of the AE model for comparison. \label{app:tab:cs3-intuitive-assumptions}}
		\end{table}
	
	
	 \begin{table}[h!]
		\centering
		\begin{tabular}{p{3.0cm}p{5.25cm}p{5.25cm}}
			\toprule
			name
			& assumption formulation
			& violation formulation
			\\ \midrule
			evidence
			&\footnotesize{the answer provides evidence that Paris is in France}
			&\footnotesize{the answer does not provide evidence that Paris is in France}
			\\
			\addlinespace[0.5em]
			naturalness
			&\footnotesize{the answer is a natural reaction to being threatened with a fork}
			&\footnotesize{the answer is not a natural reaction to being threatened with a fork}
			\\
			\addlinespace[0.5em]
			objectivity
			&\footnotesize{the answer is only conveying objective information about the world of The Lord of the Rings}
			&\footnotesize{the answer is conveying subjective information about the world of The Lord of the Rings}
			\\
			\addlinespace[0.5em]
			relevance
			&\footnotesize{the answer is relevant to building a house}
			&\footnotesize{the answer is not relevant to building a house}
			\\
			\addlinespace[0.5em]
			inform. satisfaction
			&\footnotesize{the answer conveys sufficient information to find the way to the nearest cafe}
			&\footnotesize{the answer does not convey sufficient information to find the way to the nearest cafe}
			\\
			\addlinespace[0.5em]
			inform. granularity
			&\footnotesize{the answer conveys no more information than can be expected given a sleepy speaker}
			&\footnotesize{the answer conveys more information than can be expected given a sleepy speaker}
			\\
			\bottomrule
			
		\end{tabular}
	\caption[Lexical Assumptions]{Set of ``lexical assumptions'' used by the AssumptionEvaluator an ablated version of the AE model for comparison. \label{app:tab:cs3-lexical-assumptions}}
	 \end{table}

\paragraph{Human module evaluation.}
As mentioned above, a single utterance may flout several maxims, or it may strongly depend on the context which maxim is most salient, making automated quantitative testing of this algorithmic step difficult.
To critically assess the AssumptionEvaluator module, we collected human data for empirical evaluation of its predictions.

We focused on comparing the performance of the AssumptionEvaluator on the Gricean assumption set (Table~\ref{app:tab:cs3-gricean-assumptions}) to human performance on the same evaluation task, eliciting human responses in an online experiment.
We used contexts and trigger utterances from dataset 1 introduced in Section~\ref{section:impl-materials} for this task.
The live experiment can be viewed under: \url{https://magpie-ea.github.io/magpie3-sage-module-evals/evaluator_evals/cs3_maxim_violation/}.
The code, data and analyses are available under: \url{https://github.com/magpie-ea/magpie3-sage-module-evals/blob/main/}.

\begin{figure*}[t!]
  \centering
  \includegraphics[width=0.95\linewidth]{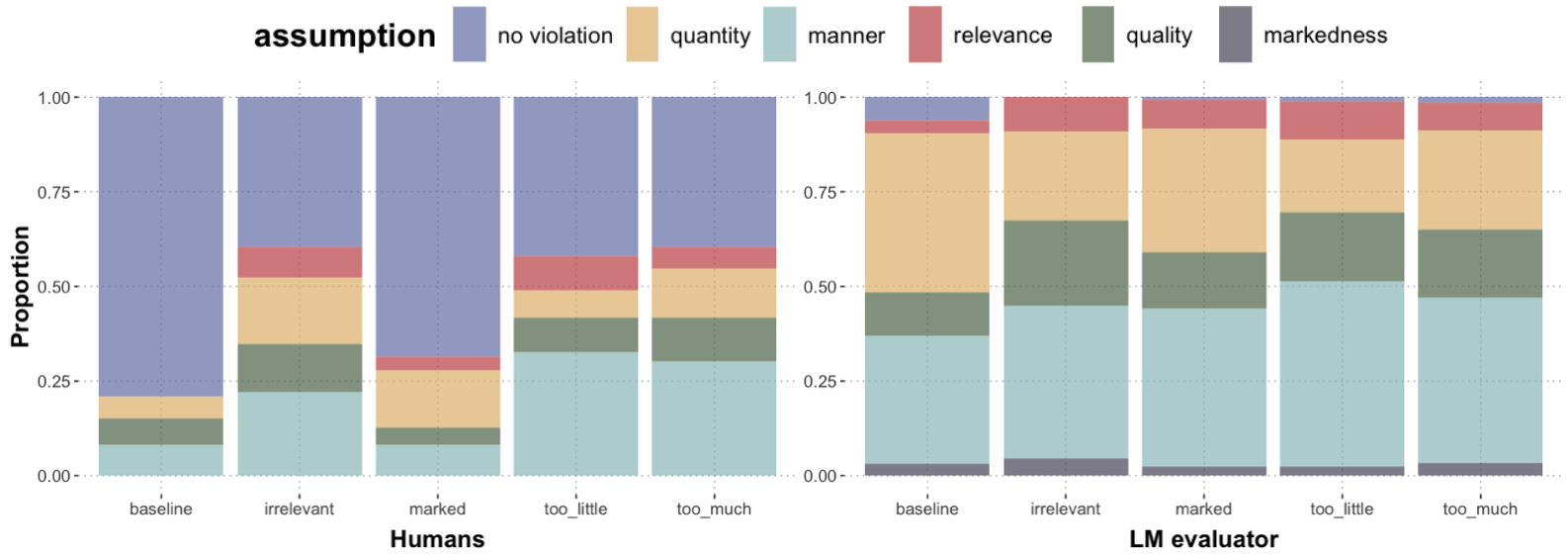}
  \caption{Results of human evaluation of the Gricean assumptions for case study III on samples from dataset 1, against the results of the \texttt{GPT-3.5-turbo}-based AssumptionEvaluator results. Human results are aggregated across items and participants, since each participant only rated one assumption per item (trial).
  \label{fig:cs3-evaluator-evals-results}}
\end{figure*}

In this experiment, we asked human participants to assess whether an assumption is likely violated, given the context and the utterance. 
We created 45 unique trials: all utterance types available in dataset 1 (baseline, irrelevant, marked, too little, too much) were combined with all possible Gricean assumptions.\footnote{Note that the AE algorithm contains a separate subloop evaluating markedness of the trigger utterance. However, since this is a different procedure and not completed by the AssumptionEvaluator which is assessed here, we did not ask human participants to evaluate markedness of the triggers in this experiment.}
Randomly sampled vignettes were used for instantiating each condition.

On each trial, participants ($N=91$) read: ``In a normal conversation, you would make the following assumption about the speaker's utterance:'', followed by an assumption sentence.
Next, they read ``Now imagine the following situation:'' followed by the context and trigger utterance. 
Then, participants answered a yes-no forced choice question whether they think the assumption is violated in the context.
Each participant saw five critical trials, with each utterance type once, combined with five distinct randomly sampled assumptions.
The critical trials were randomly shuffled with one attention checking trial which resembled visually the main trials and included an explicit instruction to answer with `Yes' or `No' in the context.
The experiment took around three minutes on average, and participants were reimbursed with \pounds0.45.
English native speaking participants based in the US and UK with at least a 95\% approval rate and at least five previously completed studies were recruited.

Five participants failed the attention checking trials and were excluded from analysis.
The rates at which remaining participants rated the assumptions as violated for different trigger types, aggregating the single assumptions by maxim the correspond to, are shown in Figure~\ref{fig:cs3-evaluator-evals-results} (left).
Visually, across utterance types, participants were most likely to not mark any assumptions as violated.
In line with intuitions, the smallest rate ofdetected violations was in the baseline trigger condition.
Among violated assumptions, assumptions of manner were judged as violated most often across most trigger conditions, but surprisingly, not the marked trigger type.
Overall, the ratings seem plausible, but no clearly different structure across utterance conditions is apparent.

Next, we focus on comparing human responses to the performance of the \texttt{GPT-3.5-turbo}-based evaluator module. The rates of assumptions identified as violated by the module are visualized in Figure~\ref{fig:cs3-evaluator-evals-results} (right). 
There is a stark contrast in the rate at which assumption violations were identified: in contrast to humans, the LM almost always evaluated the assumptions as violated. 
Although, consistent with intuitions, the proportion of no-violation evaluations was highest for the baseline utterance type in the LM results, it was below 0.1. 
However, qualitatively, similarly to human results, the proportion of manner assumption violations was higher than for other maxims.
Additionally, similar to human results, the proportion of quantity assumption violations was second-highest for most utterance types.\footnote{There is a small proportion of trials where utterances were evaluated as marked (markedness assumption) by the LM evaluator. Strictly speaking these evaluations resulted from a separate evaluation mechanism as described in Section~\ref{sec:models-CS-IIIa}, so that this assumption was not included in the human experiment. The results are visualized here for completeness.}

We note that the rates of violation judgments (both in human and LM data) might reflect the formulation of the maxims as separate assumptions.
Specifically, the maxim of manner was split into four assumptions, while the other maxims were split into two (quantity, quality) or one (relevance), which might explain the high proportion of manner violations.
Overall, while acknowledging the qualitative resemblance between the human results and the LM evaluator prediction distributions, these results suggest that more efforts are needed for the LM evaluator to provide human-like judgments of maxim flouting.

For each utterance, the AssumptionEvaluator returned the set of all assumptions for which a violation was identified, and passed it to the PlausibilityEvaluator.
This module is described next. 

\subsubsection{PlausibilityEvaluator}
\label{app:mutual-info-eval}
The traditional Gricean picture of implicature interpretation is based on interpreting the flouting of one maxim.
Therefore, we instantiated the neural PlausibilityEvaluator module for selecting one most plausible assumption violation among the set of violations returned by the AssumptionEvaluator.
This was done via identifying the most likely assumption violation, given the trigger utterance in context. 
For more robust performance, this module computed the empirical mutual information based on the (conditional) log probabilities of the violation descriptions $a$:
$$ 
\text{plausibility}(u, a) = \log \frac{P(a \mid u)}{P(a)} 
$$ 
We averaged the token-level log probabilities to compute the (conditional) log probabilities of the sequences of interest. 
To increase the plausibility of the task, we embedded the utterance in the following context:

\begin{prompt}
Context: \{context\} "\{trigger\}" \\

\noindent Someone accidentally overheard the conversation and tells their friend: 
\{assumption\}
\end{prompt}

\begin{figure*}[t!]
	\centering
	\includegraphics[scale = 0.35]{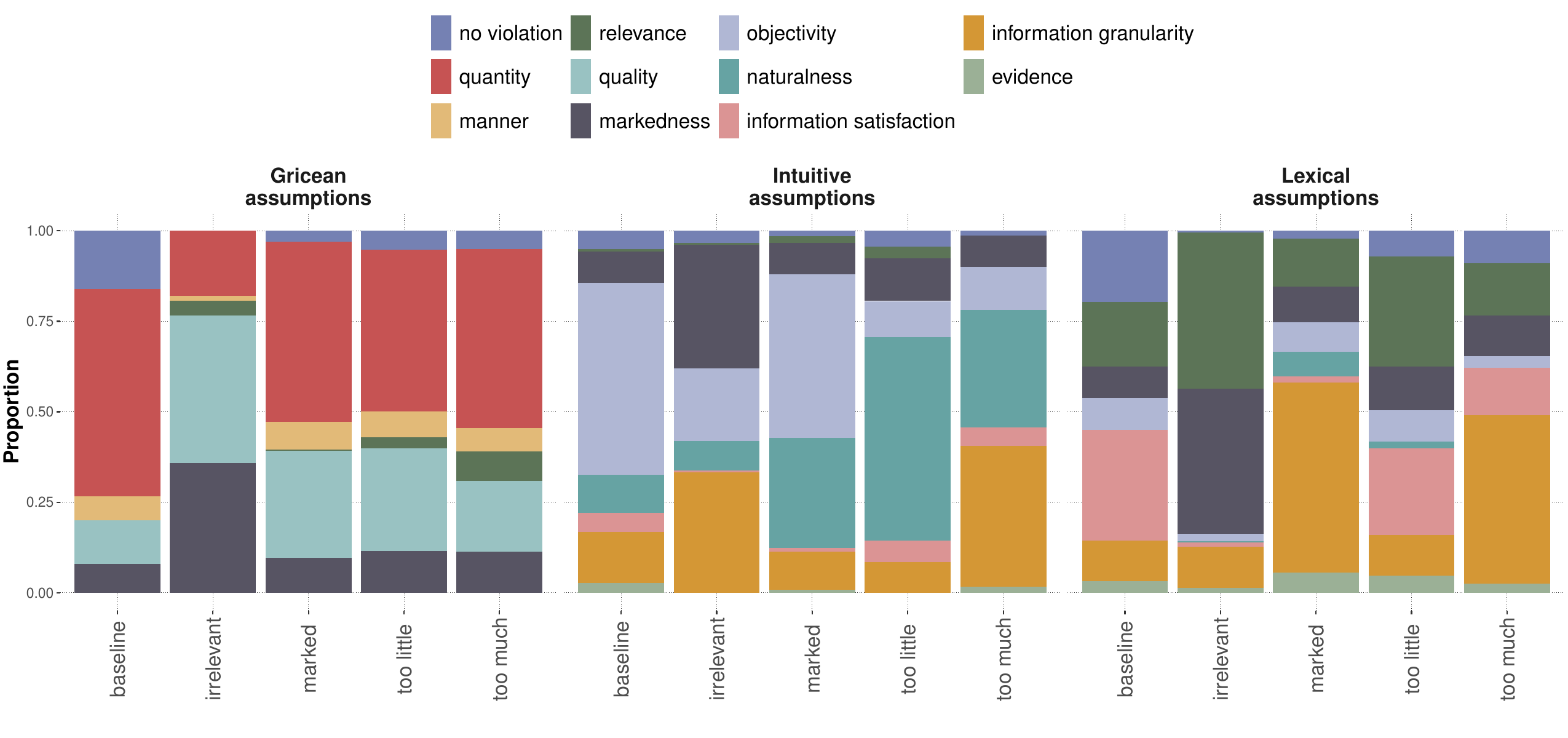}
	\caption{Most likely violations of assumptions identified by the PlausibilityEvaluator in the AE model, by assumption set (facet) and trigger type for dataset 1 (x-axis), generated by the \texttt{GPT-3.5-turbo}-based  evaluators. 
		\label{fig:app-prop-violations-by-trigger}}
\end{figure*}

Figure~\ref{fig:app-prop-violations-by-trigger} shows which assumption violation was identified as the most plausible one, for different assumption sets (facets) for given different trigger conditions (x-axis). 
We observe that the most likely assumption differed according to the trigger. 
In the Gricean assumptions AE model, matching expectations, in the too little and too much conditions, the most frequently identified assumptions are about the Maxim of Quantity. 
However, for irrelevant or marked triggers, rather unexpected violations were most likely. 
Similarly, in the baseline condition no violations were identified more frequently than in other trigger conditions, but still, violations were frequently identified. 
Given the intuitive assumption set, the naturalness and objectivity assumptions (which corresponded to the content of the Maxims of Manner and Quality) were violated very frequently, which makes intuitive sense for the marked trigger and the too little trigger conditions. 
For other triggers, the violations didn't match intuitive expectations. 
Finally, fot the AE model based on lexical assumptions, we did not formulate any particular a priori expectations about patterns of violations. 
Nonetheless, the relevance and information granularity lexical assumptions were identified as violated more frequently than other assumptions. This could be due to likelihood differences in the formulations of the lexical assumptions which led to these effects in the empirical mutual information scoring.

\subsubsection{InterpretationProposer}
\label{app:implicatures-interpretation-proposer}
The next algorithmic step according to the Gricean picture of implicature interpretation is to come up with a rationalization of the speaker's behavior, i.e., the speaker's behavior given the flouted maxim.
We implemented this with a neural InterpretationProposer, which took as input the most likely violated assumption returned by the PlausibilityEvaluator, the trigger utterance in context, and returned $n$ candidate interpretations of the utterance.
We instantiated the module using the following prompt: 

\begin{prompt}
Context: \{context\} "\{trigger utterance\}"\\
\noindent Given that \{assumption violation\}, what could \{speaker name\} be trying to convey by that answer?\\

\noindent List \{n\} guess(es). Return your answer as a list.
\end{prompt}

Note that when the set of assumptions was empty, the first part of the prompt (i.e., ``Given that \{assumption violation\}'') was dropped.
If no violations were identified in the Gricean, intuitive or lexical assumptions AE model, ``\{assumption violation\}'' was filled with a list of the (positive) assumptions about the speaker.
We used $n=4$ in all simulations.
We critically assessed the module's outputs in a human study, reported next.

\paragraph{Human module evaluation.} 
Similarly to case study II, we ran a human evaluation experiment for assessing the samples generated by the LM InterpretationProposer here.
This allows us to assess how well LMs function as proposers for interpretations, additionally to results for utterances in case study II.
The task of this module poses different challenges for the LM proposer than the task of proposing alternative utterances: it requires sampling relevant rationalizations of observed (verbal) behavior, i.e., essentially relies on abductive reasoning in search of an explanation in a potentially open-ended space.

We focused on evaluating proposals provided by the \texttt{GPT-3.5-turbo} module backbone in the Gricean AE model (i.e., operating on the full set of Gricean assumptions).
The live experiment can be found here: \url{https://magpie-ea.github.io/magpie3-sage-module-evals/proposer_evals/cs3_interpretations/}.
The code, data and analysis scripts can be found under: \url{https://github.com/magpie-ea/magpie3-sage-module-evals/blob/main/}.

We would expect that the interpretations of utterances containing a conversational implicature will contain a reference to additional information about the speaker's behavior beyond the literal content of the utterance, while interpretations of literal utterances will refer to the literal content of the utterance. 
Therefore, in this experiment, we triangulated the quality of the alternative interpretations generated by the LM proposer against manually generated pragmatic and literal interpretation options, taken from the forced choice dataset 1 (created by the authors). 
Specifically, if the proposals are natural interpretations for the given contexts, we expect them to be rated as at least equally natural as human-generated gold standard interpretations (i.e., an upper bound), 
for the four conversational implicature utterance conditions.
The literal interpretation was the upper bound option for the baseline utterance condition.
Additionally, we expect the sampled proposals to be judged as more natural interpretations than the competitor response option from the dataset (lower bound, i.e., the literal interpretation for the non-baseline utterance conditions, and the pragmatic interpretation for the baseline utterances).
To construct the materials for this experiment, we combined each trigger condition with each possible gricean assumption violation (including no violations) for randomly sampled items, resulting in 49 unique evaluation conditions. 
We then randomly sampled two proposals generated by the LM in each of these conditions.\footnote{In principle, there are 50 unique conditions, but one combination of condition and assumption violation never occurred in the simulations, so that no LM samples were available.}

On each trial, participants ($N=108$) read ``Imagine the following situation'', followed by the context, the trigger, and the assumption violation, presented as an assumption under which the interpretations should be rated.
Each trial contained two LM proposals, the upper and lower bound, presented in randomized order. 
Participants were asked to rate how natural they think the presented interpretations were for what a speaker could plausibly want to convey with the presented utterance under the assumption violation.
Participants used a slider ranging from ``terrible'' to ``natural'' (mapped to 0--100). 
As in case study II, at the beginning of the experiment, participants were told that they will help assess the quality of sentences, some of which were generated by language models.
Participants first saw an example trial in the beginning of the experiment, and there was one visually matching attention checking trials shuffled at random with the critical trials.\footnote{83\% of the participants failed the attention check by providing at least one rating that was off by more than 6 points from the instructed target rating. We ran exploratory analyses  excluding these participants and observed qualitatively matching results. Therefore, all analyses below are conducted with the full sample of participants.}
Each participant completed five critical trials from the five different trigger conditions, with randomly sampled assumptions violations.
The experiment took 7.5 minutes on average and participants were reimbursed with \pounds1.20.
English native speaking participants based in the US and UK with at least a 95\% approval rate and at least five previously completed studies were recruited.

\begin{figure*}[t!]
    \centering
    \includegraphics[width=0.7\linewidth]{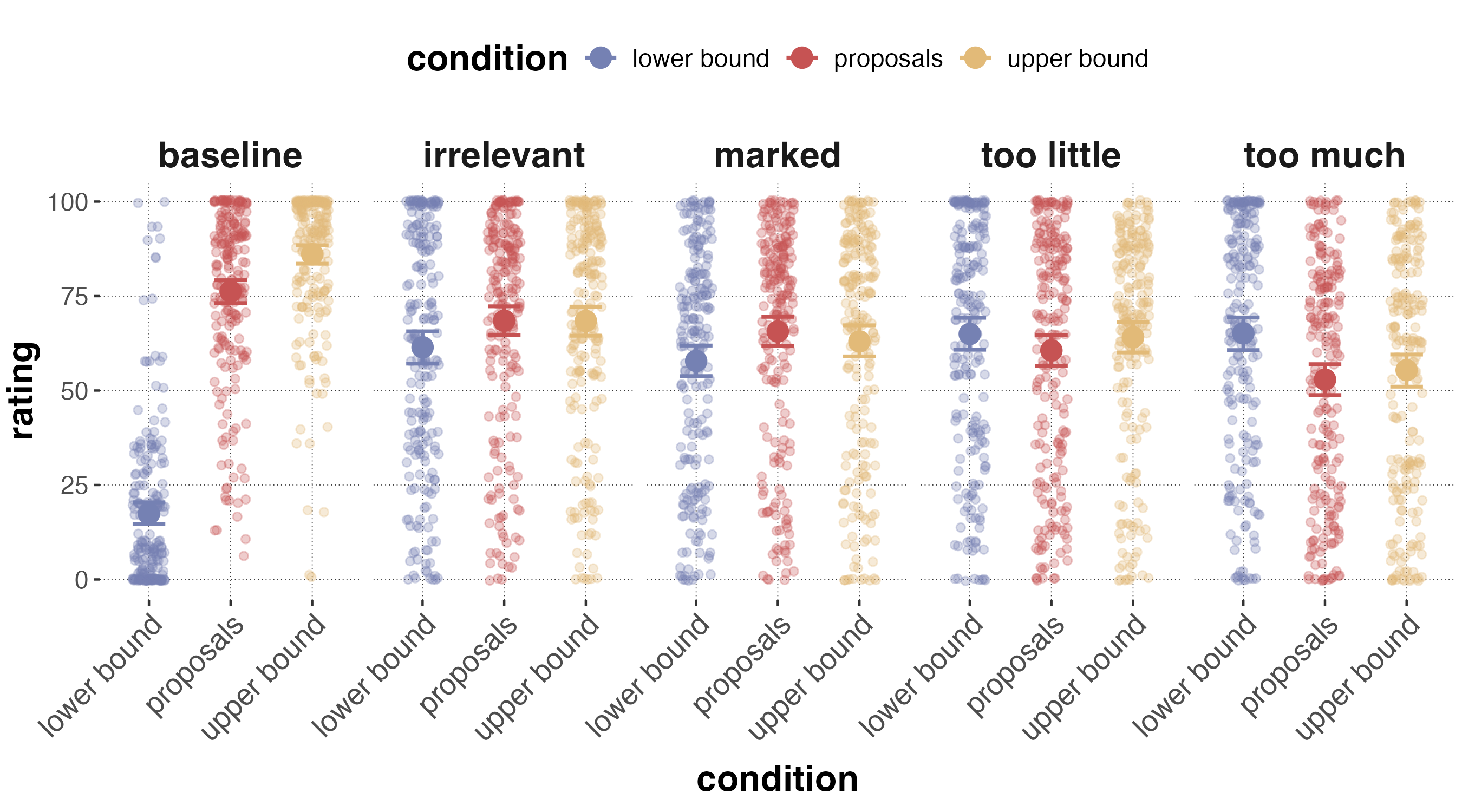}
    \caption{Results of human evaluation of the InterpretationProposer samples from case study III. \label{fig:cs3-proposer-evals-results}}
\end{figure*}

Results are shown in Figure~\ref{fig:cs3-proposer-evals-results}.
We used a Bayesian linear regression analysis, regressing the rating against main effects and the interaction of the rated utterance type (lower bound, proposal, upper bound) and inference type (utterance condition), with maximal by-subject and by-item random effect structure.\footnote{Model in R brms syntax: \texttt{rating $\sim$ condition * inference\_type + (1 | item) + (1 + condition * inference\_type | subject) + (1 + condition * inference\_type | item)}.
} 
Across utterance conditions, we found that the LM proposals were rated marginally not credibly worse than the upper bound ($\beta = 2.62 [-0.34, 5.49]$), and were rated credibly higher than the lower bound ($\beta = 11.4 [7.86, 15.09 ]$).

However, visual inspection of Figure~\ref{fig:cs3-proposer-evals-results} suggests that the rating patterns differed qualitatively by utterance type, so we compared the posterior rating contrasts separately in each of them.
For the baseline utterance type, the proposals were rated credibly worse than the upper bound ($\beta=9.88 [6.109, 13.72]$) and credibly better than the lower bound ($\beta = 58.62 [53.37, 63.62 ]$).
For the irrelevant type, the proposals were rated as higher than the lower bound, but marginally not credibly so ($\beta=6.87 [-0.7, 14.49 ]$).
For the marked type, the proposals were rated credibly higher than the lower bound ($\beta=7.9 [0.61, 14.63 ]$).
For the too much utterance type, the proposals were rated as credibly worse than the lower bound ($\beta=-12.09, [ -18.89, -5.82]$).
All other contrasts were not borne out credibly in the data.
That is, our hypothesis regarding the difference to the lower bound is borne out for all utterance types but the too little and too much types.
Our hypothesis regarding the match to the upper bound is borne out for all utterance types but the baseline.

It may at first seem puzzling that humans did not judge the human-generated lower and upper bound interpretations credibly differently in the implicature utterance conditions, except in the too-much condition; however, we note that the lower bound (i.e., literal) interpretations are still true interpretations of the contexts, and we hypothesize that humans might rate them fairly high because they might implicitly come up with additional explanations accommodating these interpretations. 
Therefore, as long as LM proposal ratings are close to the ratings of human utterances, they could arguably be considered viable alternative interpretations.

Results of the human interpretation experiment presented in Section~\ref{section:assessment-CS-III} suggest that humans did prefer one correct interpretation across utterance types, supporting the argument that human ratings presented here are not artifacts of the quality of the human-generated interpretations, but indeed provide meaningful human assessments.
Therefore, overall, we take the results of this evaluation experiment to indicate that LMs are, in principle, viable proposers for alternative interpretations of different types of implicatures, perhaps with differences across trigger conditions.
Furthermore, the performance of the AE model with no (empty) assumption set was high, indirectly suggesting that the contribution of the assumption violation information in the prompt may not have always been decisive for increasing the high baseline quality of the proposed interpretations.

\subsubsection{InterpretationMappingEvaluator}
Finally, the InterpretationMappingEvaluator module was used to map the open-ended set of proposed interpretations of the trigger utterance onto the forced choice interpretation options from the datasets.
This module was used across AE model variants.
The module identified the most likely option among the forced choice options $O$ from the dataset, given the interpretations $I$, through majority voting.
Specifically, for each generated interpretation $i_{j} \in I$, the module selects the most fitting option $o \in O$ that should be chosen if the interpretation was $i_{j}$. 
To this end, the following prompt was used:

\begin{prompt}
    \noindent Consider the following claim(s): \\

\noindent \{interpretations\} \\

\noindent Which of the following four options would you choose as a paraphrase of all/some of the claim(s) above? \\

\noindent \{options\} \\

\noindent Output the number of the most likely interpretation out of the four. Reason step by step. \\

\noindent Here the the structure of the answer:\\

\noindent"""\\
\noindent [step-by-step explanation\\
\noindent possibly over multiple lines]\\
\noindent [empty line]\\
\noindent [number] \\
\noindent """ \\

\noindent Your answer:
\end{prompt}

The final prediction of the algorithm is the most frequently selected option $o$ (with random tie breaking) over all $i \in I$. This step is called the ``MajoritySelector'' module in Algorithm~\ref{alg:interpretationAssumptionsFlow}.

\subsection{Human experiment on dataset 1}
\label{app:cs3-human-expt}
As reported in the main text (Section~\ref{section:assessment-CS-III}), for the novel materials in dataset 1, we elicited human results from 285 participants through a web-based experiment on the crowdsourcing platform Prolific.

The study was a forced-choice experiment in which participants first read instructions explaining that they will read a short description of a situation and answer a question about the situation by selecting an answer.
Each participant then completed two critical trials.
On each trial, they read the context with the trigger utterance and selected one of four interpretation options (pragmatic, literal, two distractor interpretations), presented in randomized order (see Figure~\ref{fig:flouting-implicature-example-item} for an example item).
The critical trials presented triggers in different conditions (two conditions randomly sampled from the conditions baseline, too little information, to much information, irrelevance, markedness), instantiated with different randomly sample vignettes.
The study took around two minutes and participants were reimbursed with \pounds 0.30.

\begin{figure*}[t]
\centering
\includegraphics[width=0.7\textwidth]{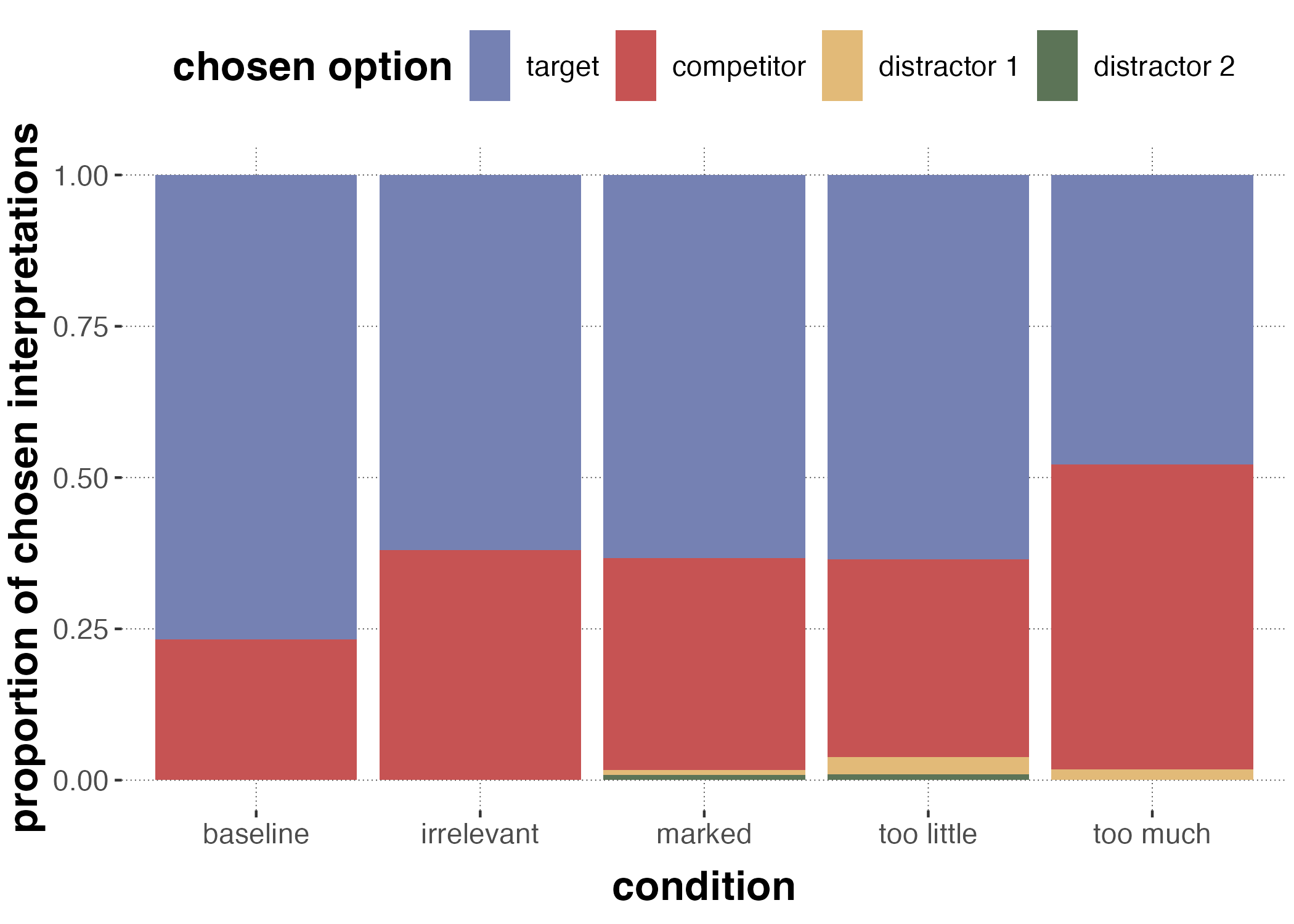}
\hfill%
\caption{
	Proportions of different responses in different trigger conditions of dataset 1 produced by humans. \label{app:fig:cs3-human-results}
}
\end{figure*}

By-condition results from the human experiment are shown in Figure~\ref{app:fig:cs3-human-results}.
First, to investigate if humans robustly preferred the pragmatic interpetation over other interpretations, we analyzed the data with a multinomial Bayesian regression model, regressing the type of chosen response (pragmatic, literal, distractor 1, distractor 2) against an intercept (pragmatic coded as the reference category).
We report posterior means and 95\% credible intervals of the contrasts.
Competitor literal responses were credibly less likely than pragmatic responses ($\beta= -0.55 [-0.73,   -0.38]$), as were distractor 1 ($\beta=-4.09 [-4.94,  -3.35]$) and distractor 2 responses ($\beta=-5.20 [-6.80, -3.99]$).\footnote{Due to an implementation mistake, in the human experiment, for item 22 the target and competitor interpretations were sometimes identical (both were target interpretation). Analyses excluding all results for item 22 revealed no qualitative or quantitative differences to the full data set. Therefore, reported analyses and figures are on the full data set.} 

Second, we explored whether there were differences in the rate of target responses between the trigger conditions (baseline, too little information, to much information, irrelevance, markedness). 
The target response was the literal option in the baseline condition, and the pragmatic option in all other conditions.
We fit a Bayesian logistic regression model predicting whether the chosen interpretation was pragmatic against a main effect of condition and by-item random intercepts. The baseline condition was coded as the reference category in dummy coding of the fixed effect.
We found that target responses were chosen less frequently than in the baseline condition in all implicature conditions ($\beta_{\text{too-little}} = -0.65 [-1.26, -0.03], \;  \beta_{\text{too-much}} = -1.27 [-1.86, -0.68], \; \beta_{\text{marked}} = -0.63 [-1.21, -0.04], \; \beta_{\text{irrelevant}} = -0.70 [-1.29,    -0.12]$).
Comparing all the implicature conditions to each other, we found a credibly lower proportion of pragmatic interpretations in the too much compared to all other conditions: the irrelevant condition ($\beta=0.56 [0.02, 1.10]$), the marked condition ($\beta=0.63 [0.10, 1.17]$) as well as the too little condition ($\beta=0.62 [0.05, 1.18]$). All other contrasts were not credible.

Taken together, this suggests that humans robustly choose target interpretations (i.e., well above chance accuracy of 25\%), although the flouting of different maxims may give rise to quantitatively different patterns observed in the different trigger conditions.

\subsection{Using different LM backbones for the AE model}
\label{app:sec:lm-configs}

The AE model was evaluated with three different LM backbones: \texttt{GPT-3.5-turbo}, \texttt{GPT-4o} and \texttt{Llama-3.1-8b-Inst.} (latter accessed through the HuggingFace platform).
Given idiosynchracies of the model performance, the interfaces to the neural modules in the AE model were minimally adjusted for each backbone as described below. 

For modules based on  \texttt{GPT-3.5-turbo} and \texttt{GPT-4o}, the mapping of the neural modules' output to structured formats expected by the symbolic modules consisted of simple string processing with regular expressions. 
The regular expressions were constructed based on the formatting instructions passed to the model (see details of the respective prompts) and manual inspections of some example outputs. The processing worked without exceptions for \texttt{GPT-4o}, while minimal occasional errors occured when \texttt{GPT-3.5-turbo} was used. 
This was possible due to the good instruction-following performance of the GPT models.

On the other hand, the instruction-following abilities of \texttt{Llama-3.1-8b-Inst.} were worse than the capabilities of the GPT models, so that more manual inspection of sample outputs was necessary for constructing custom regular expressions for output processing. 
This was due to longer chain of thought outputs of \texttt{Llama-3.1-8b-Inst.} beyond what was required, so that  retrieving the relevant part of the output was more difficult.
Based on exploratory testing and manual inspections of the outputs, the following generation parameters were used for all Llama-based neural modules: the sampling temperature was $\tau=0.8$, the maximal number of predicted tokens was set to 128, the repetition penalty to 1.8 and top $P = 0.9$.
For all backbones, if the module interfaces returned errors, respective iterations of the model simulation were discarded or repeated.

Furthermore, the \texttt{GPT-3.5-turbo}-based AE model included minor implementation inaccuracies. Specifically, the InterpretationProposer was implemented in a different way for cases where no assumption violation was identified. 
In particular, instead of not including any assumptions in the respective prompt of the InterpretationProposer, the prompt included a list of all (positive) assumptions about the speaker for proposing interpretations. 
We ran a follow-up evaluation with the corrected prompt identical to the prompt used in the no-assumptions baseline, and found no noteworthy difference, as the fraction of trials in the original results where no violations were identified was only $<5\%$. 
Specifically, the average accuracy was 0.6 for original results and 0.61 for the follow-up. 
This slight increase was mostly due to better accuracy for the lexical assumptions model (which is intuitive, as largely irrelevant information about the speaker was removed from the prompt). 

Furthermore, the markedness evaluation subloop was also inluded in the AE model with lexical assumptions, essentially adding a correct Gricean markedness assumption to the set. 
Again, follow-up experiments removing that subloop from simulations with lexical assumptions didn't reveal noteworthy differences. 
Finally, in the \texttt{GPT-3.5-turbo}-based simulations the markedness evaluation subloop relied on a manually hard-coded ranking of the typicality of the states (i.e., the pragmatic and literal interpretations passed to the subloop), instead of dynamic evaluation. 
This resulted in the fact that the markedness evaluation could artificially be blocked by the more likely literal interpretation. 
This occurred rarely for the \texttt{GPT-3.5-turbo} based implementation, so that the markedness assumption was rarely used as the most likely assumption violation overall. 
If anything, this made the \texttt{GPT-3.5-turbo} results more conservative, since results with the corrected markedness evaluation with other LMs perform better. 
We take these differences to not change anything of substance for our conclusions.

\subsection{Accuracy-based analyses}
\label{app:cs3-accuracy}

\begin{figure*}[t]
	\centering
	
	\includegraphics[width=\textwidth]{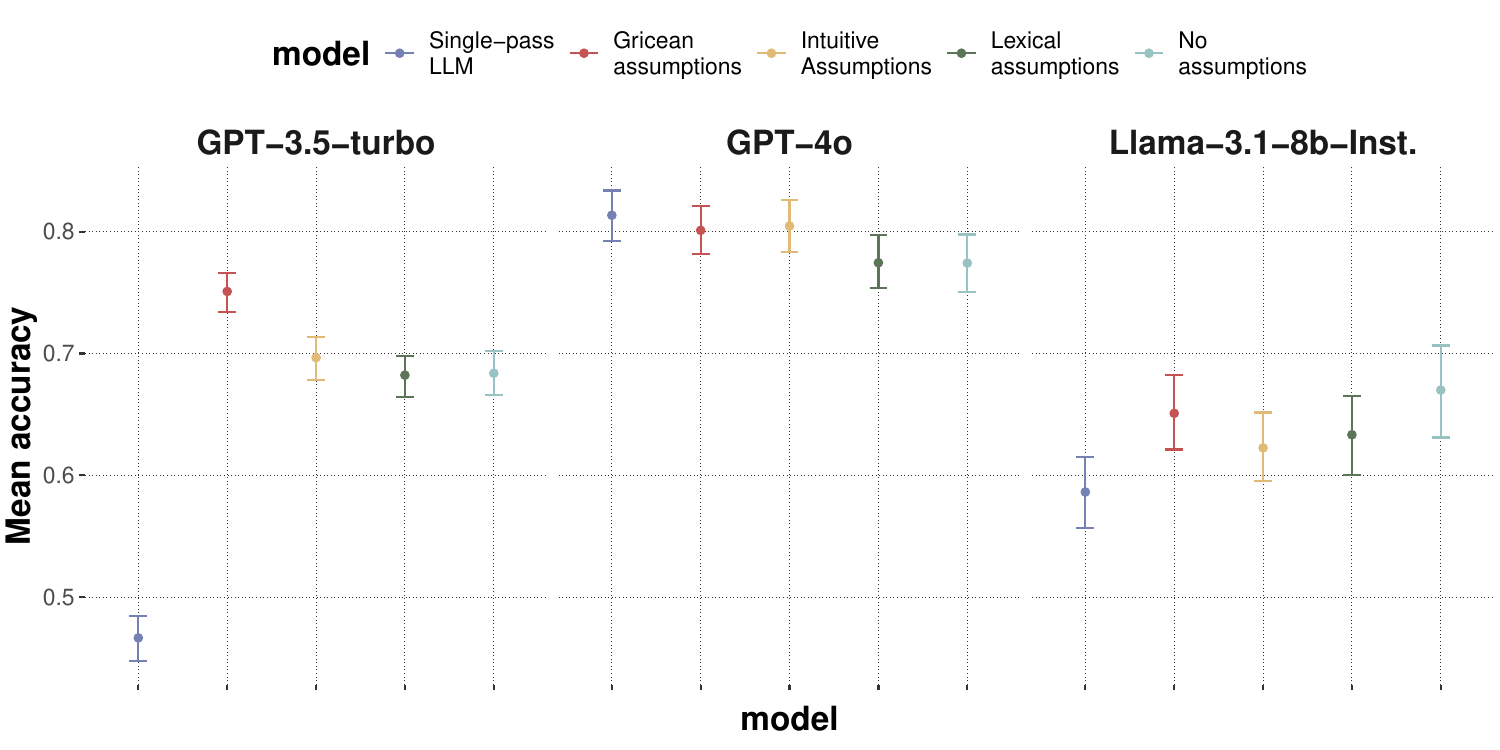}
	
	\caption{
		Interpretation results in case study III. The average accuracy (i.e., proportion of correct selections of the intended interpretation option, given the trigger utterance in context) is shown for each LM backbone (facets), across datasets and simulations (y-axis), by model (x-axis). Chance accuracy is 25\%. The error bars indicate 95-\% bootstrapped credible intervals.
		\label{fig:app-cs3-avg-acc-byBackbone}}
\end{figure*}

\begin{figure*}[h!]
\centering

\begin{subfigure}[b]{\textwidth}
    \centering
    \includegraphics[scale = 0.35]{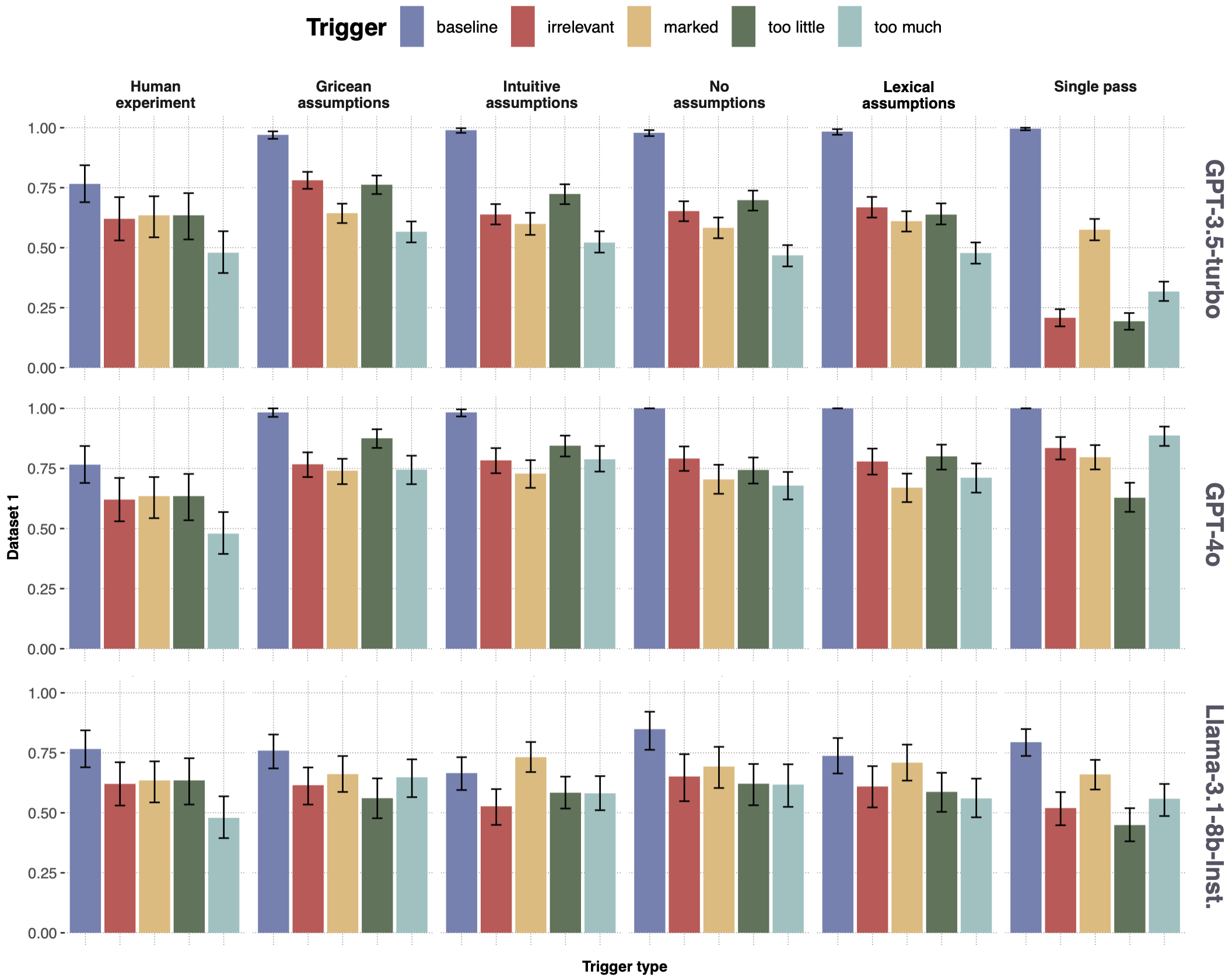}
    \caption{Implicature inference results for the novel materials introduced in this paper (dataset 1), by LM backbone (rows) and model (facets). 
    \label{fig:ours-impl-summary}}
\end{subfigure}

\begin{subfigure}[b]{\textwidth}
    \centering
    \includegraphics[scale = 0.35]{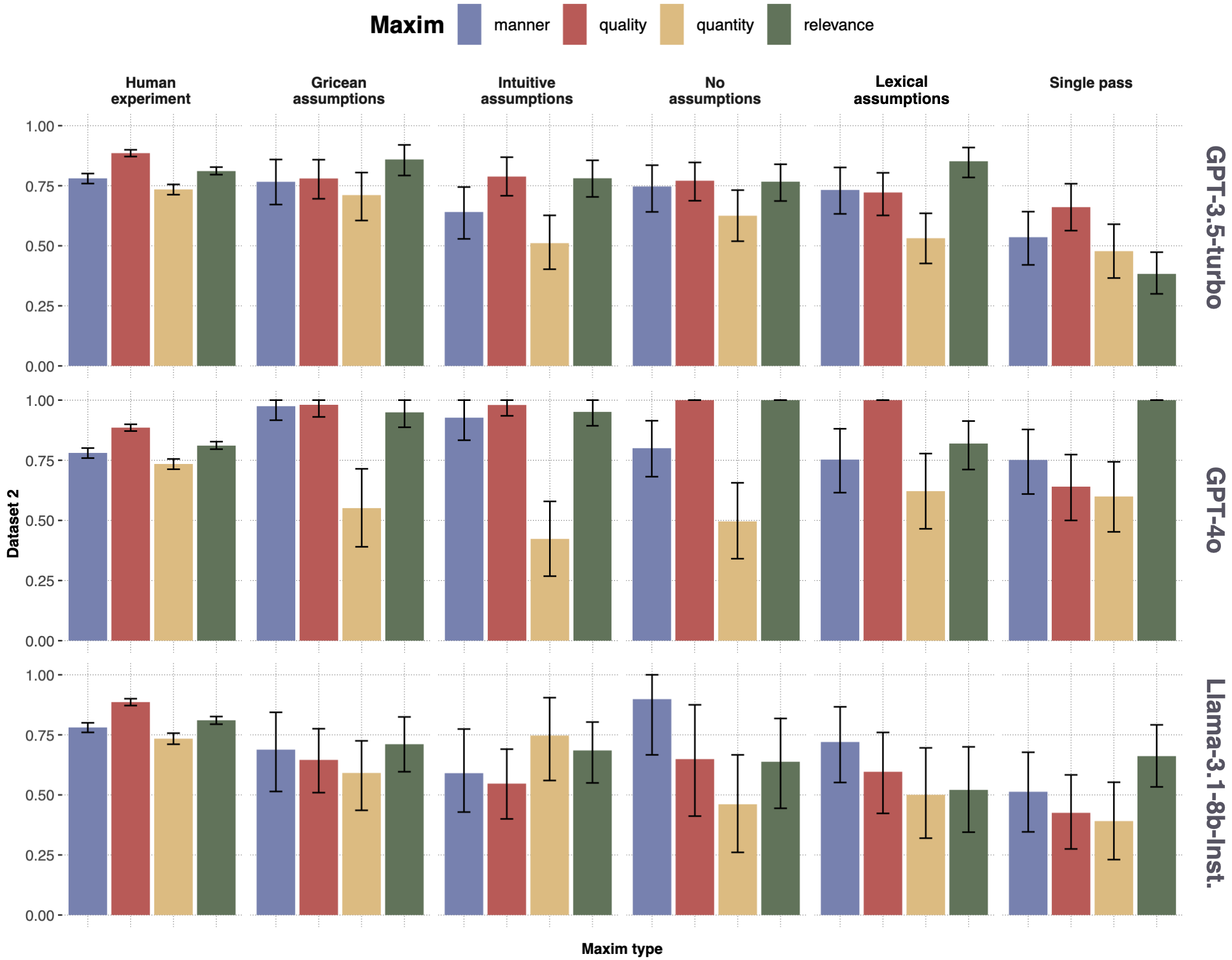}
    \caption{Implicature inference results for the materials from \citet{hu-etal-2023-fine} (dataset 2), by LM backbone (rows) and model (facets).  
    \label{fig:hu-impl-summary}}
\end{subfigure}\hfill%

\caption{
  Results from pragmatic interpretation tasks for both datasets. Y-axis shows accuracy averaged over items and different runs of the models. Error bars show bootstrapped 95\%-CIs.
  \label{fig:CS-III-results}
}

\end{figure*}

Case study III was the most extensive and ambitious case study, also investigating the effects of choosing a particular LM as the backbone for neural modules within a given SAGE model.
Therefore, next, we report more details on the analyses focusing on the accuracy of the models with different LM backbones on both datasets, expanding on the analyses reported in the main text in Section~\ref{sec:models-CS-IIIb}.

We first turn to a qualitative analysis.
Figure~\ref{fig:app-cs3-avg-acc-byBackbone} shows the average accuracy of the AE model with different assumption sets next to the single-pass LM baseline, aggregating across the two datasets and trigger conditions therein. 
%
%
Across datasets, visual inspection suggests that the AE model performed better than the SP model when instantiated with the \texttt{GPT-3.5-turbo} or \texttt{Llama-3.1-8b-Inst.} backbone, although the magnitude of the margin varied between the two.
When instantiated with \texttt{GPT-3.5-turbo}, the best model was the Gricean AE model, while the other AE models performed at a roughly similar lower accuracy.
When instantiated with \texttt{Llama-3.1-8b-Inst.}, the average model accuracy was lower than for GPT-based models; the best model was the no assumptions AE model, followed by the other AE models at a roughly similar marginally lower accuracy.
When instantiated with the \texttt{GPT-4o} backbone, the average accuracy of the models was the highest, the accuracy of the SP model being highest, on par with the Gricean and intuitive AE models, although the difference to the other two AE models did not appear very large.
All models and baselines performed well above chance (25\%).
This suggests that although the Gricean AE model was among the top performing models across LM backbones, differences between AE models based on different assumption sets were not equally pronounced between different backbones. The generally high accuracy of the no-assumptions model together with our module evaluation results indicates that, at least under the current implementation, advantages of the AE model may be driven by the performance of the UtteranceProposer, and not the AssumptionEvaluator.

Next, we report quantitative analyses of the AE model instantiated with different backbones, focusing on the role of the assumption set for accuracy of the respective model. 
We address the following questions:
\begin{enumerate}[]
	\item Does assumption evaluation matter? We operationalize this question by comparing the accuracy of the theoretically motivated AE models (i.e., with Gricean and intuitive assumptions) to the SP model, and the no-assumptions model. If yes, we expect the accuracy of the former to be higher, than of the latter.
	\item Does the formulation of assumptions matter? We operationalize this question by comparing the accuracy of the intuitive and Gricean models to the lexical model. If yes, we expect the accuracy of the former to be higher, than of the latter.
\end{enumerate}
We fit a Bayesian logistic regression model, regressing whether a response was correct against the model type, the backbone, and their interaction.\footnote{
The regression model in R syntax: \texttt{response correctness $\sim$ model * backbone}.} 
The resulting posterior means and 95\% credible intervals for contrasts of interest extracted across LM backbones and for the single LMs are reported in Table~\ref{app:tab:cs3-ds1-stats}.

\begin{table}[!ht]
	\centering
	\renewcommand{\arraystretch}{1.2}
	\small
	\begin{tabular}{|p{0.6cm}|p{3.6cm}|p{2.2cm}|p{2.2cm}|p{2.2cm}|p{2.2cm}|}
		\hline
		RQ & Contrast & GPT-3.5 & GPT-4 & Llama & Across LMs \\ \hline
		1 & Gricean > SP & 1.24 [1.12, 1.35] & -0.08 [0.27, 0.10] & 0.28 [0.08, 0.45] & 0.76 [0.68, 0.85] \\ \hline
		1 & Intuitive > SP & 0.97 [0.86, 1.08] & -0.06 [-0.26, 0.13] & 0.15 [-0.02, 0.31] & 0.57 [0.48, 0.65] \\ \hline
		1 & Lexical > SP & 0.90 [0.79, 1.01] & -0.25 [-0.43, -0.06] & 0.20 [0.003, 0.37] & 0.52 [0.44, 0.60] \\ \hline
		1 & Gricean > No & 0.33 [0.21, 0.45] & 0.16 [-0.03, 0.34] & -0.08 [-0.31, 0.12] & 0.20 [0.12, 0.29] \\ \hline
		1 & Intuitive > No & 0.06 [-0.05, 0.17] & 0.18 [-0.01, 0.37] & -0.21 [-0.43, -0.004] & 0.01 [-0.08, 0.10] \\ \hline
		1 & Lexical > No & -0.01 [-0.12, 0.11] & -0.01 [-0.19, 0.18] & -0.16 [-0.38, 0.06] & -0.04 [-0.13, 0.04] \\ \hline
		1 & avg (G + L + I) > SP & 1.03 [0.95, 1.12] & -0.13 [-0.29, 0.02] & 0.21 [0.06, 0.34] & 0.61 [0.55, 0.68] \\ \hline
		1 & avg (G + L + I) > No & 0.13 [0.04, 0.22] & 0.11 [-0.04, 0.26] & -0.15 [-0.34, 0.03] & 0.06 [-0.01, 0.13] \\ \hline
		2 & (avg (G + I) > SP) > & 0.20 [0.10, 0.30] & 0.18 [0.02, 0.33] & 0.02 [-0.15, 0.19] & 0.15 [0.07, 0.22] \\ 
		 & (L > SP) &  &  & &  \\ \hline
		2& (avg (G + I) > No) >  & 0.20 [0.10, 0.30] & 0.18 [0.02, 0.33] & 0.02 [-0.15, 0.19] & 0.11 [0.03, 0.18] \\ 
		& (L > No) &  &  &  &  \\ \hline
	\end{tabular}
	\caption{Table of posterior means and 95\% credible intervals of contrasts assessing AE models with different assumption sets across test datasets. ``G'' = Gricean, ``I'' = intuitive, ``L'' = lexical.  \label{app:tab:cs3-ds1-stats}}
\end{table}
The comparisons across backbones showed that (1) assumption evalution in the AE model mattered for improving over the SP model, but not for improving over the no-assumptions AE model; and that (2) the theoretically driven formulation of assumptions mattered for higher accuracy.
We found the same pattern for \texttt{GPT-3.5-turbo}, while for \texttt{GPT-4o} only (2) was borne out. 
For \texttt{Llama-3.1-8B-Inst.}, (1) was borne out for Gricean and lexical assumptions relative to the SP model, and (2) was not borne out. 
This suggests that a robust performance boost through assumption evaluation depended on a high-performing backbone.

Next, we inspect the accuracy of the models by condition of the trigger in the single datasets.
Figure~\ref{fig:ours-impl-summary} displays the interpretation accuracy on the first dataset (i.e., the proportion of selected target interpretations) in different trigger conditions (baseline, irrelevant, marked, too little, too much information; x-axis), in the human experiment (leftmost facet, replicating data from Figure~\ref{app:fig:cs3-human-results}) and in the different model configurations (central facets), against the single pass LM baseline (rightmost facet), compared across different LMs (panels).

Visual inspection shows that across models, performance in the literal baseline condition was better than in the remaining implicature conditions. 
The only stark contrast seemed to appear between the SP model and the AE models in the implicature conditions under \texttt{GPT-3.5-turbo}, while differences between other conditions and backbones were more nuanced. 
Within the AE models instantiated with GPT-backbones, the best-performing implicature conditions were the irrelevant and / or the too little information conditions, while for \texttt{Llama-3.1-8b-Inst.}~it was the marked condition.
The SP performance differed between the backbones, yet too little information was consistently the condition with the lowest accuracy.
On the other hand, the SP performance on the marked condition was similar to the AE models' performance.
We speculate that it might be due to the markedness condition being the only lexically driven condition which might be easier to solve for the LM without additional reasoning.  
While matching qualitatively, the figure also suggests that the \texttt{GPT-4o}-based AE model had a higher accuracy than humans. 

Turning to different assumption sets, the model with lexical assumptions showed higher performance than the SP baseline in the irrelevant, too little and too much conditions for \texttt{GPT-3.5-turbo}, and in the too little condition under \texttt{GPT-4o}, as well as in the irrelevant and too little conditions for \texttt{Llama-3.1-8b-Inst.}. 
Visually, the no-assumptions model, where the interpretation evaluation is independent of any assumption evaluation, also outperformed the SP baseline in all conditions except the baseline condition when using \texttt{GPT-3.5-turbo}, and outperformed the SP baseline in the too little but not the too much condition using \texttt{GPT-4o}. It outperformed the SP baseline in all but the marked condition for the \texttt{Llama-3.1-8b-Inst.}-based model, as well.

\begin{figure*}[t!]
\centering
\includegraphics[width=\textwidth]{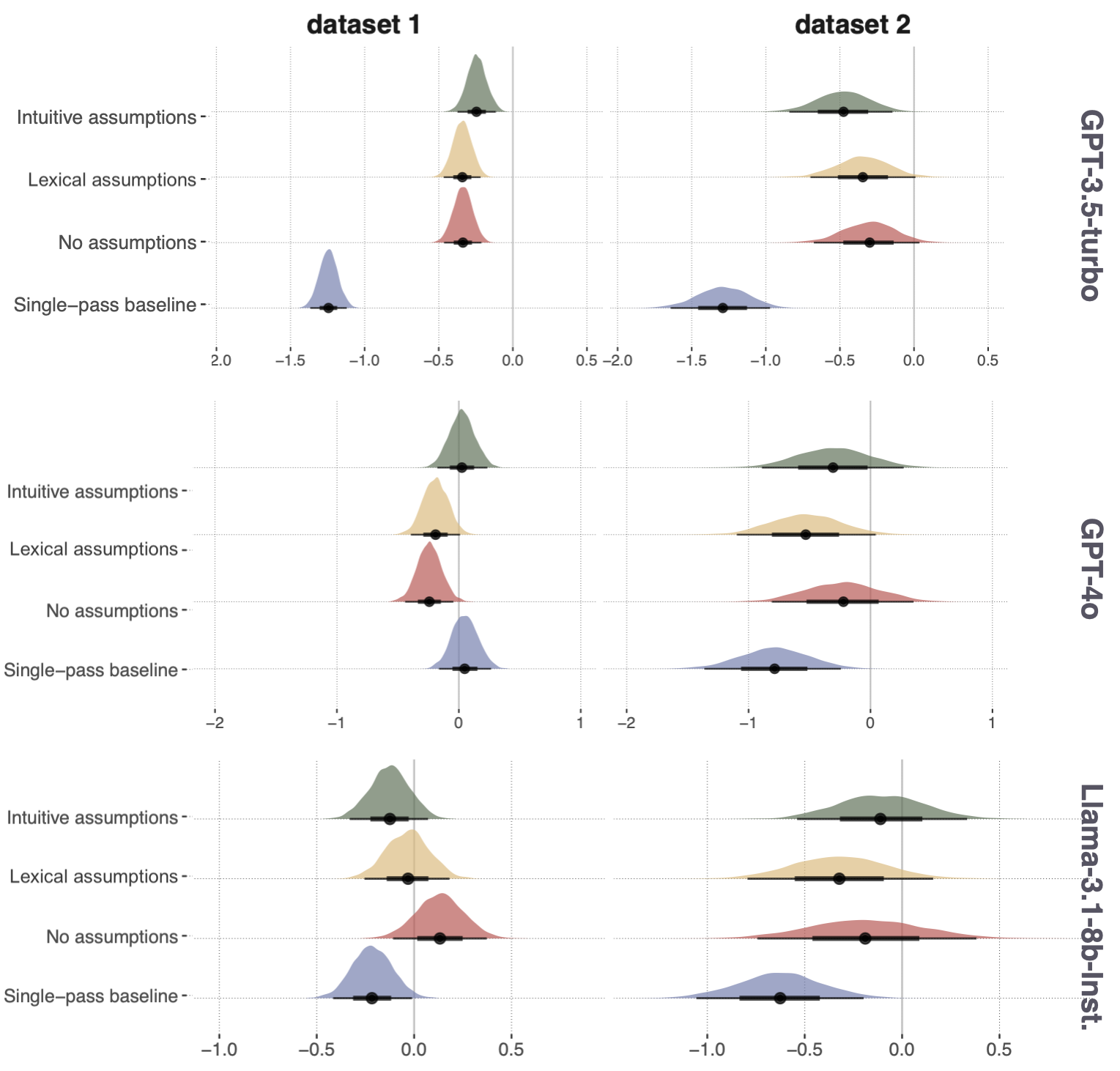}
\hfill%
\caption{
  Density plots and summary statistics for the posterior distribution of difference in (log) \textbf{accuracy} between the Gricean-assumptions model and any other model.
Black dots indicate means of posterior samples for the total log-likelihood, thick black lines indicate 66\% credible intervals, thinner black lines indicate 95\% credible intervals.  \label{fig:impl-lr-contrasts}
}
\end{figure*}

For a quantitative assessment on dataset 1, we compared the accuracy of different models by fitting a Bayesian logistic regression model on results from this dataset, regressing the correctness of the predicted responses against the model type, separately for each LM, using the R package \texttt{brms} \citep{Burkner2018:Advanced-Bayesi}.\footnote{Model in R syntax: \texttt{response $\sim$ model}.}
Using the Gricean assumptions model as reference level, Figure~\ref{fig:impl-lr-contrasts} (left half) shows the estimated posterior distribution of the contrasts of the accuracy of other models relative to the Gricean model.
For the \texttt{GPT-3.5-turbo} backbone (Figure~\ref{fig:impl-lr-contrasts}, top panel), all contrasts were credibly negative, suggesting that the Gricean assumptions model achieved the highest accuracy in this task. 
For \texttt{GPT-4o} (Figure~\ref{fig:impl-lr-contrasts}, middle panel), however, only the no-assumptions model was credibly worse than the Gricean assumptions model. 
For \texttt{Llama-3.1-8B-Inst.} (Figure~\ref{fig:impl-lr-contrasts}, bottom panel), only the SP baseline was credibly worse than the Gricean assumptions AE model.  

Turning to dataset 2 taken from \citet{hu-etal-2023-fine}, Figure~\ref{fig:hu-impl-summary} displays the interpretation accuracy (i.e., the proportion of selected target interpretations) in different trigger conditions (flouting of maxim of manner, quanlity, quantity, relevance; x-axis), in the human experiment (reported in \citet{hu-etal-2023-fine}; leftmost facet) and in the AE model configurations (middle facets), against the single pass LM baseline (rightmost facet), instantiated with different LM backbones (panels). 
First, we observed that both human experimental results and AE model results were robustly above chance (0.25). 
In addition, our best AE results outperformed the results by the best-performing model originally reported in \citep{hu-etal-2023-fine} (Fig.~\ref{fig:hu-impl-summary}; average accuracy of AE models is 0.74 with \texttt{GPT-3.5-turbo}, 0.83 with \texttt{GPT-4o}, 0.604 with \texttt{Llama-3.1-8B-Inst.}; best accuracy by \citet{hu-etal-2023-fine} was around 0.7).

We found that the AE models again outperformed the SP baseline when based on \texttt{GPT-3.5-turbo}. The improvement was smaller for the quantity condition when lexical and intuitive assumptions were used. 
Under the lexical and intuitive assumptions, visually, the models performed worse than humans in all conditions except the relevance condition. 
When \texttt{GPT-4o} was used, the AE models based on Gricean, intuitive and no assumptions outperformed humans and the SP model in all conditions except the quantity condition. 
The lexical assumptions model outperformed humans and the SP baseline in the quality condition, but performed worse that the other models in other conditions.
Under the \texttt{Llama-3.1-8b-Inst.} backbone, the no-assumptions AE model outperformed the SP model. 
For other assumption sets, the performance tended to increase for the quantity, quality and manner conditions, but not the relevance condition.
In general, for the GPT-based models, the results on dataset 2 were better than on dataset 1, while the \texttt{Llama-3.1-8b-Inst.}-based model results were better for dataset 1. 

Similarly to the statistical analysis of model performance on the first dataset, we compared the performance of different models with a Bayesian logistic regression model, regressing the correctness of the predicted responses against the main effect of model type, separately for each backbone. 
We again computed the contrasts between posterior estimates for each effect of model relative to the Gricean assumptions model. 
Posterior distributions of the contrast estimates are displayed in Figure~\ref{fig:impl-lr-contrasts} (right facet). Based on the contrast estimates, the Gricean assumptions model performed credibly better (i.e., had a higher target response choice rate across conditions) than the intuitive assumptions model and the SP baseline but not the other models, when \texttt{GPT-3.5-turbo} was used (Figure~\ref{fig:impl-lr-contrasts}, top panel). When \texttt{GPT-4o} or  \texttt{Llama-3.1-8b-Inst.} were used, the Gricean assumptions model outperformed the SP baseline (Figure~\ref{fig:impl-lr-contrasts}, middle and bottom panels), but not the other models.


\subsection{Analyses based on likelihood of human data}
\label{app:cs3-human-llh}

\subsubsection{Assessing likelihood functions based on single-example samples.}
\label{sec:assess-likel-funct}

Since SAGE models are defined as algorithmic ``feed-forward'' sampling procedures, they do not give us an explicit likelihood function over answer choices, but only produce single-example samples.
Nevertheless, it is possible to approximate the likelihoods for model predictions via sampling.
To derive our (modelers') beliefs about the likelihood prediction of model $M$ (assuming a fixed trigger condition $i$ in a given dataset), we use the samples from ``forward pass'' simulations of the model, $D^i_M = \langle c^i_t, c^i_c, c^i_d \rangle$, which give us counts for how often model $M$ chose each answer category (target, competitor, distractor).
We consider $D_M$ to be the product of a multinomial distribution with choice probabilities $p^i_M$, $D_M \sim \text{Multinom}(p^i_M)$, which are unknown. 
Assuming that, for us modelers, \textit{a priori} any probability vector is equally likely, $p^i_M \sim \text{Dirichlet}(1,1,1)$, we can use conjugacy to determine the posterior uncertainty over $p^i_M$ conditional on $D^i_M$, which captures our (modelers') uncertainty about model $M$'s predictions for condition $i$ after having observed $D^i_M$.
The modeler's uncertain distribution over the model $M$'s likelihood $LH(D^i_H, M)$ of the observed human data $D^i_H$ for condition $i$ (which also consists of counts for answer categories target, competitor and distractor) is therefore characterized by this sampling process:
\begin{align*}
    D^i_H & \sim \text{Multinom}(p^i_M)\,, \text{where} \\
    p^i_M \mid D^i_M & \sim \text{Dirichlet}(1+c^i_t, 1+c^i_c, 1+c^i_d)
\end{align*}
Intuitively, a model $M$ is a better predictor of human data $D^i_H$ for condition $i$ than another model $M'$ if $LH(D^i_H, M)$ is credibly higher than $LH(D^i_H, M')$.
A model $M$ outperforms another model $M'$ on a set of conditions, $i \in \{1, \dots, n\}$, if the total likelihood $TLH(M) = \prod_i LH(D^i_H, M)$ is credibly higher  than $TLH(M') = \prod_i LH(D^i_H, M')$.

\subsubsection{By-condition analyses}
\label{sec:more-deta-analys}

\begin{figure*}[t!]
	\centering
	\includegraphics[width=\textwidth]{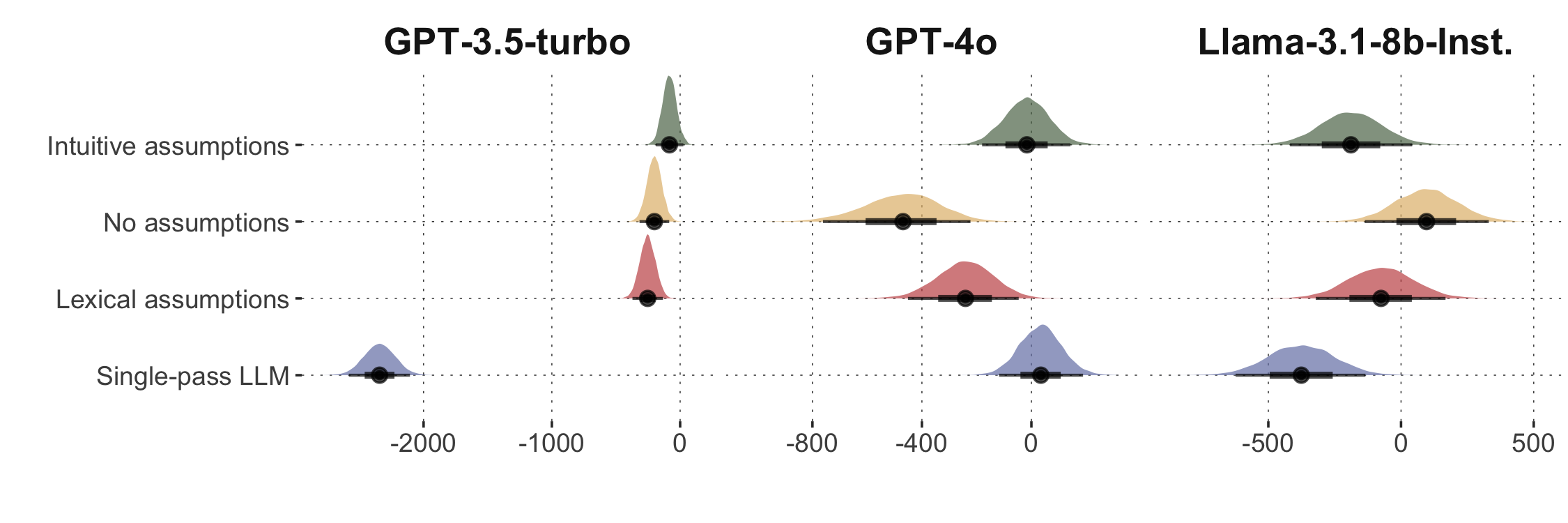}
	\hfill%
	\caption{
		Density plots and summary statistics for the posterior distribution of differences in total \textbf{log-likelihood of the human data} between the Gricean-assumptions model (lines of the plot) and any other model, for the different LM backbones (facets).
		Black dots indicate means of posterior samples for the total log-likelihood, thick black lines indicate 66\% credible intervals, thinner black lines indicate 95\% credible intervals. \label{fig:model-comparison}
	}
\end{figure*}


Extending the analyses reported in Section~\ref{sec:models-CS-IIIb}, we report the assessment of the models instantiated with different backbones in terms of their fit to human experimental data.
Figure~\ref{fig:model-comparison} shows summary statistics (means and 95\% credible intervals) from each posterior distribution of total likelihood of human data across both experiments, for each model-LM backbone pair.

Results for models based on \texttt{GPT-3.5-turbo} (Figure~\ref{fig:model-comparison} (left)) indicate that the SP baseline model was clearly a worse predictor of the distribution of human answer choices (i.e., the difference in log-likelihood under the SP model was credibly below 0).
A sampling-based approximation of the posterior probability that the model with Gricean assumptions had a higher total likelihood than the model with lexical assumptions was about $0.996$.
The posterior probability that the Gricean-assumptions model yields higher likelihood for the human data than the no-assumptions model was still high, at about $0.876$, but did not support strong conclusions.
On the other hand, for models based on the GPT-4o backbone (Figure~\ref{fig:model-comparison} (middle)), we did not observe any difference in the fit to human data between the AE and baseline model. 
The no-assumptions and the lexical assumptions AE models were credibly worse than the Gricean AE model in their fit to human data, while the intuitive assumptions model did not credibly differ from the Gricean model.

For models based on the \texttt{Llama-3.1-8B-Inst.} backbone (Figure~\ref{fig:model-comparison} (right)), the differences in the fit to human data between the models were also less pronounced. 
While the Gricean model fit human data credibly better than the SP baseline model, the other models did not credibly differ from the Gricean model.
These results highlight the importance of encompassing assessment of SAGE models beyond accuracy-based evaluations, as analyses like exemplified here may reveal a more nuanced picture of the role of the task decomposition for predicting human-like response distributions. 

\end{document}